\documentclass[twoside,11pt]{article}

\newcommand{\techrep}{}

\usepackage{jmlr2e}

\usepackage{amsmath}
\usepackage{algorithm}
\usepackage{algorithmic}
\usepackage{subfigure}
\usepackage{pstricks}
\usepackage{pst-node}
\usepackage{url}

\newcommand{\eref}[1]{(\ref{#1})}
\newcommand{\abs}[1]{\left|#1\right|}
\newcommand{\nel}[1]{\abs{#1}}
\newcommand{\avg}[1]{\left\langle #1 \right\rangle}
\newcommand{\C}[1]{\mathcal{#1}}
\newcommand{\Exp}[1]{\avg{#1}}
\newcommand{\arctanh}{\mathrm{tanh}^{-1}\,}
\newtheorem{theo}{Theorem}[section]

\newtheorem{defi}{Definition}[section]

\newcommand{\nbv}[1]{N_{#1}}                      
\newcommand{\nbf}[1]{#1}                          
\newcommand{\nbve}[2]{\nbv{#1}\setm #2}           
\newcommand{\nbfe}[2]{\nbf{#1}\setm #2}           
\newcommand{\X}[1]{\C{X}_{#1}}                    
\newcommand{\facs}{\C{F}}                         
\newcommand{\fac}[1]{\psi_{#1}}                   
\newcommand{\facx}[1]{\fac{#1}(x_{#1})}           
\newcommand{\vars}{\C{V}}                         
\newcommand{\setm}{\setminus}

\newcommand{\del}[1]{{\partial {#1}}}             
\newcommand{\dele}[2]{{\partial {#1} \setm {#2}}} 
\newcommand{\Del}[1]{{\Delta {#1}}}               
\newcommand{\Dele}[2]{{\Delta {#1} \setm {#2}}}   

\newcommand{\xd}[1]{x_\del{#1}}
\newcommand{\xD}[1]{x_\Del{#1}}

\newcommand{\xDe}[2]{x_{\Del{#1} \setm {#2}}}
\newcommand{\xnfe}[2]{x_{\nbfe{#1}{#2}}}

\newcommand{\Fac}[1]{\Psi_{#1}}
\newcommand{\Facx}[1]{\Fac{#1}(x_{\bigcup #1})}
\newcommand{\FacN}[1]{\Fac{\nbv{#1}}}
\newcommand{\FacNx}[1]{\FacN{#1}(x_\Del{#1})}
\newcommand{\FacNe}[2]{\Fac{\nbv{#1}\setm #2}}

\newcommand{\Facm}[1]{\Fac{\setm \nbv{#1}}}
\newcommand{\Facmx}[1]{\Facm{#1}(x_{\setm #1})}

\newcommand{\Zm}[1]{Z^{\setm #1}}                  
\newcommand{\Zmx}[1]{\Zm{#1}(\xd{#1})}             
\newcommand{\Za}{\zeta}                            
\newcommand{\Zma}[1]{\Za^{\setm #1}}               
\newcommand{\Zmax}[1]{\Zma{#1}(\xd{#1})}           
\newcommand{\Zi}{\zeta_0}                          
\newcommand{\Zmi}[1]{\Zi^{\setm #1}}               
\newcommand{\Zmix}[1]{\Zmi{#1}(\xd{#1})}           

\newcommand{\er}[2]{\phi^{\setm {#1}}_{#2}}        
\newcommand{\erx}[2]{\er{#1}{#2} (\xnfe{#2}{#1})}  

\newcommand{\Q}[1]{Q_{#1}}                         
\newcommand{\Qx}[1]{Q_{#1}(\xD{#1})}               

\newcommand{\M}[2]{\C{M}^{\setm #1}_{#2}}          
\newcommand{\Ma}[2]{\C{M}^{\setm #1}_{#2}}         
\newcommand{\Cu}[2]{\C{C}^{\setm #1}_{#2}}         

\newcommand{\A}{\C{A}}
\newcommand{\B}{\C{B}}
\newcommand{\ssum}[3]{\sum_{#1 \in \C{P}_{#3}(#2)}}
\newcommand{\Om}[2]{\Omega^{\setm #1}_{#2}}
\newcommand{\Ga}[2]{\Gamma^{\setm #1}_{#2}}
\newcommand{\T}[2]{T^{\setm #1}_{#2}}

\newcommand{\sAeijk}{\ssum{\A}{\dele{i}{\{j,k\}}}{+}}
\newcommand{\sAoijk}{\ssum{\A}{\dele{i}{\{j,k\}}}{-}}
\newcommand{\sAeij}{\ssum{\A}{\dele{i}{j}}{+}}
\newcommand{\sAoij}{\ssum{\A}{\dele{i}{j}}{-}}
\newcommand{\sBeji}{\ssum{\A}{\dele{j}{i}}{+}}
\newcommand{\sBoji}{\ssum{\A}{\dele{j}{i}}{-}}
\newcommand{\sAei}{\ssum{\A}{\del{i}}{+}}
\newcommand{\sAoi}{\ssum{\A}{\del{i}}{-}}

\graphicspath{{eps/}}

\ifthenelse{\isundefined{\techrep}}{\jmlrheading{vol}{year}{pages}{submitted}{published}{Joris Mooij and Bert Kappen}}{}

\firstpageno{1}

\begin{document}

\ifthenelse{\isundefined{\techrep}}{
	\title{Loop corrections for approximate inference on factor graphs}
	\ShortHeadings{Loop corrections for approximate inference on factor graphs}{Mooij and Kappen}
}{
	\title{Loop corrections for approximate inference}
	\ShortHeadings{Loop corrections for approximate inference}{Mooij and Kappen}
}

\author{\name Joris Mooij \email j.mooij@science.ru.nl \\
	\addr Department of Biophysics \\ 
	Radboud University Nijmegen \\
	6525 EZ Nijmegen, The Netherlands \\
	\AND
	\name Bert Kappen \email b.kappen@science.ru.nl \\
	\addr Department of Biophysics \\ 
	Radboud University Nijmegen \\
	6525 EZ Nijmegen, The Netherlands}

\ifthenelse{\isundefined{\techrep}}{\editor{Editor}}{\editor{}}

\maketitle

\begin{abstract}
We propose a method for improving approximate inference methods that corrects
for the influence of loops in the graphical model. The method is applicable to
arbitrary factor graphs, provided that the size of the Markov blankets is not
too large. It is an alternative implementation of an idea introduced recently
by \citet{MontanariRizzo05}.
In its simplest form, which
amounts to the assumption that no loops are present, the method reduces to the
minimal Cluster Variation Method approximation (which uses maximal factors
as outer clusters). On the other hand, using estimates of the effect of loops (obtained by some
approximate inference algorithm) and applying the Loop Correcting (LC) method usually 
gives significantly better results than applying the approximate inference algorithm
directly without loop corrections. Indeed, we often observe that the loop corrected
error is approximately the square of the error of the approximate inference method
used to estimate the effect of loops. We compare 
different variants of the Loop Correcting method with other approximate
inference methods on a variety of graphical models, including ``real world'' 
networks, and conclude that the LC approach generally obtains the most accurate 
results.
\end{abstract}


\section{Introduction}

In recent years, much research has been done in the field of approximate
inference on graphical models. One of the goals is to obtain accurate approximations of
marginal probabilities of complex probability distributions defined over many
variables, using limited computation time and memory. This research has led to
a large number of approximate inference methods. Apart from sampling
(``Monte Carlo'') methods, the most well-known methods and algorithms are
variational approximations such as Mean Field (MF), which originates in
statistical physics \citep{Parisi88}; Belief Propagation (BP), also known as
the Sum-Product Algorithm and as Loopy Belief Propagation
\citep{Pearl88,KschischangFreyLoeliger01}, which is directly related to the
Bethe approximation used in statistical physics
\citep{Bethe35,YedidiaFreemanWeiss05}; the Cluster Variation Method (CVM)
\citep{Pelizzola05} and other region-based approximation methods
\citep{YedidiaFreemanWeiss05}, which are related to the Kikuchi approximation
\citep{Kikuchi51}, a generalization of the Bethe approximation using larger
clusters; Expectation Propagation (EP) \citep{Minka01}, which includes TreeEP
\citep{MinkaQi04} as a special case.  To calculate the results of CVM and other
region based approximation methods, one can use the Generalized Belief
Propagation (GBP) algorithm \citep{YedidiaFreemanWeiss05} or double-loop algorithms that
have guaranteed convergence \citep{Yuille02,HeskesAlbersKappen03}.

It is well-known that Belief Propagation yields exact results if the graphical
model is a tree, or, more generally, if each connected component is a tree.  If
the graphical model does contain loops, BP can still yield surprisingly
accurate results using little computation time. However, if the influence of
loops is large, the approximate marginals calculated by BP can have large
errors and the quality of the BP results may not be satisfactory. One way to
correct for the influence of short loops is to increase the cluster size of the
approximation, using CVM (GBP) with clusters that subsume as many loops as
possible. However, choosing a good set of clusters is highly nontrivial
\citep{WellingMinkaTeh05}, and in general this method will only work if the
clusters do not have many intersections, or in other words, if the loops do not
have many intersections. Another method that corrects for loops to a certain
extent is TreeEP, which does exact inference on the base tree, a subgraph of
the graphical model which has no loops, and approximates the other
interactions. This corrects for the loops that consist of part of the base tree
and exactly one additional factor and yields good results if the graphical
model is dominated by the base tree, which is the case in very sparse models.
However, loops that consist of two or more interactions that are not part of
the base tree are approximated in a similar way as in BP. Hence, for 
denser models, the improvement of TreeEP over BP usually diminishes. 

In this article we propose a method that takes into account \emph{all} the
loops in the graphical model in an approximate way and therefore obtains more
accurate results in many cases. Our method is a variation on the theme
introduced by \cite{MontanariRizzo05}. The basic idea is to first estimate the
cavity distributions of all variables and subsequently improve these estimates
by cancelling out errors using certain consistency constraints. A cavity
distribution of some variable is the probability distribution on its Markov
blanket (all its neighbouring variables) of a modified graphical model, in
which all factors involving that variable have been removed. The removal of the
factors breaks all the loops in which that variable takes part. This allows an
approximate inference algorithm to estimate the strength of these loops in
terms of effective interactions or correlations between the variables of the
Markov blanket. Then, the influence of the removed factors is taken into
account, which yields accurate approximations to the probability distributions
of the original graphical model. Even more accuracy is obtained by imposing
certain consistency relations between the cavity distributions, which results in
a cancellation of errors to some extent. This error cancellation is done by a
message passing algorithm which can be interpreted as a generalization of BP in
the pairwise case and of the minimal CVM approximation in general. Indeed,
the assumption that no loops are present, or equivalently, that the cavity
distributions factorize, yields the BP / minimal CVM results. On the other
hand, using better estimates of the effective interactions in the cavity
distributions yields accurate loop corrected results.

Although the basic idea underlying our method is very
similar to that described in \citep{MontanariRizzo05}, the alternative
implementation that we propose here offers two advantages. Most importantly, it
is directly applicable to arbitrary factor graphs, whereas the original method
has only been formulated for the rather special case of graphical models with
binary variables and pairwise factors, which excludes e.g.\ many interesting
Bayesian networks. Furthermore, our implementation appears to be more robust
and also gives improved results for relatively strong interactions, as will be
shown numerically.

This article is organised as follows. First we explain the theory behind our
proposed method and discuss the differences with the original method by
\cite{MontanariRizzo05}. Then we report extensive numerical experiments
regarding the quality of the approximation and the computation time, where we
compare with other approximate inference methods. Finally, we discuss the
results and state conclusions.

\section{Theory}\label{sec:theory}

In this work, we consider graphical models such as Markov random fields and
Bayesian networks. We use the general factor graph representation since it
allows for formulating approximate inference algorithms in a unified way
\citep{KschischangFreyLoeliger01}. In the next subsection, we introduce our
notation and basic definitions.

\subsection{Graphical models and factor graphs}

Consider $N$ discrete random variables $\{x_i\}_{i\in\vars}$ with $\vars := \{1,\dots,N\}$.
Each variable $x_i$ takes values in a discrete domain $\X{i}$. 
We will use the following multi-index notation: for any subset $I \subseteq
\vars$, we write $x_I := (x_{i_1}, x_{i_2}, \dots, x_{i_m})$ if $I = \{i_1, i_2,
\dots, i_m\}$ and $i_1 < i_2 < \dots i_m$.
We consider a probability distribution over $x = (x_1,\dots,x_N)$ that can be written
as a product of factors $\fac{I}$:
  \begin{equation}\label{eq:probability_distribution}
  P(x) = \frac{1}{Z} \prod_{I\in\facs} \facx{I}, \qquad Z = \sum_{x} \prod_{I \in \facs} \facx{I}.
  \end{equation}
The factors (which we will also call ``interactions'') are indexed by (small)
subsets of $\vars$, i.e.\ $\facs \subseteq \C{P}(\vars) := \{I : I \subseteq \vars\}$. 
Each factor is a non-negative function $\fac{I} : \prod_{i\in I}\X{i} \to
[0,\infty)$.  For a Bayesian network, the factors are conditional probability
tables. In case of Markov random fields, the factors are often called
potentials (not to be confused with statistical physics terminology, where
``potential'' refers to minus the logarithm of the factor instead). Henceforth,
we will refer to a triple $(\vars,\facs,\{\fac{I}\}_{I\in\facs})$ that
satisfies the description above as a discrete \emph{graphical model} (or
\emph{network}).

In general, the normalizing constant $Z$ is not known and exact computation of
$Z$ is infeasible, due to the fact that the number of terms to be summed is
exponential in $N$. Similarly, computing marginal distributions $P(x_J)$ of $P$
for subsets of variables $J \subseteq \vars$ is intractable in general.
In this article, we focus on the task of accurately approximating single node
marginals $P(x_i) = \sum_{x_{\vars\setm i}}P(x)$.

We can represent the structure of the probability distribution
\eref{eq:probability_distribution} using a \emph{factor graph}. This is a
bipartite graph, consisting of \emph{variable nodes} $i \in \vars$ and
\emph{factor nodes} $I \in \facs$, with an edge between $i$ and $I$ if and only
if $i \in I$, i.e.\ if $x_i$ participates in the factor $\fac{I}$. We will
represent factor nodes visually as rectangles and variable nodes as circles.
See Figure \ref{fig:cavities}(a) for an example of a factor graph.  We
denote the neighbouring nodes of a variable node $i$ by $\nbv{i} := \{I \in
\facs : i \in I\}$ and the neighbouring nodes of a factor node $I$ simply by $\nbf{I}
= \{i \in \vars : i \in I\}$. Further, we define for each variable $i \in \vars$ the
set $\Del{i} := \bigcup \nbv{i}$ consisting of all variables that appear in some
factor in which variable $i$ participates, and the set $\del{i} := \Del{i}
\setm \{i\}$, the \emph{Markov blanket} of $i$.

In the following, we will often abbreviate the set theoretical notation $X
\setm Y$ (i.e.\ all elements in $X$ that are not in $Y$) by $\setm Y$ if it is
obvious from the context what the set $X$ is. Further, we will use lowercase for
variable indices and uppercase for factor indices. For convenience, we will define
for any subset $\C{I} \subset \facs$ the product of the corresponding factors:
  \begin{equation*}
  \Facx{\C{I}} := \prod_{I \in \C{I}} \facx{I}.
  \end{equation*}

\subsection{Cavity networks and loop corrections}

The notion of a \emph{cavity} stems from statistical physics, where it was used
originally to calculate properties of random ensembles of certain graphical models
\citep{MezardParisiVirasoro87}. A cavity is obtained by removing one variable
from the graphical model, together with all the factors in which that variable
participates.

\begin{figure}[t]
\centering
\subfigure[Original factor graph]{\psset{unit=1cm}
\begin{pspicture}(0,0)(5.639,5.417)
\tiny
\cnodeput(3.181,1.944){i}{$i$}
\cnodeput(4.806,3.264){j}{$j$}
\cnodeput(4.792,1.097){k}{$k$}
\cnodeput(2.639,0.264){l}{$l$}
\cnodeput(1.333,1.972){m}{$m$}
\cnodeput(2.736,3.764){n}{$n$}
\cnodeput(4.181,5.153){o}{$o$}
\rput(4.056,2.583){\rnode{I}{\psframebox{$I$}}}
\rput(3.625,0.861){\rnode{J}{\psframebox{$J$}}}
\rput(2.264,2.681){\rnode{K}{\psframebox{$K$}}}
\rput(3.903,4.097){\rnode{L}{\psframebox{$L$}}}
\rput(5.375,2.181){\rnode{M}{\psframebox{$M$}}}
\rput(0.264,2.208){\rnode{N}{\psframebox{$N$}}}
\rput(1.583,0.806){\rnode{O}{\psframebox{$O$}}}
\ncline{I}{i}
\ncline{I}{j}
\ncline{J}{i}
\ncline{J}{k}
\ncline{J}{l}
\ncline{K}{i}
\ncline{K}{m}
\ncline{K}{n}
\ncline{L}{j}
\ncline{L}{n}
\ncline{L}{o}
\ncline{M}{j}
\ncline{M}{k}
\ncline{N}{m}
\ncline{O}{l}
\ncline{O}{m}
\end{pspicture}}
\qquad\qquad\qquad
\subfigure[Cavity graph of $i$]{\psset{unit=1cm}
\begin{pspicture}(0,0)(5.639,5.417)
\tiny
\cnodeput(4.806,3.264){j}{$j$}
\cnodeput(4.792,1.097){k}{$k$}
\cnodeput(2.639,0.264){l}{$l$}
\cnodeput(1.333,1.972){m}{$m$}
\cnodeput(2.736,3.764){n}{$n$}
\cnodeput(4.181,5.153){o}{$o$}
\rput(3.903,4.097){\rnode{L}{\psframebox{$L$}}}
\rput(5.375,2.181){\rnode{M}{\psframebox{$M$}}}
\rput(0.264,2.208){\rnode{N}{\psframebox{$N$}}}
\rput(1.583,0.806){\rnode{O}{\psframebox{$O$}}}
\ncline{L}{j}
\ncline{L}{n}
\ncline{L}{o}
\ncline{M}{j}
\ncline{M}{k}
\ncline{N}{m}
\ncline{O}{l}
\ncline{O}{m}
\end{pspicture}}
\caption{\small\label{fig:cavities}
(a) Original factor graph, corresponding to the probability distribution $P(x) = \frac{1}{Z} \fac{L}(x_j,x_n,x_o)
\fac{I}(x_i,x_j)\fac{M}(x_j,x_k)\fac{N}(x_m)\fac{K}(x_i,x_m,x_n)\fac{J}(x_i,x_k,x_l)\fac{O}(x_l,x_m)$;
(b) Factor graph corresponding to the cavity network of variable $i$, obtained by removing variable $i$ and the factor nodes that contain $i$
(i.e.\ $I$,$J$ and $K$). The Markov blanket of $i$ is $\del{i} = \{j,k,l,m,n\}$. The cavity distribution 
$\Zmx{i}$ is the (unnormalized) marginal on $\xd{i}$ of the probability distribution corresponding to the
cavity graph (b).}
\end{figure}
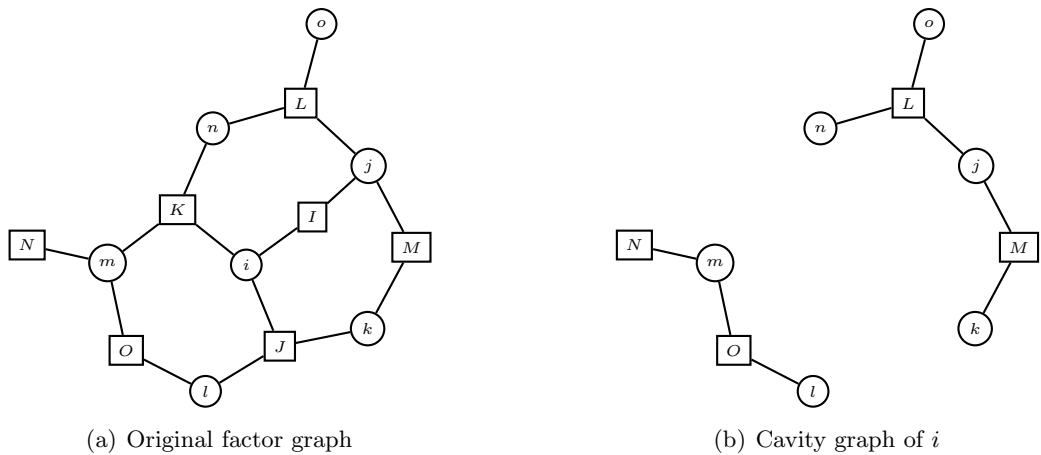

In our context, we define cavity networks as follows (see also Figure \ref{fig:cavities}):
\begin{defi}
Given a graphical model $(\vars,\facs,\{\fac{I}\}_{I\in\facs})$ and a variable 
$i \in \vars$, the \emph{cavity network} of variable $i$ is the graphical model 
$(\vars\setm i, \facs \setm \nbv{i}, \{\fac{I}\}_{I\in\facs\setm\nbv{i}})$.
\end{defi}
The probability distribution corresponding to the cavity network of variable $i$
is thus proportional to:
  \begin{equation*}
  \Facmx{i} = \prod_{\substack{I\in\facs\\ i \not\in I}} \facx{I}.
  \end{equation*}
Summing out all the variables, except for the neighbours $\del{i}$ of $i$, gives
what we will call the \emph{cavity distribution}:
\begin{defi}
Given a graphical model $(\vars,\facs,\{\fac{I}\}_{I\in\facs})$ and a variable $i \in \vars$, the 
\emph{cavity distribution} of $i$ is
  \begin{equation}\label{eq:def_cavity_dist}
  \Zmx{i} := \sum_{x_{\setm \Del{i}}} \Facmx{i}.
  \end{equation}
\end{defi}
Thus the cavity distribution of $i$ is proportional to the marginal of the
cavity network of $i$ on the Markov blanket $\del{i}$. The cavity distribution
describes the \emph{effective} interactions (or correlations) induced by the cavity
network on the neighbours $\del{i}$ of variable $i$. Indeed, from equations
\eref{eq:probability_distribution} and \eref{eq:def_cavity_dist} and the
trivial observation that $\Fac{\facs} = \FacN{i} \Facm{i}$ we conclude:
\begin{equation}\label{eq:cav2org}
  P(\xD{i}) 
  \propto \Zmx{i} \FacNx{i}.
  \end{equation}
Thus given the cavity distribution $\Zmx{i}$, one can calculate the marginal
distribution of the original graphical model $P$ on $\xD{i}$, provided that the
cardinality of $\X{\Del{i}}$ is not too large. 

In practice, exact cavity distributions are not known, and the only way to
proceed is to use approximate cavity distributions.  Given some approximate
inference method (e.g.\ BP), there are two ways to calculate $P(\xD{i})$:
either use the method to approximate $P(\xD{i})$ directly, or use the method to
approximate $\Zmx{i}$ and use relation \eref{eq:cav2org} to obtain an
approximation to $P(\xD{i})$. The latter method generally gives more accurate
results, since the complexity of the cavity network is less than that of the
original network. In particular, the cavity network of variable $i$ contains no
loops involving that variable, since all factors in which $i$ participates have
been removed (e.g.\ the loop $i-J-l-O-m-K-i$ in the original network,
Figure \ref{fig:cavities}(a), is not present in the cavity network, Figure
\ref{fig:cavities}(b)). 
Thus the latter method of calculating $P(\xD{i})$ takes into account loops
involving variable $i$, although in an approximate way. It does not, however,
take into account the other loops in the original graphical model. The basic
idea of the loop correction approach of \cite{MontanariRizzo05} is to use the
latter method for all variables in the network, but to adjust the approximate
cavity distributions in order to cancel out approximation errors before
\eref{eq:cav2org} is used to obtain the final approximate marginals. This
approach takes into account \emph{all} the loops in the original network, in an
approximate way.

This basic idea can be implemented in several ways. Here we propose an
implementation which we will show to have certain advantages over the original
implementation proposed in \citep{MontanariRizzo05}. In particular, it is
directly applicable to arbitrary factor graphs with variables taking an
arbitrary (discrete) number of values and factors that may contain zeroes and
consist of an arbitrary number of variables. In the remaining subsections, we
will first discuss our proposed implementation in detail. In section
\ref{sec:diffMontanariRizzo} we will discuss differences with the
original approach.

\subsection{Combining approximate cavity distributions to cancel out errors}

Suppose that we have obtained an initial approximation $\Zmix{i}$ of the
(exact) cavity distribution $\Zmx{i}$, for each $i \in \vars$. Let $i \in \vars$
and consider the approximation error of the cavity distribution of $i$,
i.e.\ the exact cavity distribution of $i$ divided by its approximation:
  \begin{equation*}
  \frac{\Zmx{i}}{\Zmix{i}}.
  \end{equation*}
In general, this is an arbitrary function of the variables $\xd{i}$. However, for
our purposes, we can \emph{approximate} the error as 
a product of factors defined on small subsets of $\del{i}$ in the following way:
  \begin{equation*}
  \frac{\Zmx{i}}{\Zmix{i}} \approx \prod_{I \in \nbv{i}} \erx{i}{I}.
  \end{equation*}
Thus we assume that the approximation error lies near a submanifold parameterized
by the error factors $\{\erx{i}{I}\}_{I\in\nbv{i}}$. If we were able to calculate these error 
factors, we could improve our initial approximation $\Zmix{i}$ 
by replacing it with the product 
  \begin{equation}\label{eq:def_Zma}
  \Zmax{i} := \Zmix{i} \prod_{I\in\nbv{i}} \erx{i}{I} \approx \Zmx{i}.
  \end{equation}
Using \eref{eq:cav2org}, this would then yield an improved approximation of $P(\xD{i})$.

It turns out that the error factors can indeed be calculated by exploiting the
redundancy of the information in the initial cavity approximations
$\{\Zmi{i}\}_{i\in\vars}$. The fact that all $\Zma{i}$ provide approximations to
marginals of the \emph{same} probability distribution $P(x)$ via
\eref{eq:cav2org} can be used to obtain consistency constraints. The number of
constraints obtained in this way is enough to solve for the unknown error factors
$\{\erx{i}{I}\}_{i\in\vars, I \in \nbv{i}}$. 

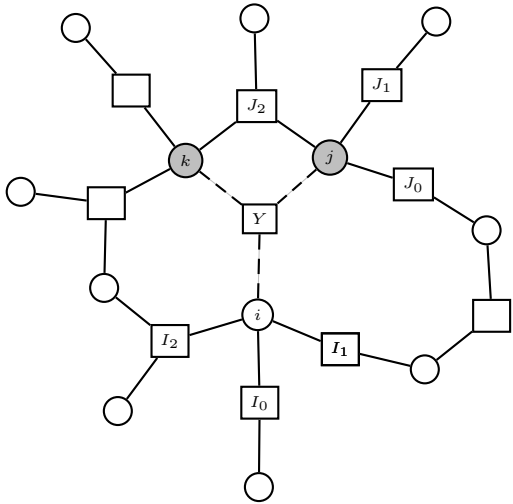
\begin{figure}[t]
\centering
\psset{unit=1cm}
\begin{pspicture}(0,0)(7,7)
\tiny
\rput(3.444,1.389){\rnode{I0}{\psframebox{$I_0$}}}
\rput(4.514,2.097){\rnode{I1}{\psframebox{$I_1$}}}
\rput(4.514,2.097){\rnode{I1}{\psframebox{$I_1$}}}
\rput(2.250,2.208){\rnode{I2}{\psframebox{$I_2$}}}
\rput(5.486,4.292){\rnode{J0}{\psframebox{$J_0$}}}
\rput(5.069,5.611){\rnode{J1}{\psframebox{$J_1$}}}
\rput(3.403,5.333){\rnode{J2}{\psframebox{$J_2$}}}
\rput(6.528,2.542){\rnode{F0}{\psframebox{\phantom{$I_0$}}}}
\rput(1.736,5.542){\rnode{F1}{\psframebox{\phantom{$I_0$}}}}
\rput(1.403,4.042){\rnode{F2}{\psframebox{\phantom{$I_0$}}}}
\rput(3.444,3.833){\rnode{Y}{\psframebox{$Y$}}}
\cnodeput(3.417,2.556){i}{$i$}
\psset{radius=0.2}
\rput(3.431,0.264){\Cnode{v0}{}}
\rput(5.639,1.833){\Cnode{v1}{}}
\rput(6.458,3.681){\Cnode{v2}{}}
\rput(1.556,1.278){\Cnode{v3}{}}
\rput(1.375,2.917){\Cnode{v4}{}}
\rput(5.792,6.458){\Cnode{v5}{}}
\rput(3.375,6.500){\Cnode{v6}{}}
\rput(1.000,6.375){\Cnode{v7}{}}
\rput(0.264,4.194){\Cnode{v8}{}}
\psset{fillcolor=lightgray,fillstyle=solid}
\rput(4.375,4.653){\circlenode{j}{$j$}}
\rput(2.458,4.611){\circlenode{k}{$k$}}
\ncline{I0}{i}
\ncline{I0}{v0}
\ncline{I1}{i}
\ncline{I1}{v1}
\ncline{I2}{i}
\ncline{I2}{v3}
\ncline{I2}{v4}
\ncline{J0}{j}
\ncline{J0}{v2}
\ncline{J1}{j}
\ncline{J1}{v5}
\ncline{J2}{j}
\ncline{J2}{k}
\ncline{J2}{v6}
\ncline{F0}{v1}
\ncline{F0}{v2}
\ncline{F1}{k}
\ncline{F1}{v7}
\ncline{F2}{k}
\ncline{F2}{v4}
\ncline{F2}{v8}
\ncline[linestyle=dashed]{Y}{i}
\ncline[linestyle=dashed]{Y}{j}
\ncline[linestyle=dashed]{Y}{k}
\end{pspicture}
\caption{\label{fig:Y_min_i} Part of the factor graph, illustrating the derivation
of \eref{eq:LCBP}. The two grey variable nodes correspond to $\nbfe{Y}{i} = \{j,k\}$.}
\end{figure}

Here we propose the following consistency constraints.
Let $Y \in \facs$, $i \in \nbf{Y}$ and $j \in \nbf{Y}$ with $i \ne j$ (see also Figure
\ref{fig:Y_min_i}). Consider the graphical model $(\vars,\facs \setm Y,\{\fac{I}\}_{I\in\facs\setm Y})$ that is
obtained from the original graphical model by removing factor $\fac{Y}$.
The product of all factors (except $\fac{Y}$) obviously satisfies:
  \begin{equation*}
  \Fac{\setm Y} = \Fac{\nbve{i}{Y}} \Facm{i} = \Fac{\nbve{j}{Y}} \Facm{j}.
  \end{equation*}
Using \eref{eq:def_cavity_dist} and summing over all $x_k$ for $k \not\in \nbfe{Y}{i}$,
we obtain the following equation, which holds for the exact cavity distributions $\Zm{i}$ and $\Zm{j}$:
  \begin{equation*}
  \sum_{x_i} \sum_{\xDe{i}{Y}} \FacNe{i}{Y} \Zm{i} = \sum_{x_i} \sum_{\xDe{j}{Y}} \FacNe{j}{Y} \Zm{j}.
  \end{equation*}
Substituting our basic assumption \eref{eq:def_Zma} on both sides and
pulling the factor $\erx{i}{Y}$ in the l.h.s.\ through the summation, we obtain:
\begin{equation}
  \er{i}{Y}\sum_{x_i} \sum_{\xDe{i}{Y}} \FacNe{i}{Y} \Zmi{i} \prod_{I \in \nbve{i}{Y}} \er{i}{I} = 
  \sum_{x_i} \sum_{\xDe{j}{Y}} \FacNe{j}{Y} \Zmi{j} \prod_{J \in \nbv{j}} \er{j}{J} 
\end{equation}
This should hold for each $j \in \nbfe{Y}{i}$. We can thus take the geometrical average of
the r.h.s.\ over all $j \in \nbfe{Y}{i}$. After rearranging, this yields:
  \begin{equation}\label{eq:LCBP}
  \er{i}{Y} = \frac{\displaystyle \left(\prod_{j\in \nbfe{Y}{i}} \sum_{x_i} \sum_{\xDe{j}{Y}} \FacNe{j}{Y} \Zmi{j} \prod_{J \in \nbv{j}} \er{j}{J} \right)^{1/\nel{\nbfe{Y}{i}}}}{\displaystyle {\sum_{x_i}} \sum_{\xDe{i}{Y}} \FacNe{i}{Y} \Zmi{i} \prod_{I \in \nbve{i}{Y}} \er{i}{I}}
  \qquad \text{for all $i \in \vars$, $Y \in \nbv{i}$.}
  \end{equation}
Note that the numerator is an approximation of the joint marginal $P^{\setm Y}(\xnfe{Y}{i})$ 
of the modified graphical model $(\vars,\facs \setm Y,\{\fac{I}\}_{I\in\facs\setm Y})$ on the variables $\nbfe{Y}{i}$.

Solving the consistency equations \eref{eq:LCBP} simultaneously for the error factors
$\{\er{i}{I}\}_{i\in\vars,I\in\nbv{i}}$ can be done using a simple fixed point
iteration algorithm, e.g.\ Algorithm \ref{alg:LCBP}. The input consists of the
initial approximations $\{\Zmi{i}\}_{i\in\vars}$ to the cavity distributions.
It calculates the error factors that satisfy \eref{eq:LCBP} by fixed point
iteration and from the fixed point, it calculates improved approximations of
the cavity distributions $\{\Zma{i}\}_{i\in\vars}$ using relation
\eref{eq:def_Zma}. \footnote{Alternatively, one could formulate the updates directly in
terms of the cavity distributions $\{\Zma{i}\}$.}
From the improved cavity distributions, we can calculate the loop corrected
approximations to the single variable marginals of the original probability
distribution \eref{eq:probability_distribution} as follows:
  \begin{equation}\label{eq:single_var_beliefs}
  b_i(x_i) \propto \sum_{\xd{i}} \FacNx{i} \Zmax{i}
  \end{equation}
where the factor $\fac{Y}$ is now included.
Algorithm \ref{alg:LCBP} uses a sequential update scheme, but other update schemes
are possible (e.g.\ random sequential or parallel). In practice, the fixed 
sequential update scheme often converges without the need for damping. 

\begin{algorithm}[bt]
\caption{\label{alg:LCBP}Loop Correcting algorithm}
\begin{tabular}{ll}
{\bf Input}:  & initial approximate cavity distributions $\{\Zmi{i}\}_{i\in\vars}$\\
{\bf Output}: & improved approximate cavity distributions $\{\Zma{i}\}_{i\in\vars}$
\end{tabular}
\medskip
\begin{algorithmic} [1]
\REPEAT
\FORALL{$i\in\vars$}
\FORALL{$Y\in \nbv{i}$}
\STATE $\erx{i}{Y} \leftarrow \frac{\displaystyle \left(\prod_{j\in \nbfe{Y}{i}} \sum_{x_i} \sum_{\xDe{j}{Y}} \FacNe{j}{Y} \Zmi{j} \prod_{J \in \nbv{j}} \er{j}{J} \right)^{1/\nel{\nbfe{Y}{i}}}}{\displaystyle {\sum_{x_i}} \sum_{\xDe{i}{Y}} \FacNe{i}{Y} \Zmi{i} \prod_{I \in \nbve{i}{Y}} \er{i}{I}}$\label{eq:LCBP_update}
\ENDFOR
\ENDFOR
\UNTIL convergence
\FORALL{$i\in\vars$}
\STATE $\Zmax{i} \leftarrow \Zmix{i} \prod_{I\in\nbv{i}} \erx{i}{I}$
\ENDFOR
\end{algorithmic}
\end{algorithm}

Alternatively, one can formulate Algorithm \ref{alg:LCBP} in terms of the ``beliefs''
  \begin{equation}\label{eq:pancakes}
  \Qx{i} \propto \FacNx{i} \Zmix{i} \prod_{I \in \nbv{i}} \erx{i}{I} = \FacNx{i} \Zmax{i}.
  \end{equation}
As one easily verifies, the following update equation
  \begin{equation*}
  \Q{i} \leftarrow \Q{i} \, \frac{\displaystyle \prod_{j \in \nbfe{Y}{i}}\left( \sum_{x_{\Dele{j}{(\nbfe{Y}{i})}}} \Q{j}\,\fac{Y}^{-1} \right)^{1/\nel{\nbfe{Y}{i}}}}{\displaystyle \sum_{x_{\Dele{i}{(\nbfe{Y}{i})}}} \Q{i} \, \fac{Y}^{-1}}
  \end{equation*}
is equivalent to line \ref{eq:LCBP_update} of Algorithm \ref{alg:LCBP}.
Intuitively, the update improves the approximate distribution $\Q{i}$ on 
$\Del{i}$ by replacing its marginal on $\nbfe{Y}{i}$ (in the absence of $Y$)
by a more accurate approximation of this marginal, namely the numerator.
Written in this form, the algorithm is reminiscent of Iterative Proportional
Fitting (IPF). However, contrary to IPF, the desired marginals are also updated
each iteration.
Note that after convergence, the large beliefs $\Qx{i}$ need not be consistent,
i.e.\ in general $\sum_{\xDe{i}{J}} \Q{i} \neq \sum_{\xDe{j}{J}} \Q{j}$ for $i, j \in \vars$,
$J \subseteq \Del{i} \cap \Del{j}$.

\subsection{A special case: factorized cavity distributions}

In the previous subsection we have discussed how to improve approximations of
cavity distributions. We now discuss what happens when we use the simplest
possible initial approximations $\{\Zmi{i}\}_{i\in\vars}$, namely constant
functions, in Algorithm \ref{alg:LCBP}. This amounts to the assumption that no
loops are present.  We will show that if the factor graph does not contain
short loops consisting of four nodes, fixed points of the standard BP algorithm
are also fixed points of Algorithm \ref{alg:LCBP}. In this sense, Algorithm
\ref{alg:LCBP} can be considered to be a generalization of the BP algorithm. In
fact, this holds even if the initial approximations factorize in a certain way, 
as we will show below.

If all factors involve at most two variables, one can easily arrange for the
factor graph to have no loops of four nodes. See figure
\ref{fig:cavities}(a) for an example of a factor graph which has no
loops of four nodes. The factor graph depicted in Figure \ref{fig:Y_min_i}
does have a loop of four nodes: $k-Y-j-J_2-k$.

\begin{theo} If the factor graph corresponding to \eref{eq:probability_distribution}
has no loops of exactly four nodes, and all initial approximate cavity distributions factorize 
in the following way:
  \begin{equation}\label{eq:factorized_Zmix}
  \Zmix{i} = \prod_{I\in\nbv{i}} \xi^{\setm i}_I(\xnfe{I}{i}) \qquad \forall i\in\vars,
  \end{equation}
then fixed points of the BP algorithm can be mapped to fixed points of Algorithm 
\ref{alg:LCBP}. Furthermore, the corresponding variable and factor marginals obtained from
\eref{eq:pancakes} are identical to the BP beliefs.
\end{theo}

\begin{proof}
Note that replacing the initial cavity approximations by 
  \begin{equation}\label{eq:invariance}
  \Zmix{i} \mapsto \Zmix{i} \prod_{I\in\nbv{i}} \epsilon^{\setm i}_I(\xnfe{I}{i})
  \end{equation}
for arbitrary positive functions $\epsilon^{\setm i}_I(\xnfe{I}{i})$ does not 
change the beliefs \eref{eq:pancakes} corresponding to the fixed points of \eref{eq:LCBP}.
Thus, without loss of generality, we can assume $\Zmix{i} = 1$ for all $i \in \vars$. 
The BP update equations are:
\begin{equation}\label{eq:BP_updates}
  \begin{array}{l@{{}\propto{}}ll}
  \mu_{j\to I}(x_j) & \displaystyle \prod_{J \in \nbve{j}{I}} \mu_{J\to j}(x_j) & \qquad j \in \vars, I \in \nbv{j} \\
  \mu_{I\to i}(x_i) & \displaystyle \sum_{\xnfe{I}{i}} \facx{I} \prod_{j\in \nbfe{I}{i}} \mu_{j\to I}(x_j) & \qquad I \in \facs, i \in \nbf{I}
  \end{array}
\end{equation}
in terms of messages $\{\mu_{J\to j}(x_j)\}_{j\in\vars, J\in\nbv{j}}$ and $\{\mu_{j\to J}(x_j)\}_{j\in\vars, J\in\nbv{j}}$.
Assume that the messages $\mu$ are a fixed point of \eref{eq:BP_updates} and take the \emph{Ansatz}
  \begin{equation*}
  \erx{i}{I} = \prod_{k \in \nbfe{I}{i}} \mu_{k \to I}(x_k) \qquad \text{for $i \in \vars, I \in \nbv{i}$}.
  \end{equation*}
Then, for $i \in \vars$, $Y \in \nbv{i}$, $j \in \nbfe{Y}{i}$, we can write out part of
the numerator of \eref{eq:LCBP} as follows:
  \begin{equation*}\begin{split}
  \sum_{x_i}\sum_{\xDe{j}{Y}} \FacNe{j}{Y} \Zmi{j} \prod_{J \in \nbv{j}} \er{j}{J}
  & = \sum_{x_i}\sum_{\xDe{j}{Y}} \er{j}{Y} \prod_{J\in\nbve{j}{Y}} \fac{J} \er{j}{J} \\
  & = \sum_{x_i}\left(\prod_{k\in\nbfe{Y}{j}} \mu_{k\to Y}\right) \prod_{J\in\nbve{j}{Y}} \sum_{\xnfe{J}{j}} \fac{J} \prod_{k\in\nbfe{J}{j}} \mu_{k\to J} \\
  & = \sum_{x_i}\left(\prod_{k\in\nbfe{Y}{j}} \mu_{k\to Y}\right) \mu_{j\to Y} 
    = \sum_{x_i}\prod_{k\in\nbf{Y}} \mu_{k\to Y} 
    \propto \prod_{k\in\nbfe{Y}{i}} \mu_{k\to Y} \\
  & = \er{i}{Y},
  \end{split}\end{equation*}
where we used the BP update equations \eref{eq:BP_updates} and rearranged the 
summations and products using the assumption that the factor graph has no loops of
four nodes.
Thus, the numerator of the r.h.s.\ of \eref{eq:LCBP} is simply $\er{i}{Y}$.
Using a similar calculation, one can derive that the denominator of the r.h.s.\ of 
\eref{eq:LCBP} is constant, and hence equation 
\eref{eq:LCBP} is valid (up to an irrelevant constant).

For $Y \in \facs$, $i \in \nbf{Y}$, the marginal on $x_Y$ of the belief
\eref{eq:pancakes} can be written in a similar way:
  \begin{equation*}\begin{split}
  \sum_{\xDe{i}{Y}} \Q{i}
  & \propto \sum_{\xDe{i}{Y}} \FacN{i} \prod_{I \in \nbv{i}} \er{i}{I}
    = \sum_{\xDe{i}{Y}} \prod_{I\in\nbv{i}} \fac{I} \prod_{k\in\nbfe{I}{i}} \mu_{k\to I} \\
  & = \fac{Y} \left(\prod_{k\in\nbfe{Y}{i}} \mu_{k\to Y}\right) \prod_{I\in\nbve{i}{Y}} \sum_{\xnfe{I}{i}} \fac{I} \prod_{k\in\nbfe{I}{i}} \mu_{k\to I} \\
  & = \fac{Y} \left(\prod_{k\in\nbfe{Y}{i}} \mu_{k\to Y}\right) \prod_{I\in\nbve{i}{Y}} \mu_{I\to i}
    = \fac{Y} \left(\prod_{k\in\nbfe{Y}{i}} \mu_{k\to Y}\right) \mu_{i\to Y} \\
  & = \fac{Y} \prod_{k\in\nbf{Y}} \mu_{k\to Y}.
  \end{split}\end{equation*}
which is proportional to the BP belief $b_Y(x_Y)$ on $x_Y$. Hence, also the single variable marginal
$b_i$ defined in \eref{eq:single_var_beliefs} corresponds to the BP single variable
belief, since both are marginalizations of $b_Y$ for $Y \in \nbv{i}$.
\end{proof}

If the factor graph does contain loops of four nodes, we find empirically that
the fixed point of Algorithm \ref{alg:LCBP}, when using factorized initial cavity
approximations as in \eref{eq:factorized_Zmix}, corresponds to the ``minimal'' 
CVM approximation, i.e.\ the CVM
approximation that uses all (maximal) factors as outer clusters
\citep{Kikuchi51,Pelizzola05}.\footnote{Provided that the factor graph is
connected.} In that case, the factor beliefs found by Algorithm \ref{alg:LCBP}
are consistent, i.e.  $\sum_{\xDe{i}{\nbf{Y}}} \Q{i} = \sum_{\xDe{j}{\nbf{Y}}}
\Q{j}$ for $i, j \in \nbf{Y}$, and are identical to the minimal CVM factor
beliefs. Thus it appears that Algorithm \ref{alg:LCBP} can be considered as a
generalization of the minimal CVM approximation (which can e.g.\ be calculated
using the GBP algorithm \citep{YedidiaFreemanWeiss05} or a double-loop
implementation \citep{HeskesAlbersKappen03}).

We have not yet been able to prove this, so currently this claim stands as a
conjecture, which we have empirically verified to be true for all the graphical
models used for numerical experiments in section \ref{sec:experiments}.  Note
that in case the factor graph has no loops of length four, the minimal CVM
approximation reduces to the Bethe approximation, which yields a proof
for this case. The proof in the general case is expected to be more involved,
since one needs to keep track of various overlapping sets and it requires a
translation of the structure of \eref{eq:LCBP} (where the basic sets of
variables involved are of three types, namely $i$, $Y\setm i$, and $\Del{i}$)
and the GBP equations or Lagrange multiplier equations corresponding to the
minimal CVM approximation (where the basic variable sets are those that can be
written as an intersection of a finite number of factors).

\subsection{Obtaining initial approximate cavity distributions}\label{sec:initial_cavities_full}

There is no principled way to obtain the initial cavity approximations
$\Zmix{i}$. In the previous subsection, we saw that factorized cavity
approximations result in the minimal CVM approximation, which does not yet take
into account the effect of loops in the cavity network. More sophisticated
approximations that do take into account the effect of loops can significantly
enhance the accuracy of the final result. In principle, there are many
possibilities to obtain the initial cavity approximations. Here, we will
describe one method, which uses BP on clamped cavity networks. This method
captures all interactions in the cavity distribution of $i$ in an approximate
way and can lead to very accurate results. Instead of BP, any other approximate
inference method that gives an approximation of the normalizing constant $Z$ in
\eref{eq:probability_distribution} can be used, such as Mean Field, TreeEP
\citep{MinkaQi04}, a double-loop version of BP \citep{HeskesAlbersKappen03}
which has guaranteed convergence towards a minimum of the Bethe free energy, or
some variant of GBP \citep{YedidiaFreemanWeiss05}.  One could also choose the
method for each cavity separately, trading accuracy versus computation time.
We focus here on BP because it is a very fast and often relatively
accurate algorithm. 

Let $i \in \vars$ and consider the cavity network of $i$. For each possible
state of $\xd{i}$, run BP on the cavity network clamped to that state $\xd{i}$
and calculate the corresponding Bethe free energy $F^{\setm i}_{Bethe}(\xd{i})$
\citep{YedidiaFreemanWeiss05}. Take as initial approximate cavity distribution: 
  \begin{equation*} 
  \Zmix{i} \propto e^{-F^{\setm i}_{Bethe}(\xd{i})}.
  \end{equation*} 
This procedure is
exponential in the size of $\del{i}$: it uses $\prod_{j\in\del{i}} \nel{\X{j}}$
BP runs. However, many networks encountered in applications are relatively
sparse and have limited cavity size and the computational cost may be acceptable.

This particular way of obtaining initial cavity distributions has the following
interesting property: in case the factor graph contains only a single loop, the
final beliefs \eref{eq:pancakes} resulting from Algorithm \ref{alg:LCBP} are
exact. This can be shown using an argument similar to that given in
\citep{MontanariRizzo05}. Suppose that the graphical model contains exactly one
loop and let $i \in \vars$. Consider first the case that $i$ is part of the
loop; removing $i$ will break the loop and the remaining cavity network will be
singly connected.  The cavity distribution approximated by BP will thus be
exact. Now if $i$ is not part of the loop, removing $i$ will divide the network
into several connected components, one for each neighbour of $i$. This implies
that the cavity distribution calculated by BP contains no higher order
interactions, i.e.\ $\Zmi{i}$ is exact modulo single variable interactions.
Because the final beliefs \eref{eq:pancakes} are invariant under perturbation
of the $\Zmi{i}$ by single variable interactions, the final beliefs calculated
by Algorithm \ref{alg:LCBP} are exact.

If all interactions are pairwise and each variable is binary and has exactly
$\nel{\del{i}} = d$ neighbours, the time complexity of the resulting ``Loop
Corrected BP'' (LCBP) algorithm is given by $N 2^d T_{BP} + N d 2^{d+1}
N_{LC}$, where $T_{BP}$ is the average time of one BP run on a clamped cavity
network and $N_{LC}$ is the number of iterations needed to obtain convergence
in Algorithm \ref{alg:LCBP}.

\subsection{Differences with \cite{MontanariRizzo05}}\label{sec:diffMontanariRizzo}

As mentioned before, the idea of estinating the cavity distributions and
imposing certain consistency relations amongst them has been first presented in
\cite{MontanariRizzo05}. In its simplest form (i.e.\ the so-called first order
correction), the implementation of that basic idea as proposed by
\cite{MontanariRizzo05} differs from our proposed implementation in the
following aspects.

First, the original method described by \cite{MontanariRizzo05} is only
formulated for the rather special case of binary variables and pairwise
interactions. In contrast, our method is formulated in a general way that makes
it applicable to factor graphs with variables having more than two possible
values and factors consisting of more than two variables. Also, factors may
contain zeroes. The generality that our implementation offers is important for
many practical applications.\footnote{The method by \cite{MontanariRizzo05}
could probably be generalized in a way that stays closer to the original one
than our proposal, but it is not so obvious how to do this.} In the rest of this
section, we will assume that the graphical model
\eref{eq:probability_distribution} belongs to the special class of binary
variables with pairwise interactions, allowing further comparison of both
implementations.

An important difference is that \cite{MontanariRizzo05} suggest to 
deform the initial approximate cavity distributions by altering certain 
\emph{cumulants} (also called ``connected correlations''), instead of
altering certain interactions.
In general, for a set $\C{A}$ of $\pm1$-valued random variables 
$\{x_i\}_{i\in\C{A}}$, one can define for any subset $\C{B} \subseteq \C{A}$
the \emph{moment}
  \begin{equation*}
  M_{\C{B}} := \sum_{x_{\C{A}}} P(x_{\C{A}}) \prod_{j\in\C{B}} x_j.
  \end{equation*}
The moments $\{M_{\C{B}}\}_{\C{B} \subseteq \C{A}}$ are a parameterization of
the probability distribution $P(x_{\C{A}})$. An alternative parameterization is
given in terms of the cumulants. The \emph{(joint) cumulants} 
$\{C_{\C{E}}\}_{\C{E} \subseteq \C{A}}$ are certain polynomials of the moments,
defined implicitly by the following relations:
  \begin{equation*}
  M_{\C{B}} = \sum_{\C{C} \in \mathrm{Part}(\C{B})} \prod_{\C{E} \in \C{C}} C_{\C{E}}
  \end{equation*}
where $\mathrm{Part}(\C{B})$ is the set of partitions of $\C{B}$.\footnote{For a set $X$, 
a \emph{partition} of $X$ is a nonempty set $Y$ such that
each $Z \in Y$ is a nonempty subset of $X$ and $\bigcup Y = X$.} In particular,
$C_i = M_i$ and $C_{ij} = M_{ij} - M_i M_j$ for all $i,j \in \C{A}$ with $i\ne j$.
\cite{MontanariRizzo05} propose to approximate the cavity distributions by
estimating the pair cumulants and assuming higher order cumulants to be zero.
Then, the singleton cumulants (i.e.\ the single node marginals) are altered,
keeping higher order cumulants fixed,
in such a way as to impose consistency of the single node marginals, in the
absence of interactions shared by two neighbouring cavities.
We refer the reader to the appendix for a more detailed description of the
implementation in terms of cumulants suggested by \cite{MontanariRizzo05}. 

A minor difference lies in the method to obtain initial approximations to the
cavity distributions. \cite{MontanariRizzo05} propose to use BP in combination with
linear response theory to obtain the initial pairwise cumulants. This
difference is not very important, since one could also use BP on clamped cavity
networks instead, which turns out to give almost identical results. 

As we will show in section \ref{sec:experiments}, our method of altering
interactions appears to be more robust and still works in regimes with strong
interactions, whereas the cumulant implementation suffers from convergence
problems for strong interactions.

An advantage of the cumulant based scheme is that it allows for a linearized
version (by expanding up to first order in terms of the pairwise cumulants, see
appendix) which is quadratic in the size of the cavity. This means that this
linearized, cumulant based version is currently the only one that can be
applied to networks with large Markov blankets (cavities), i.e.\ where the
maximum number of states $\max_{i\in\vars} \nel{\X{\Del{i}}}$ is large
(provided that all variables are binary and interactions are pairwise).

\section{Numerical experiments}\label{sec:experiments}

We have performed various numerical experiments to compare the quality of the results
and the computation time of the following approximate inference methods:
\begin{description}
\item[MF] Mean Field, with a random sequential update scheme and no damping.
\item[BP] Belief Propagation. We have used the recently proposed update scheme 
\cite{ElidanMcGrawKoller06}, which converges also for difficult problems 
without the need for damping.
\item[TreeEP] TreeEP \citep{MinkaQi04}, without damping.
We generalized the method of choosing the base tree described in \cite{MinkaQi04} 
to multiple variable factors as follows: when
estimating the mutual information between $x_i$ and $x_j$, we take the product
of the marginals on $\{i,j\}$ of all the factors that involve $x_i$ and/or $x_j$.
Other generalizations of TreeEP to higher order factors
are possible (e.g.\ by clustering variables), but it is not clear how to do
this in general in an optimal way.
\item[LCBP] (``Loop Corrected Belief Propagation'') Algorithm \ref{alg:LCBP}, 
where the approximate cavities are initialized according to the description
in section \ref{sec:initial_cavities_full}.
\item[LCBP-Cum] The original cumulant based loop correction scheme by
\cite{MontanariRizzo05}, using response propagation (also known as linear response;
see \citep{WellingTeh04}) to approximate the initial pairwise cavity
cumulants. The full update equations \eref{eq:MR_full} are used and higher
order cumulants are assumed to vanish.
\item[LCBP-Cum-Lin] Similar to LCBP-Cum, but instead of the full update equations
\eref{eq:MR_full}, the linearized update equations \eref{eq:MR_linear} are used.
\item[CVM-Min] A double-loop implementation \citep{HeskesAlbersKappen03} 
of the minimal CVM approximation, which uses (maximal) factors as outer clusters. 
\item[CVM-$\Delta$] A double-loop implementation 
of CVM using the sets $\{\Del{i}\}_{i\in\vars}$ as outer clusters. These are the
same sets of variables as used by LCBP (c.f.\ \eref{eq:pancakes}) and therefore
it is interesting to compare both algorithms.
\item[CVM-Loop$k$] A double-loop implementation of CVM, using as outer clusters
all (maximal) factors together with all loops in the factor graph that consist of up 
to $k$ different variables (for $k=3,4,5,6,8$).
\end{description}

We have used a double-loop implementation of CVM instead of GBP because the
former is guaranteed to converge to a local minimum of the Kikuchi free energy
\citep{HeskesAlbersKappen03} without damping, whereas the latter often only
converges with strong damping. The difficulty with damping is that the optimal
damping constant is not known \emph{a priori}, which necessitates multiple trial
runs with different damping constants, until a suitable one is found. Using
too much damping slows down convergence, whereas a certain amount of damping is
required to obtain convergence in the first place. Therefore, in general we
expect that GBP is not much faster than a double-loop implementation because of 
the computational cost of finding the optimal damping constant.

To be able to assess the errors of the various approximate methods, we have 
only considered problems for which exact inference (using a standard JunctionTree
method) was still feasible.

For each approximate inference method, we report the maximum $\ell_\infty$
error of the approximate single node marginals $b_i$, calculated as follows:
  \begin{equation}\label{eq:max_err}
  \text{Error} := \max_{i\in\vars} \max_{x_i\in\X{i}} \abs{b_i(x_i) - P(x_i)}
  \end{equation}
where $P(x_i)$ is the exact marginal calculated using the JunctionTree method.

The computation time was measured as CPU time in seconds on a 2.4 GHz AMD
Opteron 64bits processor with 4 GB memory. The timings should be seen as
indicative because we have not spent equal amounts of effort optimizing each
method.\footnote{Our C++ implementation of various approximate inference
algorithms is free/open source software and can be downloaded from 
\url{http://www.mbfys.ru.nl/~jorism/libDAI}}

We consider an iterative method to be ``converged'' after $T$ timesteps if for
each variable $i \in \vars$, the $\ell_\infty$ distance between the approximate
probability distributions of that variable at timestep $T$ and $T+1$ is less
than $\epsilon = 10^{-9}$.

We have studied four different model classes: (i) random graphs of uniform
degree with pairwise interactions and binary variables; (ii) random factor
graphs with binary variables and factor nodes of uniform degree $k = 3$; 
(iii) the ALARM network, which has variables taking on more than two possible
values and factors consisting of more than two variables; (iv) PROMEDAS networks,
which have binary variables but factors consisting of more than two variables.

\subsection{Random regular graphs with binary variables}

We have compared various approximate inference methods on random graphs,
consisting of $N$ binary ($\pm1$-valued) variables, having only pairwise
interactions, where each variable has the same degree $\nel{\del{i}} = d$.
In this case, the probability distribution \eref{eq:probability_distribution} can 
be written in the following way:
  \begin{equation*}
  P(x) = \frac{1}{Z} \exp\left( \sum_{i\in\vars} \theta_i x_i + \frac{1}{2} \sum_{i\in\vars}\sum_{j\in\del{i}} J_{ij} x_i x_j \right),
  \end{equation*}
where the parameters $\{\theta_i\}_{i\in\vars}$ are called the \emph{local fields}
and the parameters $\{J_{ij} = J_{ji}\}_{i\in\vars, j\in\del{i}}$ are called the
\emph{couplings}. The graph structure and the parameters $\theta$ and $J$ were drawn 
randomly for each instance. 

The local fields $\{\theta_i\}$ were drawn independently from a 
$\C{N}(0,\beta \Theta)$ distribution (i.e.\ a normal distribution with mean 0 and 
standard deviation $\beta \Theta$). For the couplings $\{J_{ij}\}$, we 
distinguished two different
cases: mixed (``spin-glass'') and attractive (``ferromagnetic'') couplings. The
couplings were drawn independently from the following distributions:
  \begin{align*}
  J_{ij} & \sim \C{N}\left(0,\beta\, \arctanh \frac{1}{\sqrt{d-1}}\right) & \qquad & \text{mixed couplings}\\
  J_{ij} & = \abs{J_{ij}'}, \quad J_{ij}' \sim \C{N}\left(0, \beta\,\arctanh \frac{1}{d-1}\right) & \qquad & \text{attractive couplings}
  \end{align*}
The constant $\beta$ (called ``inverse temperature'' in statistical physics)
controls the overall interaction strength and thereby the difficulty of the
inference problem, larger $\beta$ corresponding usually to more difficult
problems. The constant $\Theta$ controls the relative strength of the local
fields, where larger $\Theta$ result in easier inference problems. The
particular $d$-dependent scaling of the couplings is used in order to obtain
roughly $d$-independent behaviour. In case of mixed couplings, for $\Theta = 0$
and for $\beta \approx 1$ a phase transition occurs in the limit of
$N\to\infty$, going from an easy ``paramagnetic'' phase for $\beta<1$ to a
complicated ``spin-glass'' phase for $\beta > 1$.  In the case of attractive
couplings and $\Theta = 0$, a phase transition also occurs at $\beta = 1$, now
going from the easy ``paramagnetic'' phase for $\beta < 1$ to a 
``ferromagnetic'' phase for $\beta > 1$.\footnote{More precisely,
in case of zero local fields ($\Theta=0$), the PA-SG phase transition occurs at
$(d-1) = \avg{\tanh^2 (\beta J_{ij})}$, where $\avg{\cdot}$ is the average over
all $J_{ij}$, and the PA-FE phase transition occurs at $(d-1) = \avg{\tanh
(\beta J_{ij})}$ \citep{MooijKappen05c}. What happens for $\Theta > 0$ is
not known, to the best of our knowledge.}

\subsubsection{$N = 100$, $d = 3$, mixed couplings, strong local fields ($\Theta=2$)}

\begin{figure}[p]
\centering
\includegraphics[scale=0.666666]{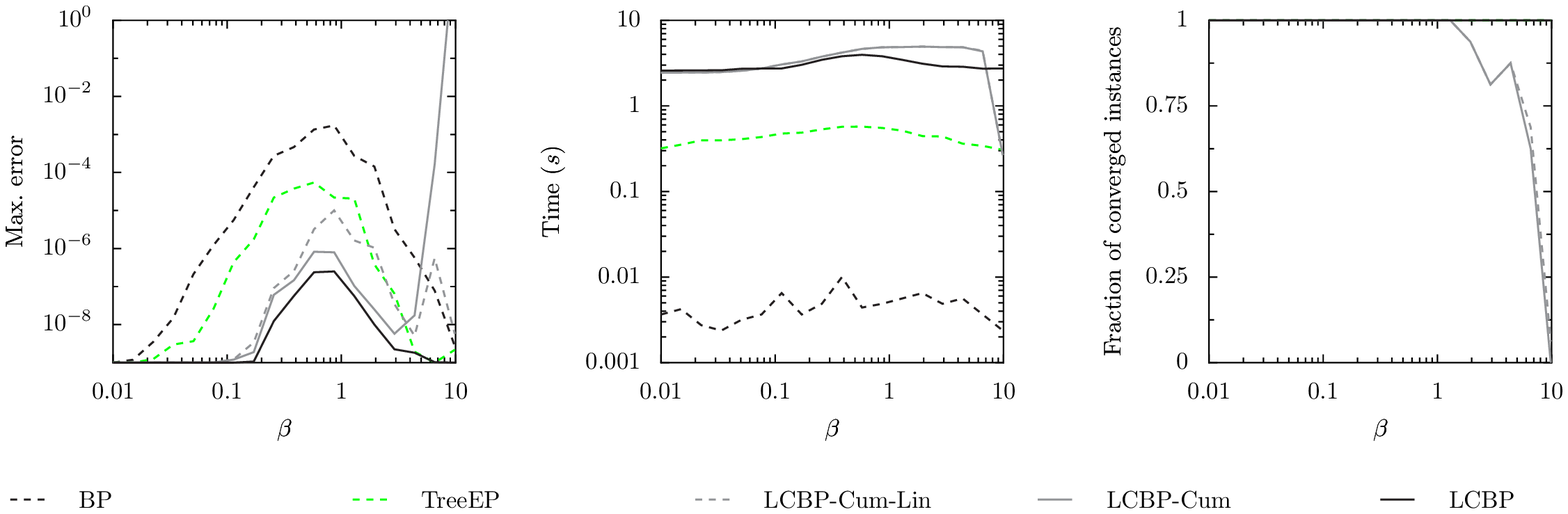}\\[1cm]
\begin{tabular}{ccc}
  \includegraphics[width=0.3\textwidth]{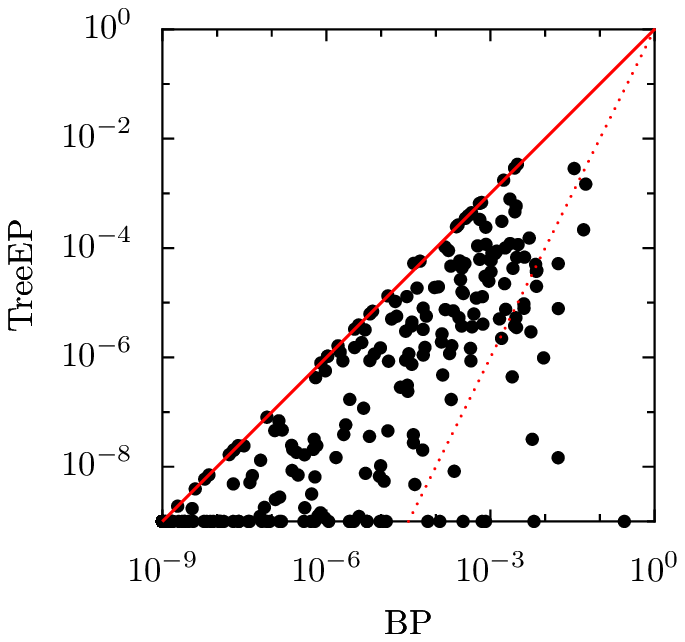}
& \includegraphics[width=0.3\textwidth]{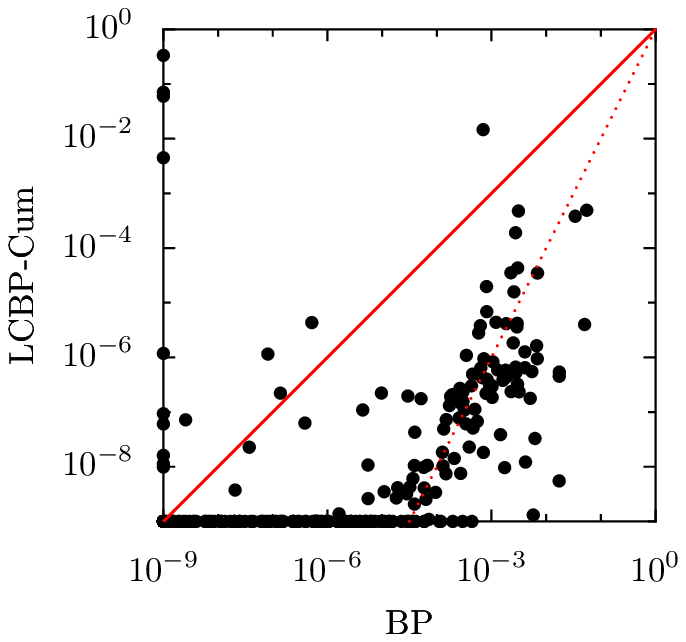}
& \includegraphics[width=0.3\textwidth]{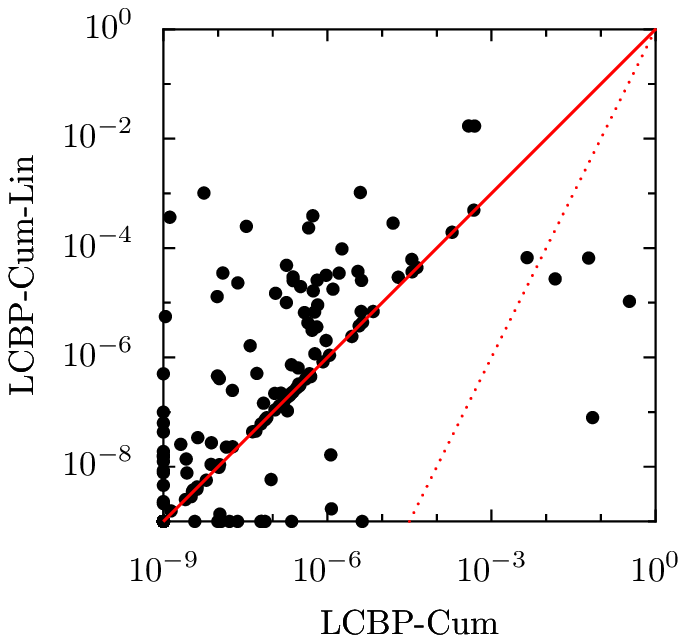}\\[0.5cm]
  \includegraphics[width=0.3\textwidth]{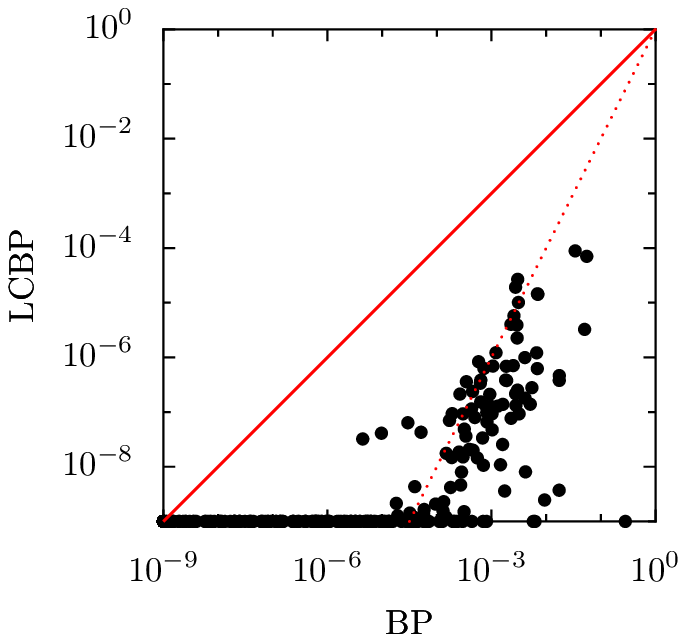}
& \includegraphics[width=0.3\textwidth]{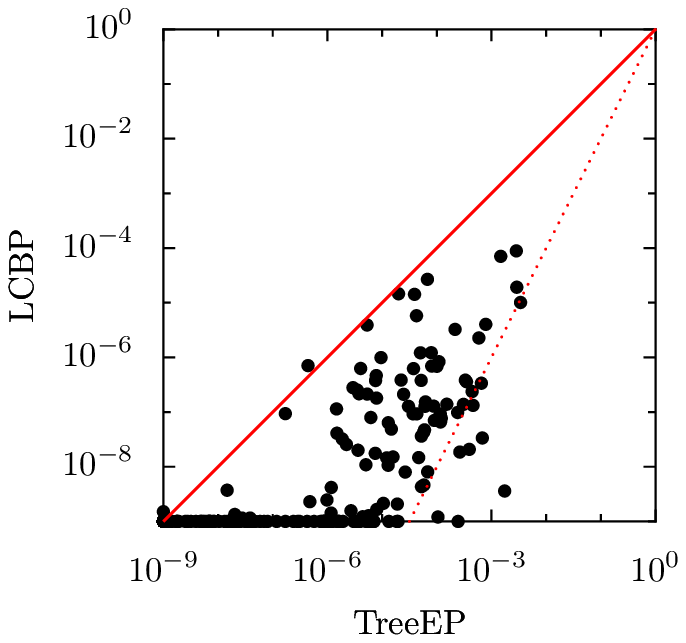}
& \includegraphics[width=0.3\textwidth]{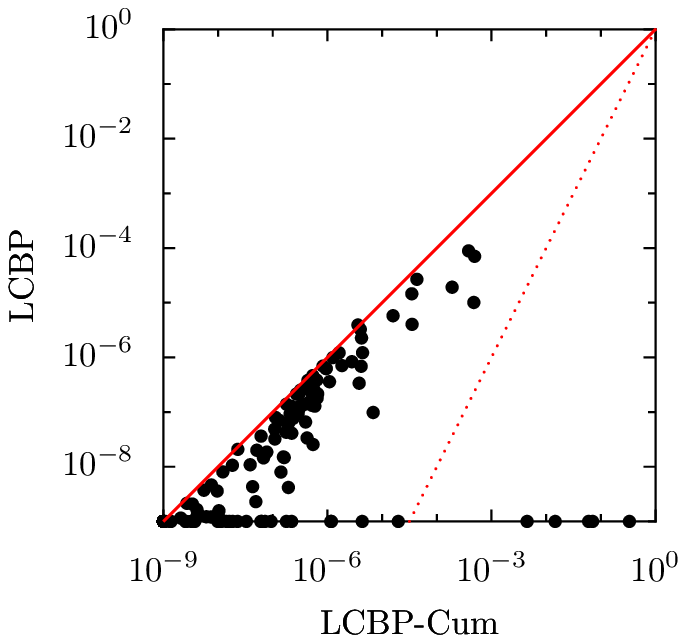}
\end{tabular}
\caption{\label{fig:dreg_d3_N100_sg}
Results for $(N=100,d=3)$ regular random graphs with mixed couplings and 
strong local fields $\Theta=2$. First row, from left to right: error, 
computation time and fraction of converged instances, as a function of 
$\beta$ for various methods, averaged over 16 randomly generated 
instances (where results are only included if the method has converged).
For the same instances, scatter plots of errors are shown in the 
next rows for various pairs of methods. The solid red lines correspond with 
$y=x$, the dotted red lines with $y=x^2$. Only points have been plotted
for which both approximate inference methods converged.}
\end{figure}

\begin{figure}[p]
\centering
\includegraphics[scale=0.666666]{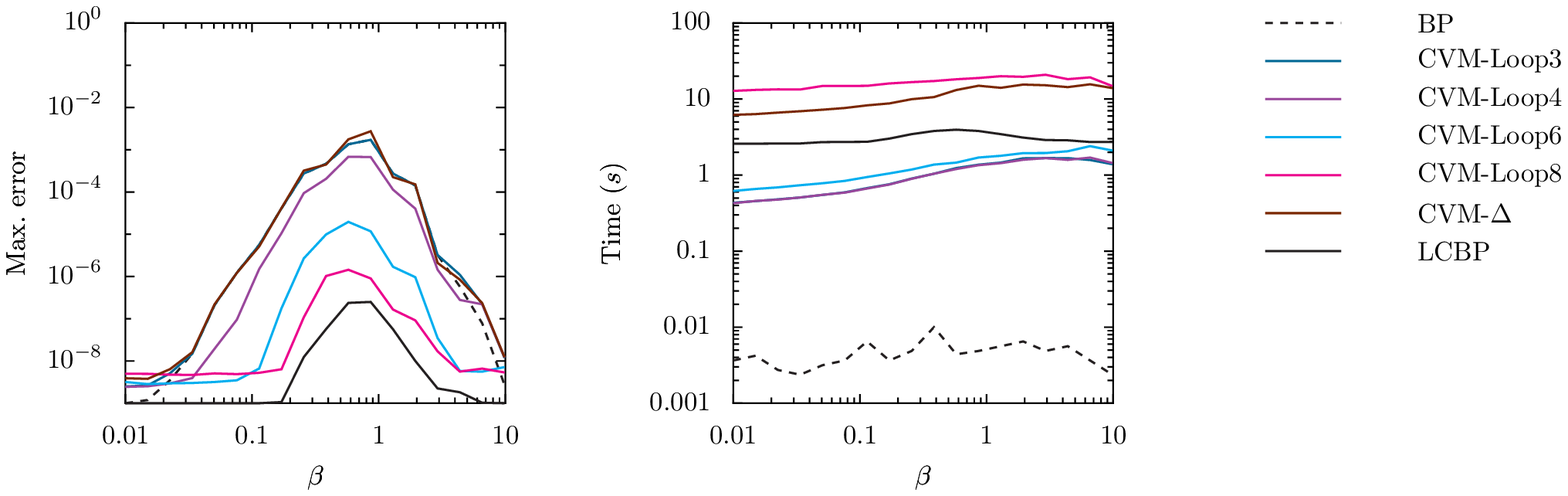}\\[1cm]
\begin{tabular}{ccc}
  \includegraphics[width=0.3\textwidth]{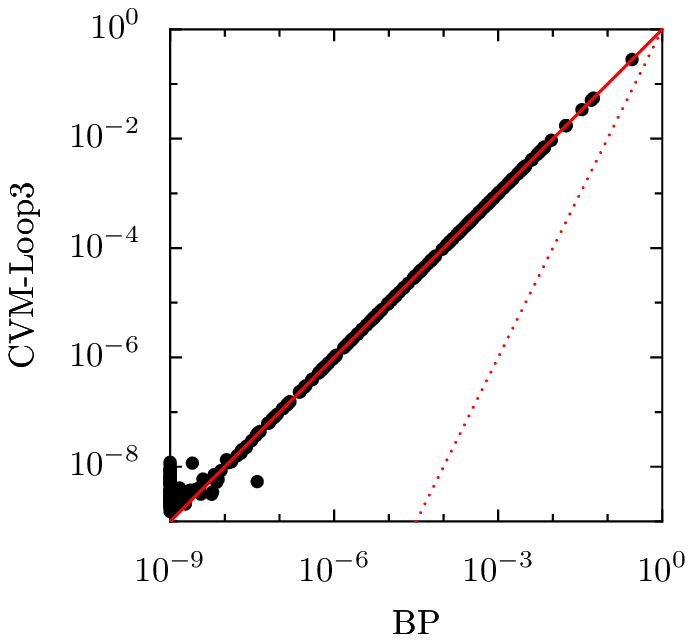}
& \includegraphics[width=0.3\textwidth]{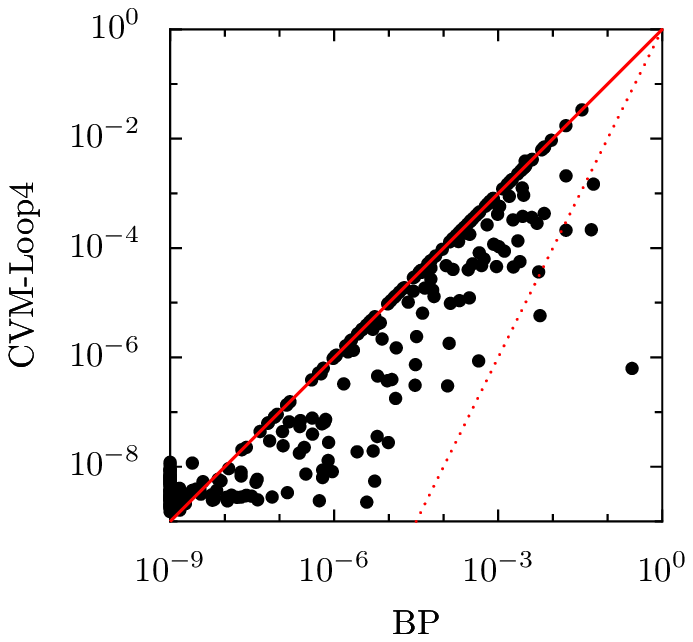}
& \includegraphics[width=0.3\textwidth]{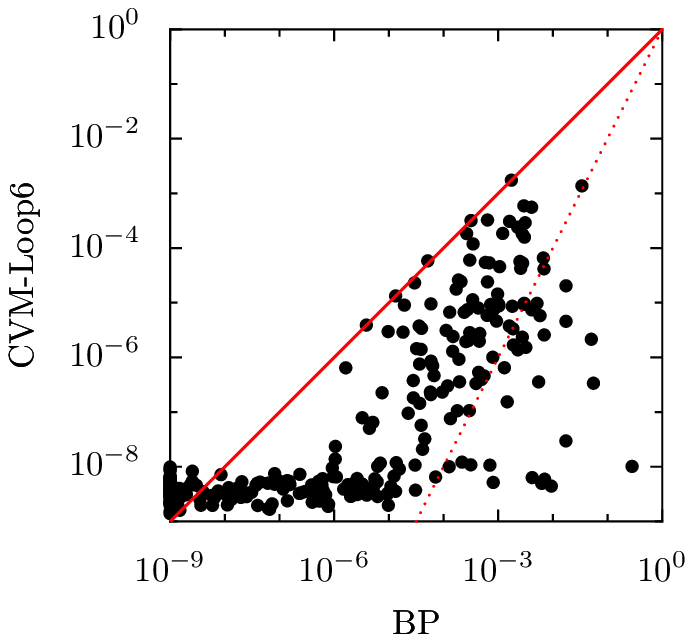}\\[0.5cm]
  \includegraphics[width=0.3\textwidth]{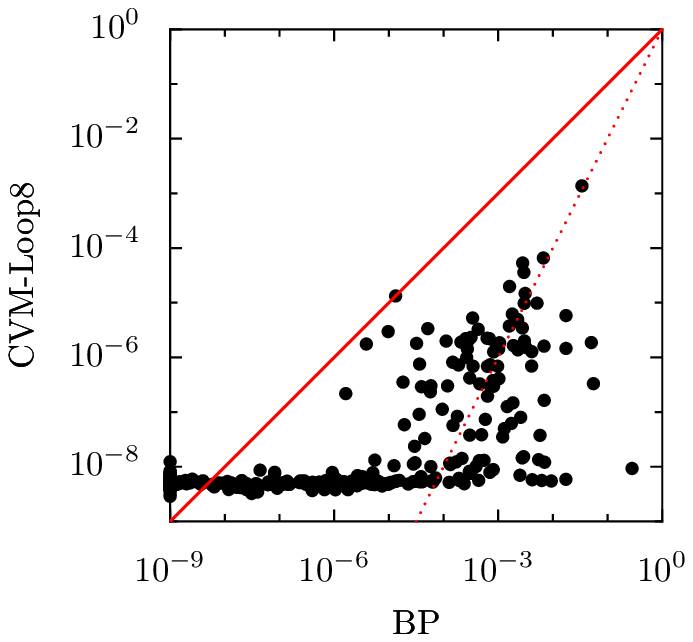}
& \includegraphics[width=0.3\textwidth]{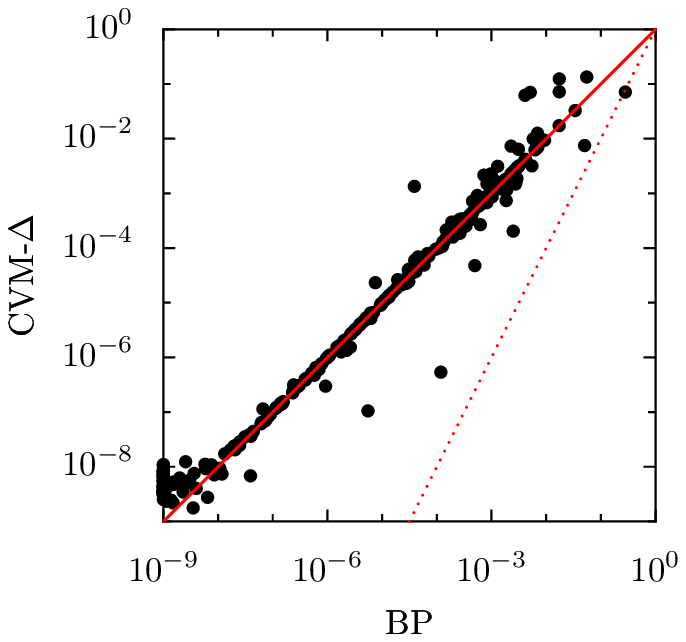}
& \includegraphics[width=0.3\textwidth]{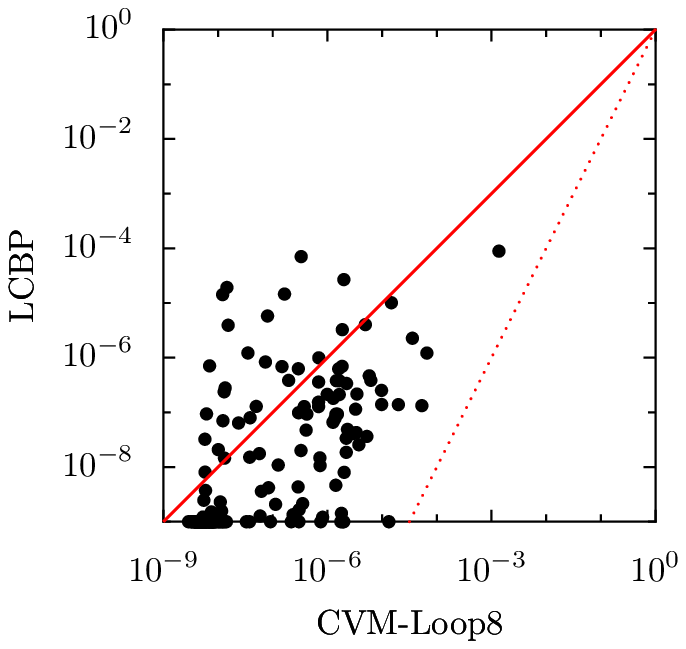}
\end{tabular}
\caption{\label{fig:dreg_d3_N100_sg_b}
Additional results for $(N=100,d=3)$ regular random graphs with mixed couplings and 
strong local fields $\Theta=2$, for the same instances as in Figure \ref{fig:dreg_d3_N100_sg}. 
All methods converged on all instances.}
\end{figure}

In this section we study regular random graphs of low degree $d = 3$,
consisting of $N = 100$ variables, with mixed couplings and relatively strong
local fields of strength $\Theta=2$.  We considered various overall interaction
strengths $\beta$ between $0.01$ and $10$. For each value of $\beta$, we used
16 random instances. On each instance, we ran various approximate inference
algorithms.  Figures \ref{fig:dreg_d3_N100_sg} and \ref{fig:dreg_d3_N100_sg_b}
show selected results.\footnote{We apologize to readers for the use of colours;
we saw no viable alternative for creating clear plots.}

Both figures consist of two parts; the first row in both figures shows averages
(in the logarithmic domain) of errors and computation time as a function of
$\beta$ for various methods. In addition, Figure \ref{fig:dreg_d3_N100_sg}
shows the fraction of instances on which each method converged; for Figure
\ref{fig:dreg_d3_N100_sg_b}, all methods converged for all values of $\beta$.
The averages of errors and computation time were calculated from the converged
instances only. The other rows in the figures contain scatter plots that
compare errors of various methods one-to-one. The solid red lines in the
scatter plots indicate equality; the dotted red lines indicate that the error
of the method on the vertical axis is the square of the error on the horizontal
axis. Saturation of errors around $10^{-9}$ is an artefact due to the convergence
criterion.  The CVM methods are often seen to saturate around $10^{-8}$, which
indicates that single iterations are less effective than for other methods.

We conclude from both figures that BP is the fastest but also the least
accurate method and that LCBP is the most accurate method and that it converges
for all $\beta$. Furthermore, the error of LCBP is approximately the square of
the BP error.

Figure \ref{fig:dreg_d3_N100_sg} shows further that TreeEP is able to obtain a
significant improvement over BP using little computation time.  For small
values of $\beta$, LCBP-Cum and LCBP-Cum-Lin both converge and yield high
quality results and the error introduced by the linearization is relatively
small. However, for larger values of $\beta$, both methods get more and more
convergence problems, although for the few cases where they do converge, they
still yield accurate results. At $\beta \approx 10$, both methods have
completely stopped converging.  The error introduced by the linearization
increases for larger values of $\beta$.  The computation times of LCBP-Cum,
LCBP-Cum-Lin and LCBP do not differ substantially in the regime where all
methods converge. The difference in quality between LCBP and LCBP-Cum is mainly
due to the fact that LCBP does take into account triple interactions in the
cavity (however, extending LCBP-Cum in order to take into account triple 
interactions is easy for this case of low $d$). 

The break-down of the cumulant based LCBP methods for high $\beta$ is probably
due to the choice of cumulants for parameterizing cavity distributions, which
seem to be less robust than interactions. Indeed, consider two random variables
$x_1$ and $x_2$ with fixed pair interaction $\exp (J x_1 x_2)$. By altering the
singleton interactions $\exp (\theta_1 x_1)$ and $\exp (\theta_2 x_2)$, one can
obtain any desired marginals of $x_1$ and $x_2$.  However, a fixed pair
cumulant $C_{12} = \Exp{x_1 x_2} - \Exp{x_1}\Exp{x_2}$ imposes a constraint on
the range of possible expectation values $\Exp{x_1}$ and $\Exp{x_2}$ (hence on
the single node marginals of $x_1$ and $x_2$); the freedom of choice in these
marginals becomes less as the pair cumulant becomes stronger.  We believe that
something similar happens for LCBP-Cum: for strong interactions, the
approximate pair cumulants in the cavity are strong, and even tiny errors can
lead to inconsistencies.\footnote{Indeed, for strong interactions, the update
equations \eref{eq:MR_full} often yield values for the $\Ma{i}{j}$ outside of
the valid interval $[-1,1]$. In this case, we project these values back into
the valid interval in the hope that the method will converge to a valid result,
which it sometimes does. This phenomenon also indicates the
lack of robustness of a cumulant parameterization in the regime of strong
interactions.}

The results of the CVM approach to loop correcting is shown in
\ref{fig:dreg_d3_N100_sg_b}. The CVM-Loop methods, with clusters reflecting the
short loops present in the factor graph, do improve on BP. The use of larger
clusters that subsume longer loops improves the results, but computation time
quickly increases. CVM-Loop3 does not obtain any improvement over BP, simply
because there are (almost) no loops of 3 variables present. The most accurate
CVM method, CVM-Loop8, needs more computation time than LCBP, whereas the quality
of its results is not as good.  Surprisingly, although CVM-$\Delta$ uses larger
cluster than BP, its quality is similar to that of BP and its computation time
is enormous. This is remarkable, since one would expect that CVM-$\Delta$
should improve on BP because it uses larger clusters. In any case, we conclude
that although LCBP and CVM-$\Delta$ use identical clusters, the nature of both
approximations is very different. 

\ifthenelse{\isundefined{\techrep}}{
We have also done experiments for weak local fields ($\Theta = 0.2$).
The behaviour is similar to that of strong local fields, apart from the
following differences. First, the influence of the phase transition is more
pronounced; many methods have severe convergence problems around $\beta = 1$.
Further, the negative effect of linearization on the error (LCBP-Cum-Lin 
compared to LCBP-Cum) is smaller.
}{
We have also done experiments for weak local fields ($\Theta = 0.2$).
The behaviour is similar to that of strong local fields, apart from the
following differences. First, the influence of the phase transition is more
pronounced; many methods have severe convergence problems around $\beta = 1$.
Further, the negative effect of linearization on the error (LCBP-Cum-Lin 
compared to LCBP-Cum) is smaller.

\subsubsection{Fixed $\beta$ and varying relative local field strength $\Theta$}

In addition, we have done experiments for fixed $\beta=1.0$ for various values 
of the relative local field strength $\Theta$ between 0.01 and 10. The results 
are shown in Figures \ref{fig:dreg_d3_N100_sg_theta} and \ref{fig:dreg_d3_N100_sg_theta_b}.

\begin{figure}[p]
\centering
\includegraphics[scale=0.666666]{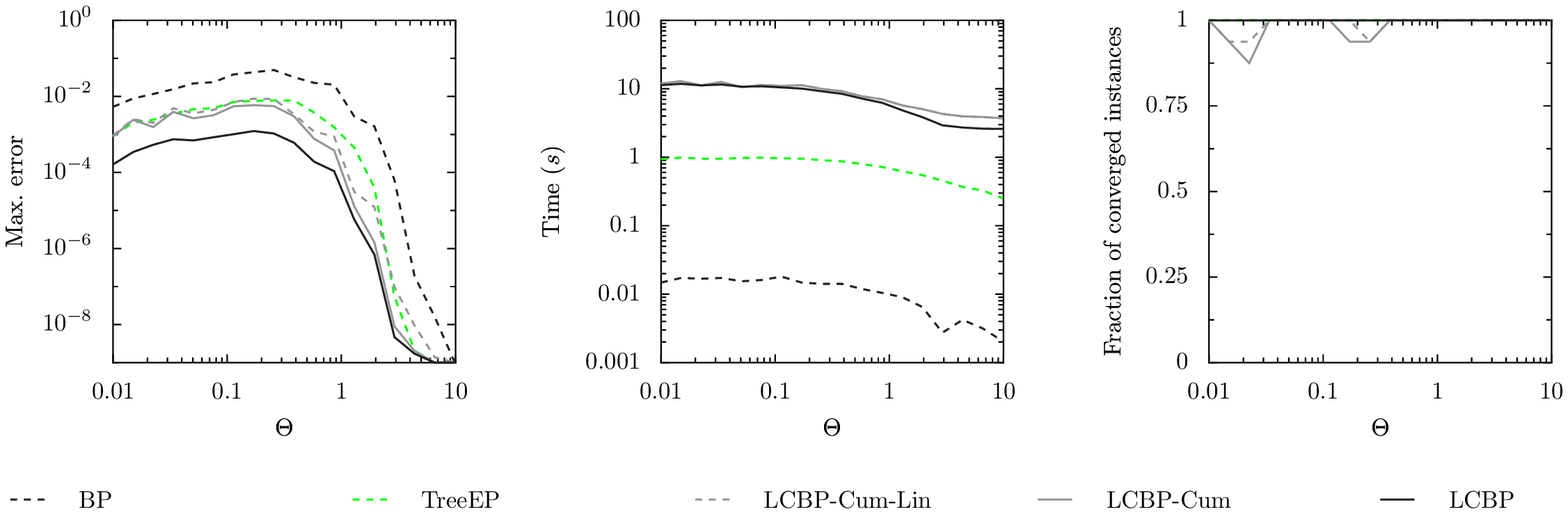}\\[1cm]
\begin{tabular}{ccc}
  \includegraphics[width=0.3\textwidth]{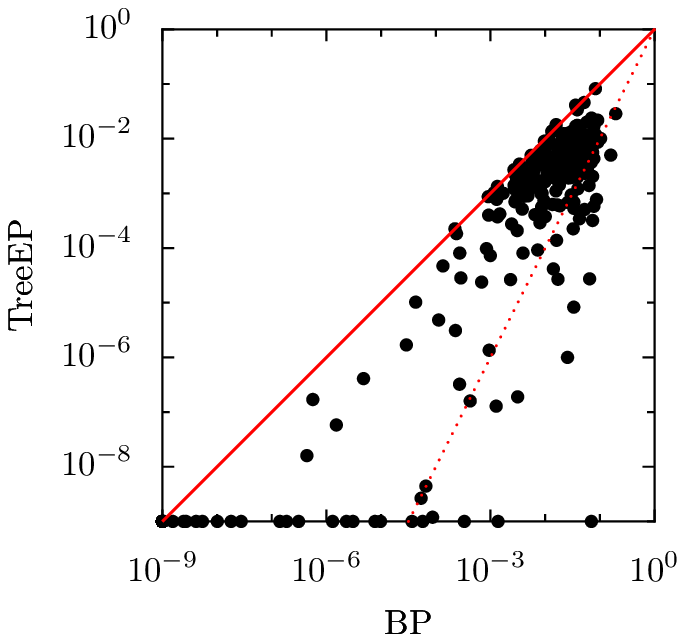}
& \includegraphics[width=0.3\textwidth]{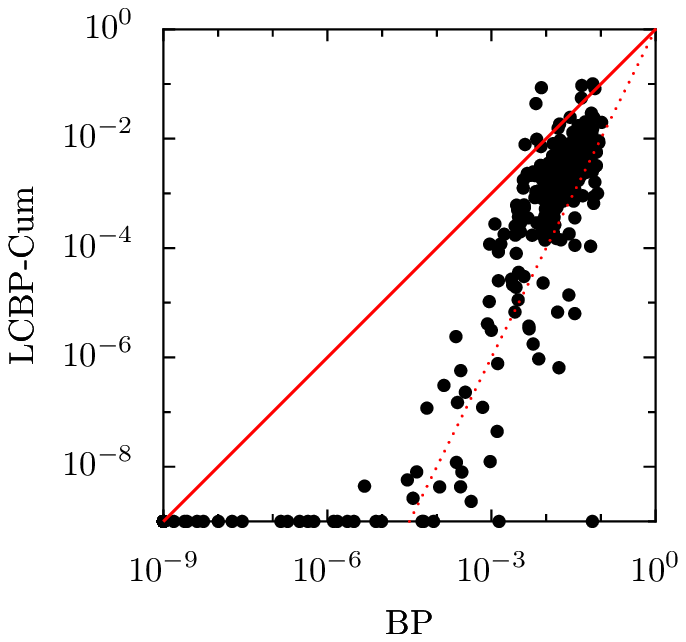}
& \includegraphics[width=0.3\textwidth]{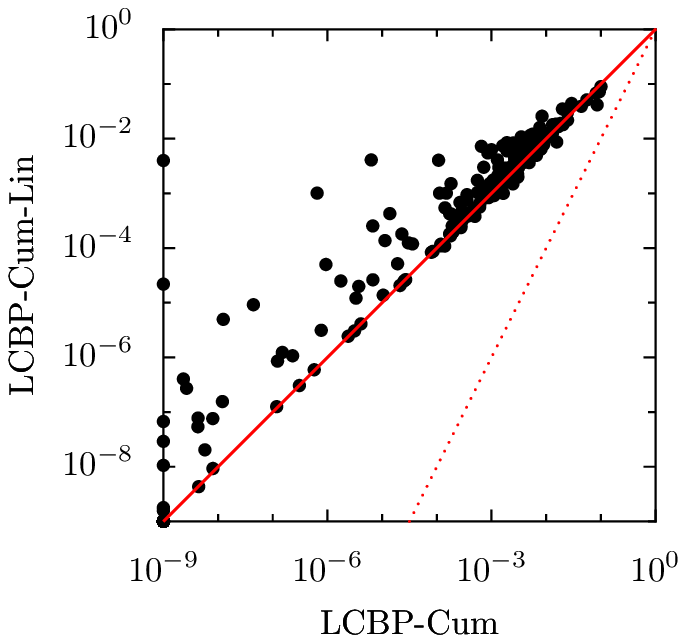}\\[0.5cm]
  \includegraphics[width=0.3\textwidth]{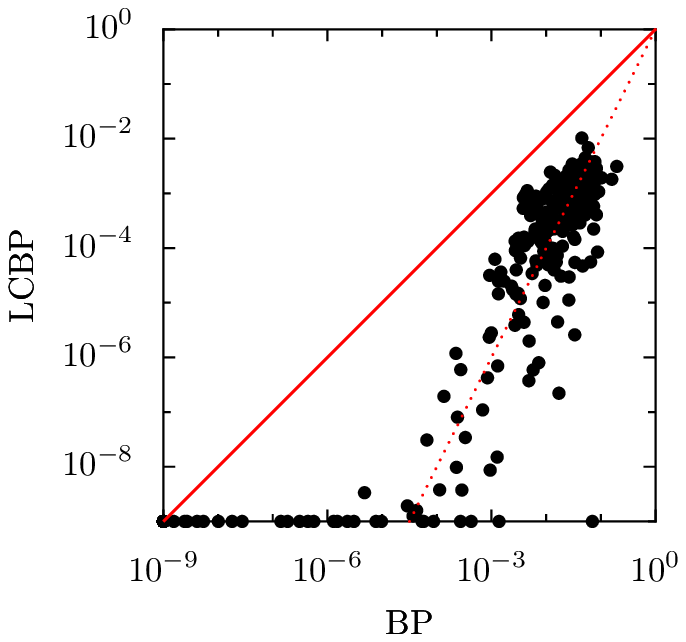}
& \includegraphics[width=0.3\textwidth]{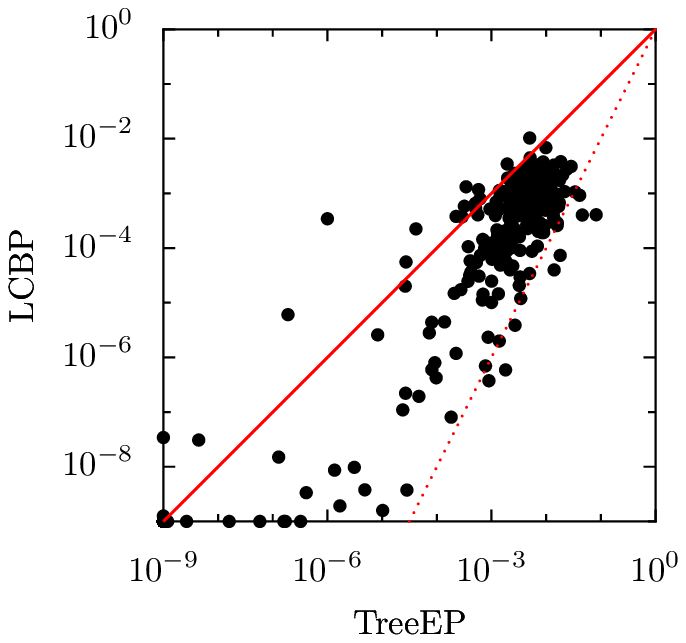}
& \includegraphics[width=0.3\textwidth]{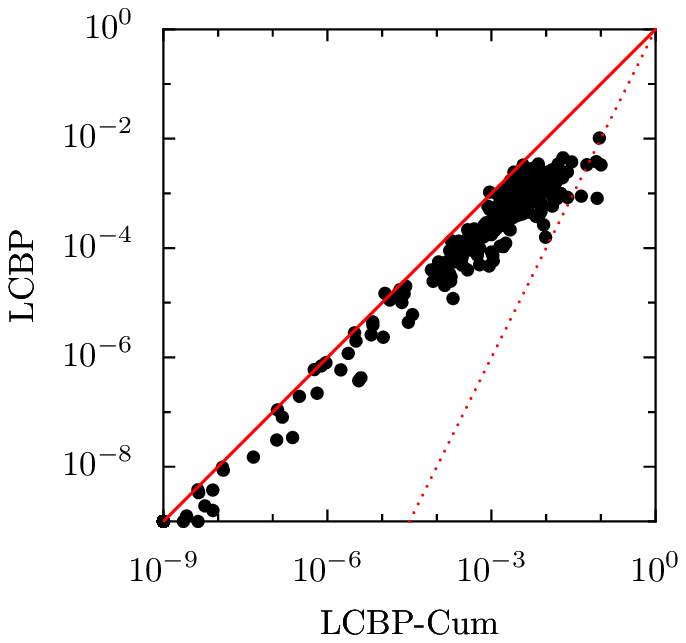}
\end{tabular}
\caption{\label{fig:dreg_d3_N100_sg_theta}
Results for $(N=100,d=3)$ regular random graphs with mixed couplings and 
$\beta=1.0$. First row, from left to right: error, 
computation time and fraction of converged instances, as a function of 
relative local field strength $\Theta$ for various methods, averaged over 16 randomly generated 
instances (where results are only included if the method has converged).
For the same instances, scatter plots of errors are shown in the 
next rows for various pairs of methods.}
\end{figure}

\begin{figure}[p]
\centering
\includegraphics[scale=0.666666]{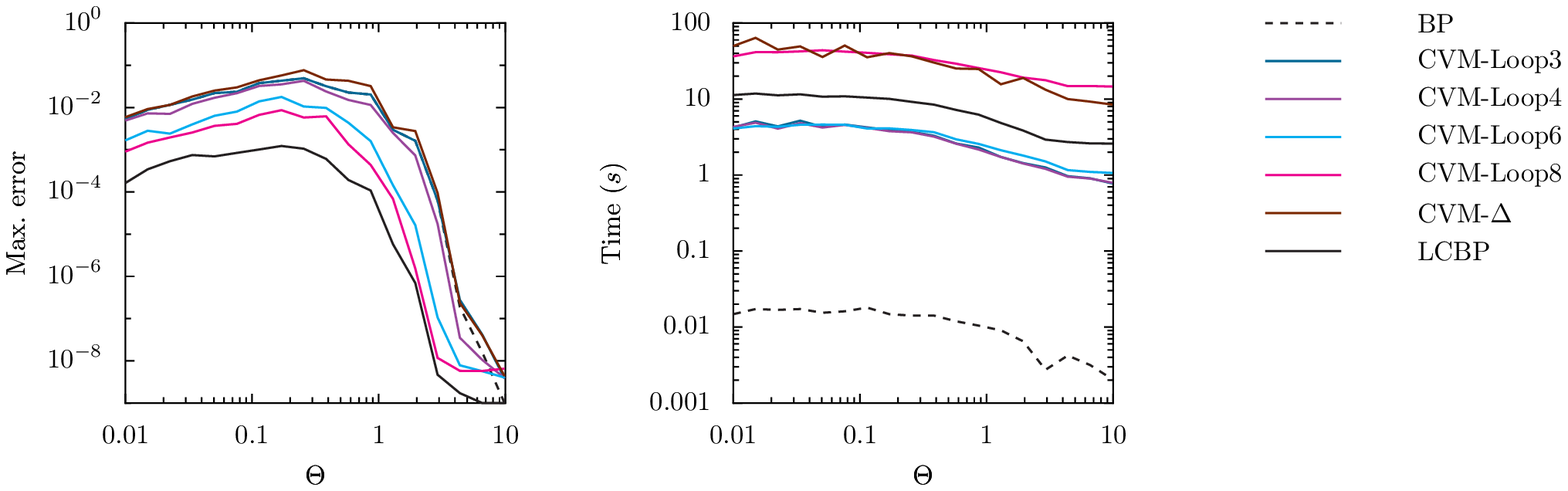}\\[1cm]
\begin{tabular}{ccc}
  \includegraphics[width=0.3\textwidth]{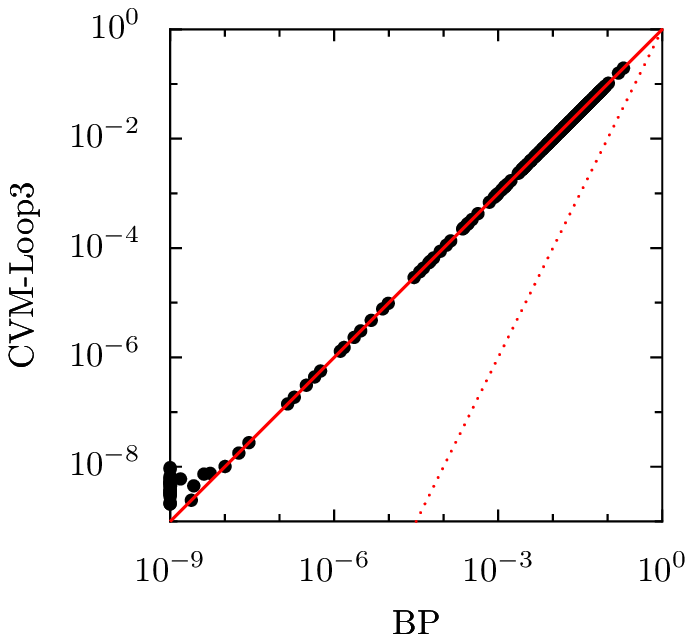}
& \includegraphics[width=0.3\textwidth]{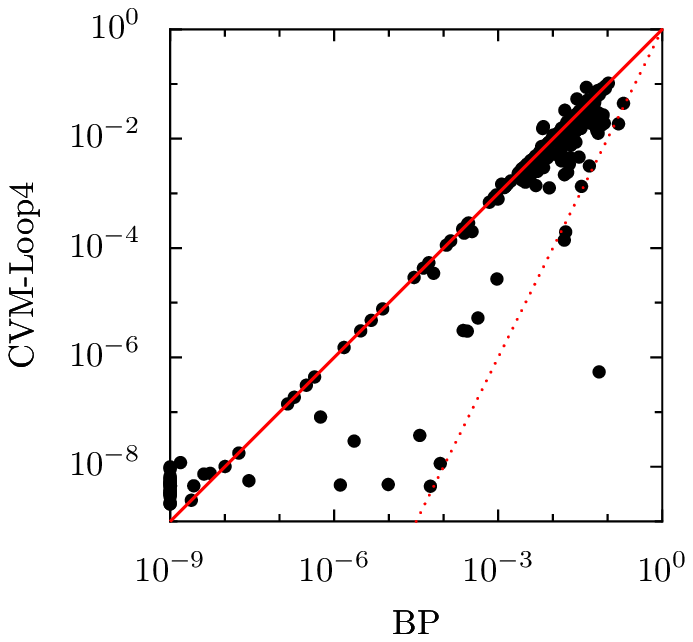}
& \includegraphics[width=0.3\textwidth]{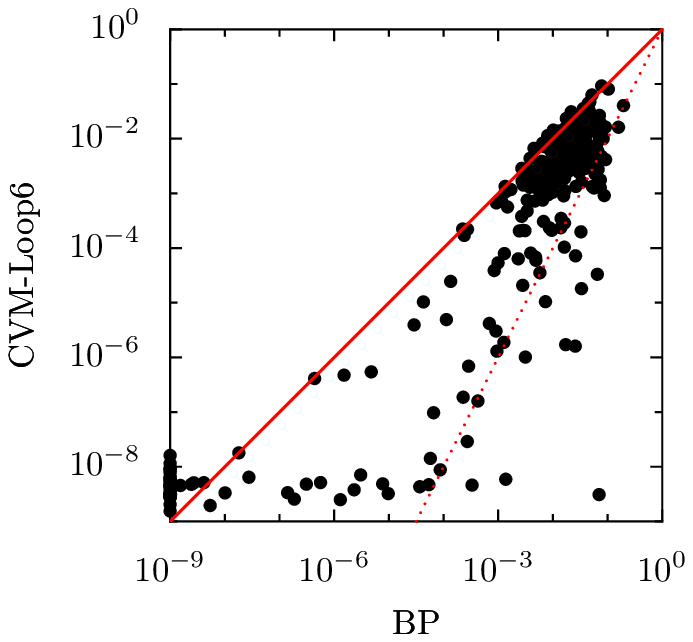}\\[0.5cm]
  \includegraphics[width=0.3\textwidth]{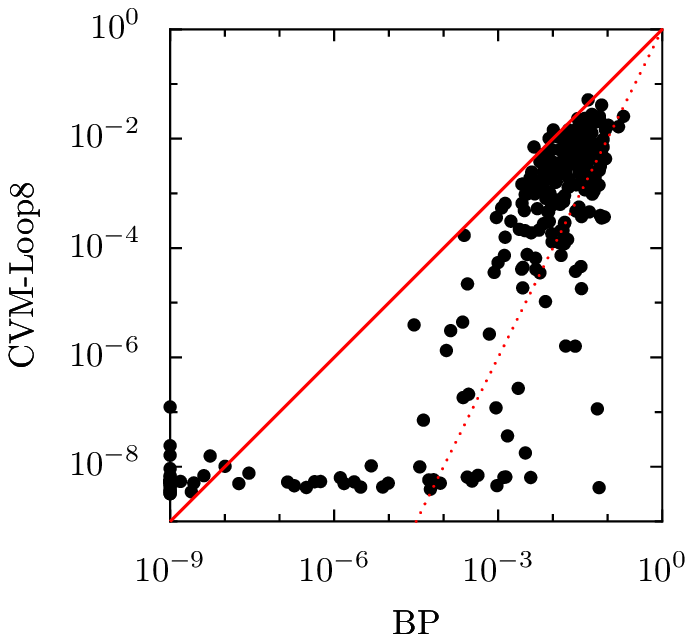}
& \includegraphics[width=0.3\textwidth]{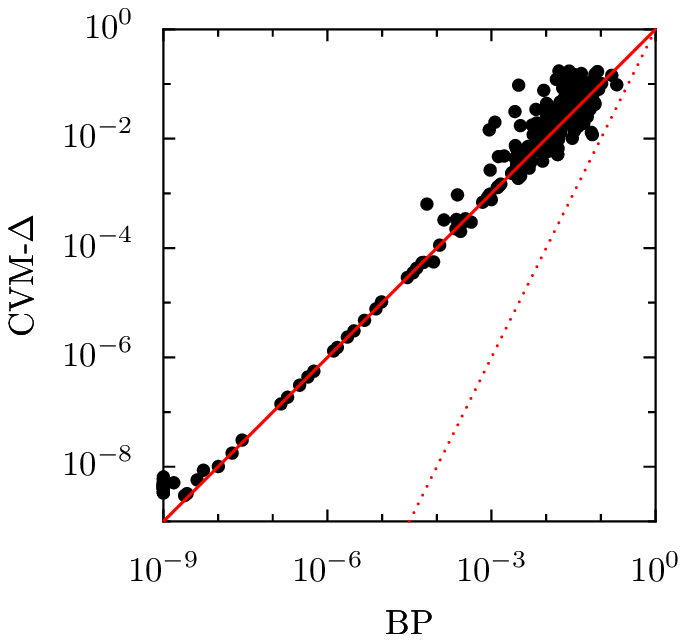}
& \includegraphics[width=0.3\textwidth]{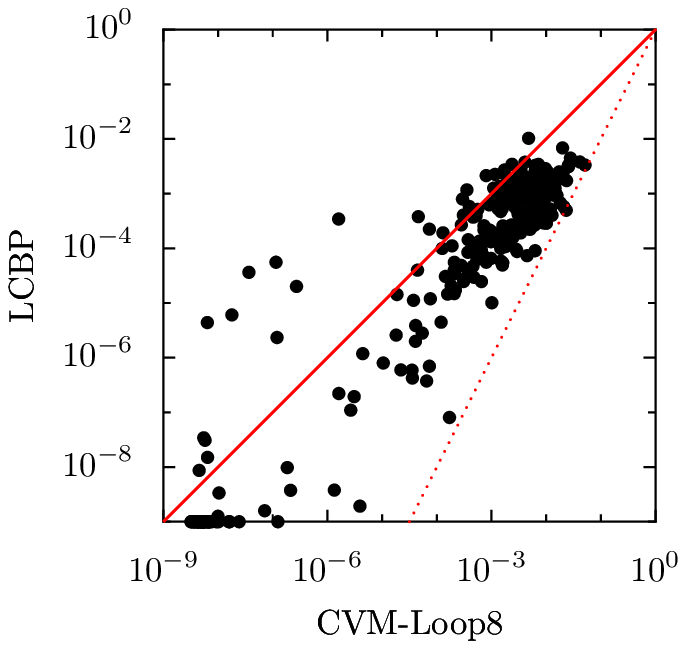}
\end{tabular}
\caption{\label{fig:dreg_d3_N100_sg_theta_b}
Additional results for $(N=100,d=3)$ regular random graphs with mixed couplings and 
$\beta=1.0$, for the same instances as in Figure \ref{fig:dreg_d3_N100_sg_theta}. 
All methods converged on all instances.}
\end{figure}

Computation time is seen to decrease with increasing local field strength
$\Theta$.  The errors on the other hand first increase slowly, and then
suddenly decrease rapidly. Again, LCBP-Cum and LCBP-Cum-Lin are the only
methods that have convergence problems. The ranking in terms of accuracy of
various methods does not depend on the local field strength, nor does the
ranking in terms of computation time. 
}

\subsubsection{Larger degree ($d = 6$)}

\begin{figure}[p]
\centering
\ifthenelse{\isundefined{\techrep}}{
	\includegraphics[scale=0.666666]{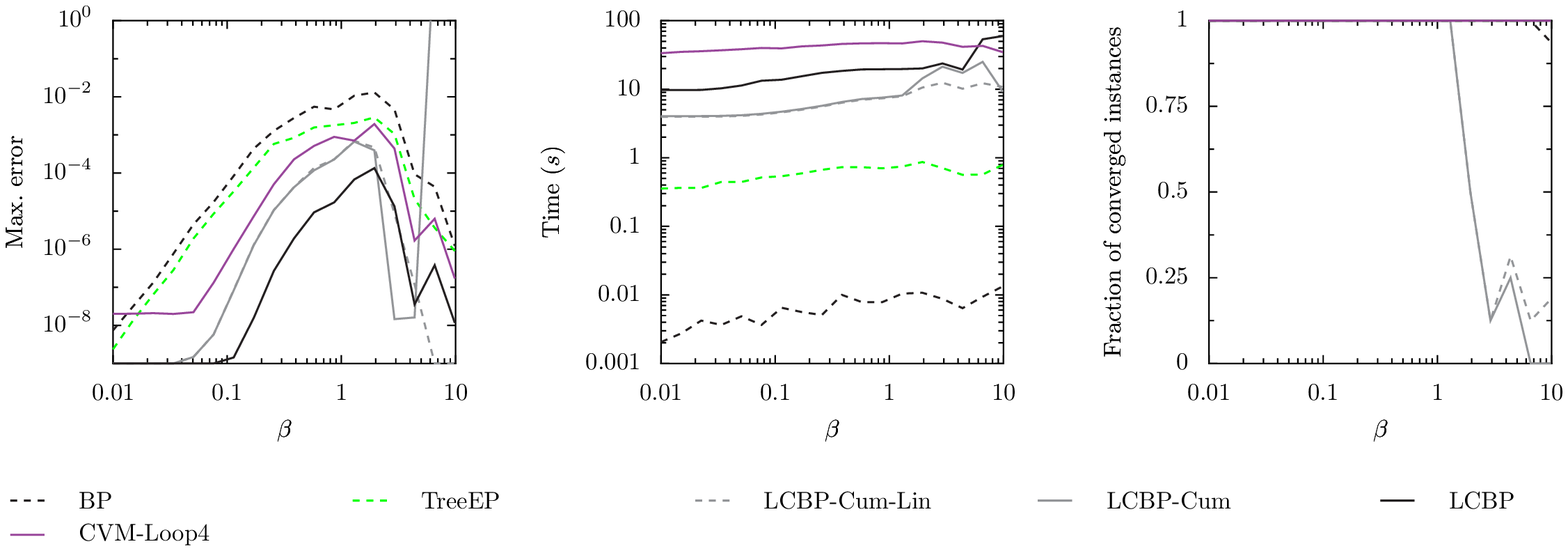}\\[0.5cm]
	\begin{tabular}{ccc}
	  \includegraphics[width=0.3\textwidth]{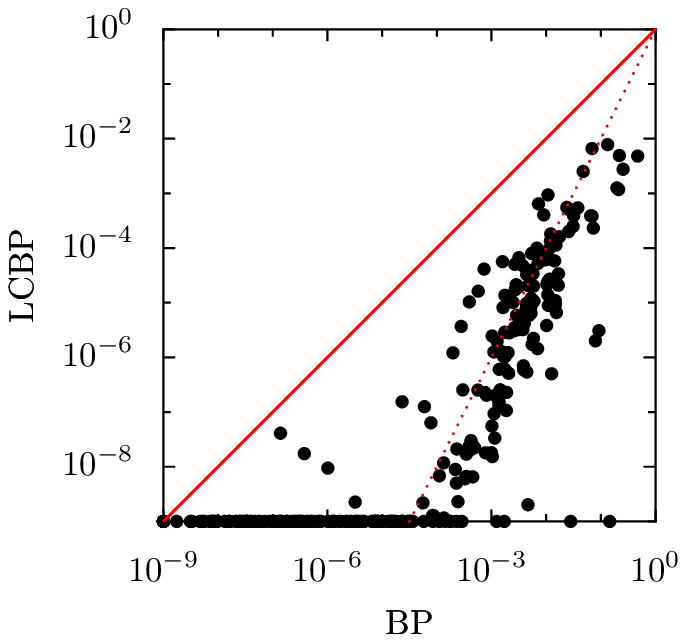}
	& \includegraphics[width=0.3\textwidth]{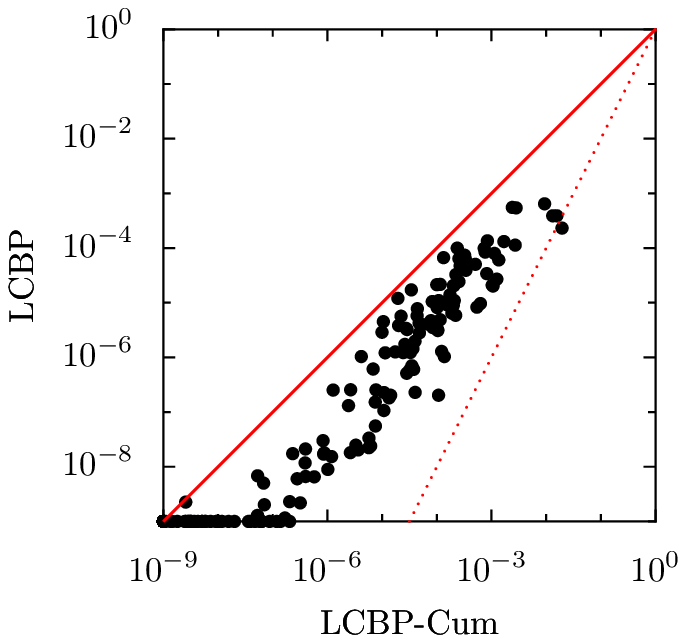} 
	& \includegraphics[width=0.3\textwidth]{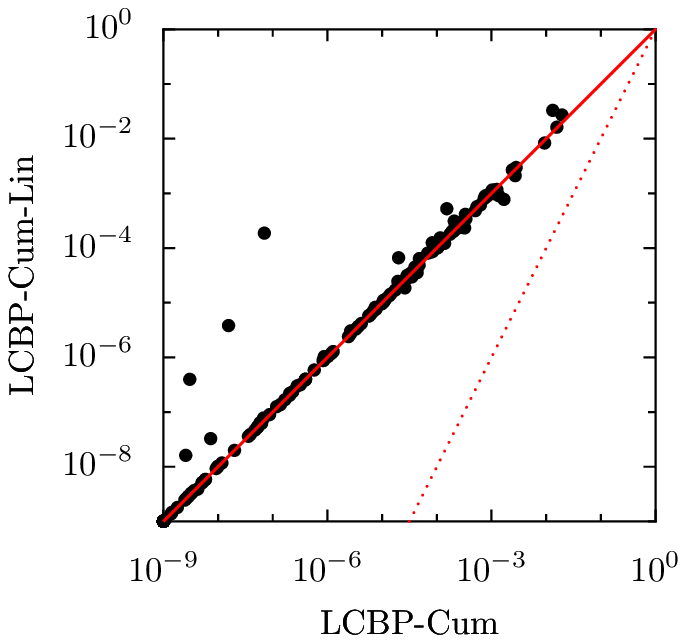}
	\end{tabular}
}{
	\includegraphics[scale=0.666666]{dreg_d6_N50_sg.eps}\\[0.5cm]
	\begin{tabular}{ccc}
	  \includegraphics[width=0.3\textwidth]{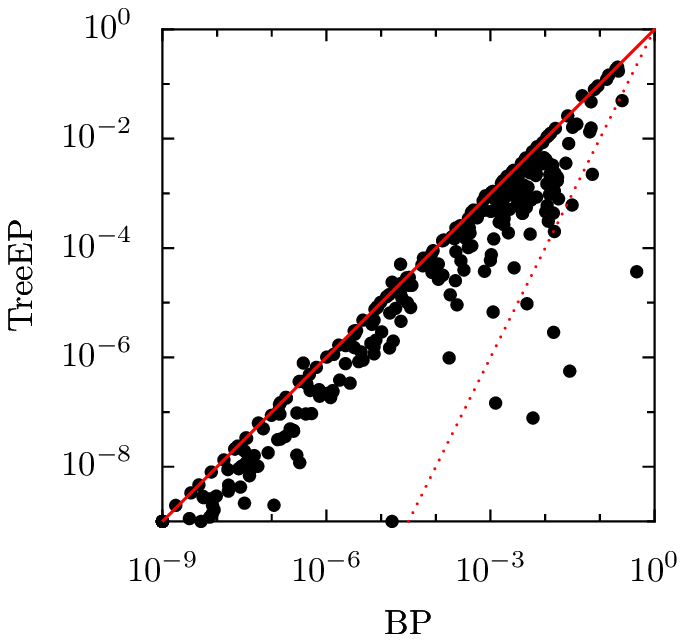}
	& \includegraphics[width=0.3\textwidth]{dreg_d6_N50_sg_BP_LCBP.eps}
	& \includegraphics[width=0.3\textwidth]{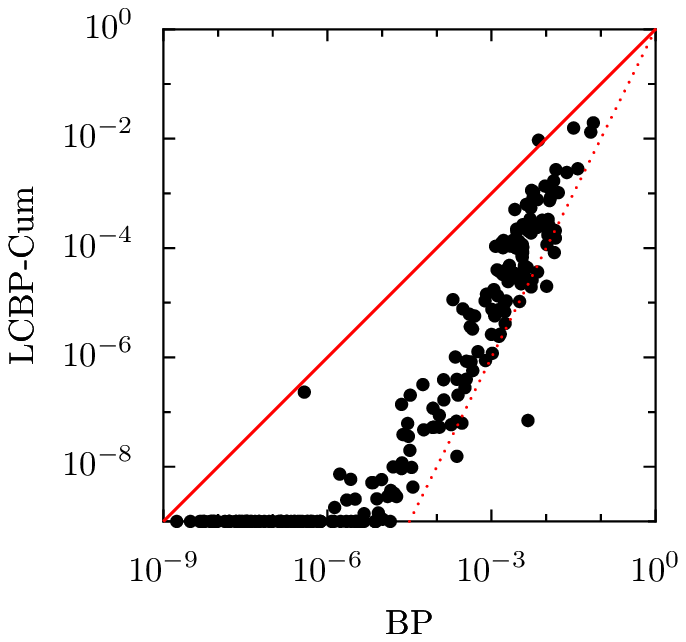}\\[0.4cm]
	  \includegraphics[width=0.3\textwidth]{dreg_d6_N50_sg_MR_LCBP.eps}
	& \includegraphics[width=0.3\textwidth]{dreg_d6_N50_sg_MR_MRLINEAR.eps}
	& \includegraphics[width=0.3\textwidth]{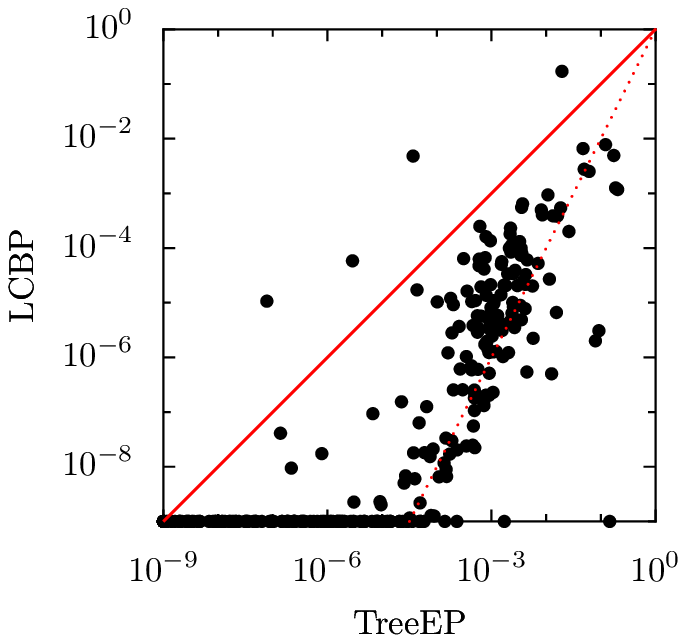}\\[0.4cm]
	  \includegraphics[width=0.3\textwidth]{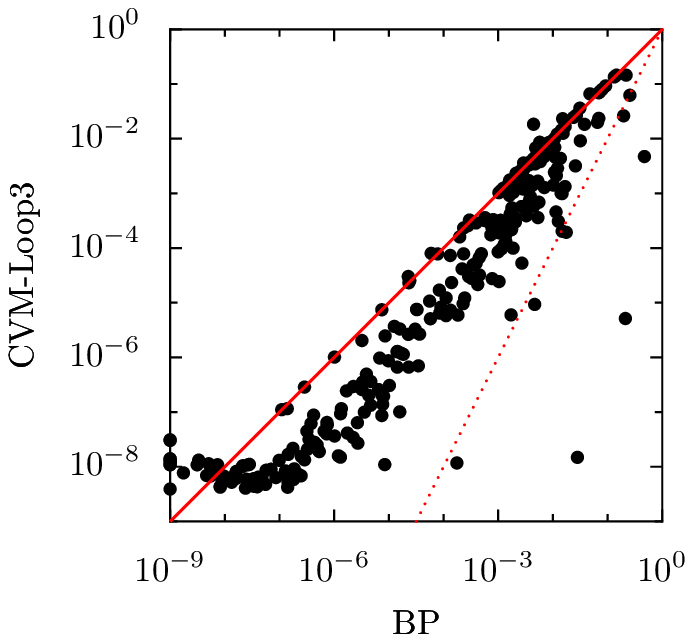}
	& \includegraphics[width=0.3\textwidth]{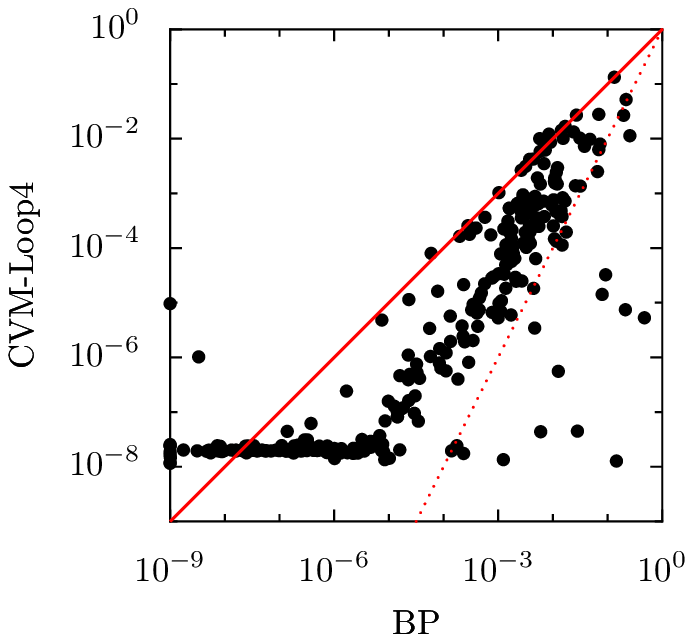}
	\end{tabular}
}
\caption{\label{fig:dreg_d6_N50_sg}
Results for $(N=50,d=6)$ regular random graphs with mixed couplings and 
strong local fields $\Theta=2$.
}
\end{figure}

To study the influence of the degree $d = \nel{\del{i}}$, we have done
additional experiments for $d = 6$. We had to reduce the number of variables to
$N = 50$, because exact inference was infeasible for larger values of $N$ due to
quickly increasing treewidth. The results are shown in Figure \ref{fig:dreg_d6_N50_sg}.

As in the previous experiments, BP is the fastest and least accurate method,
whereas LCBP yields the most accurate results, even for high $\beta$.

The differences with the case of low degree ($d = 3$) are the following.  The
relative improvement of TreeEP over BP has decreased. This could have been
expected, because in denser networks, the effect of taking out a tree becomes
less. Further, the relative improvement of CVM-Loop4 over BP has increased,
probably because there are more short loops present. On the other hand,
computation time of CVM-Loop4 has also increased and it is the slowest of all
methods. We decided to abort the calculations for CVM-Loop6 and CVM-Loop8,
because computation time was prohibitive due to the enormous amount of short
loops present.  We conclude that the CVM-Loop approach to loop correcting is not
very efficient.
Surprisingly, the results of LCBP-Cum-Lin are now very similar in quality
to the results of LCBP-Cum, except for a few isolated cases (presumably 
on the edge of the convergence region).
LCBP now clearly needs more computation time than LCBP-Cum and LCBP-Cum-Lin,
but also obtains significantly better results due to the fact that it takes 
into account higher order cavity interactions.

\subsubsection{Influence of the coupling type}

\begin{figure}[p]
\centering
\ifthenelse{\isundefined{\techrep}}{
	\includegraphics[scale=0.666666]{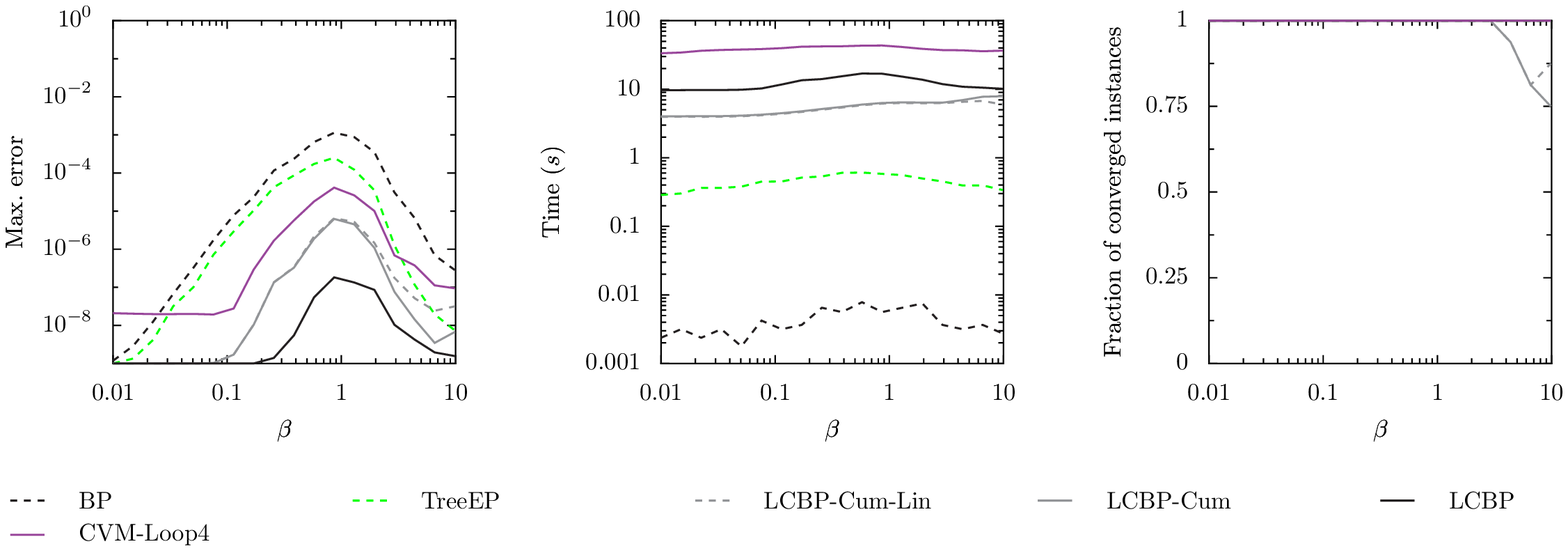}
}{
	\includegraphics[scale=0.666666]{dreg_d6_N50_fe.eps}\\[0.5cm]
	\begin{tabular}{ccc}
	  \includegraphics[width=0.3\textwidth]{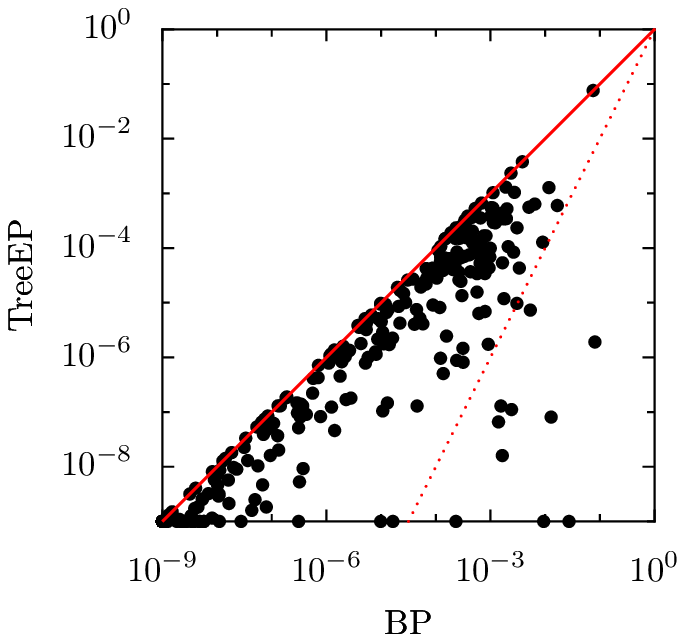}
	& \includegraphics[width=0.3\textwidth]{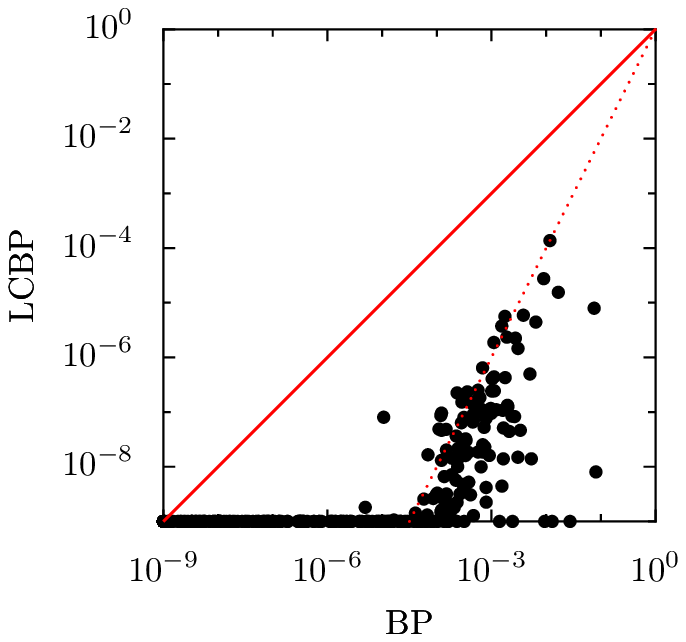}
	& \includegraphics[width=0.3\textwidth]{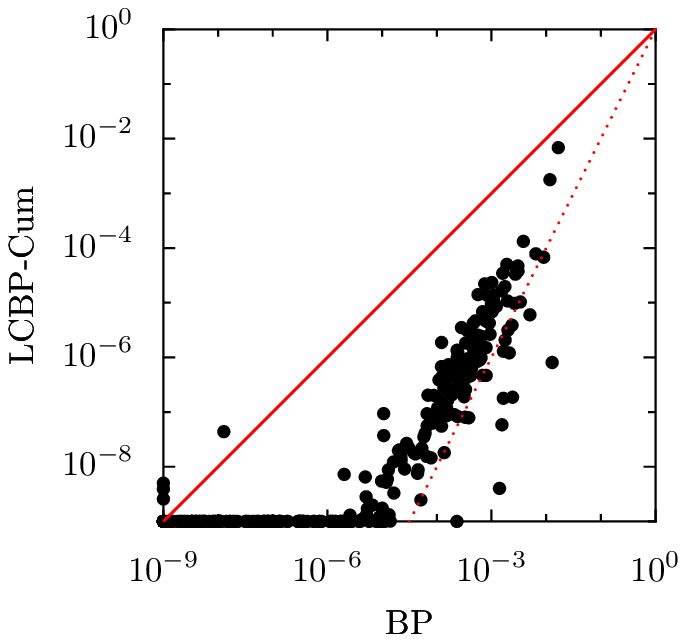}\\[0.4cm]
	  \includegraphics[width=0.3\textwidth]{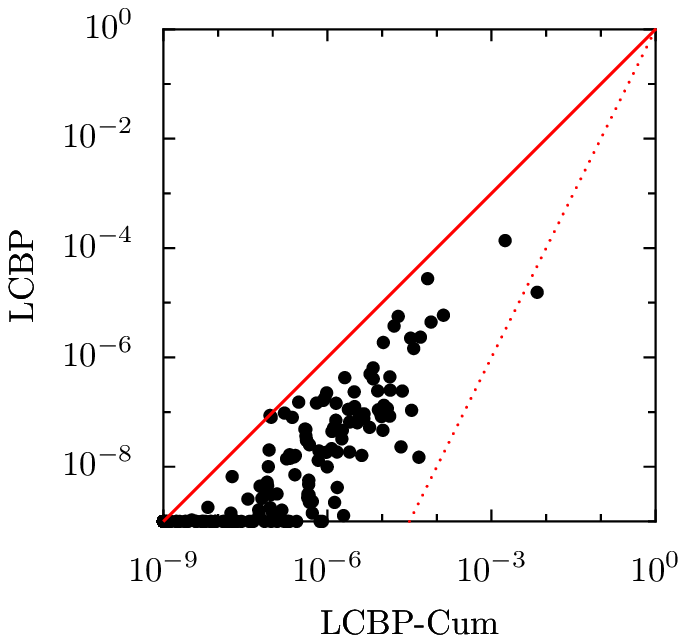}
	& \includegraphics[width=0.3\textwidth]{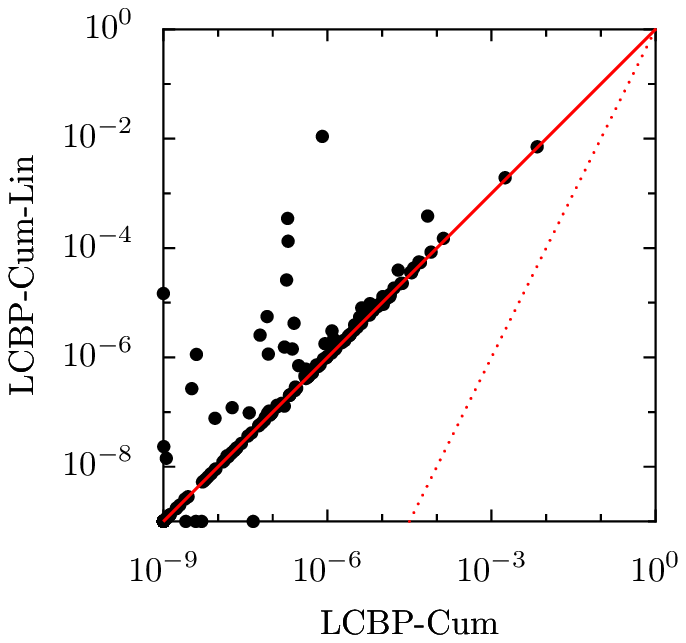}
	& \includegraphics[width=0.3\textwidth]{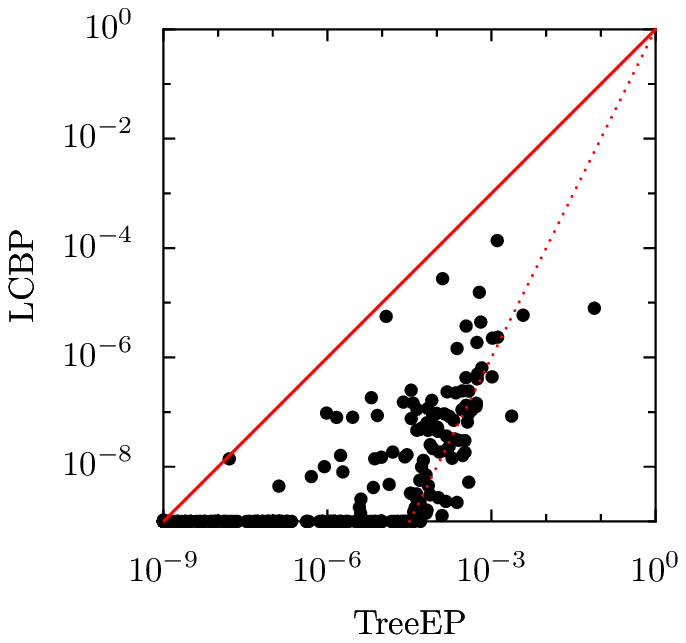}\\[0.4cm]
	  \includegraphics[width=0.3\textwidth]{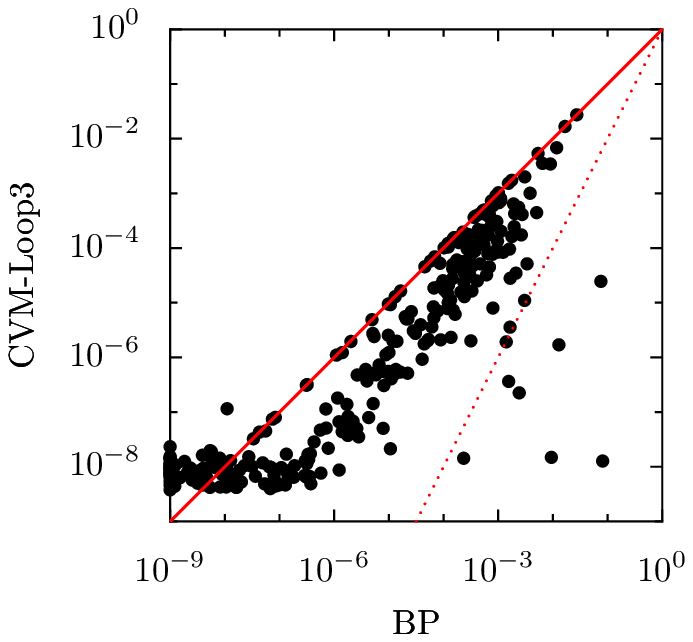}
	& \includegraphics[width=0.3\textwidth]{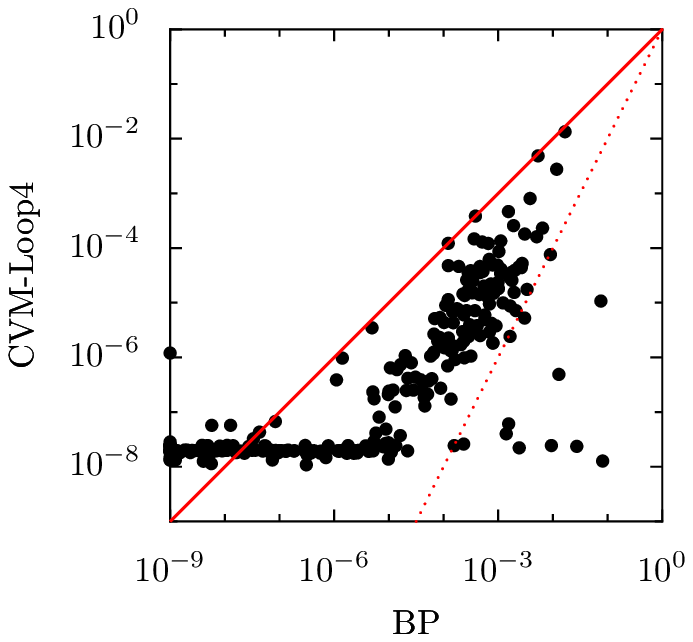}
	\end{tabular}
}
\caption{\label{fig:dreg_d6_N50_fe}
Results for $(N=50,d=6)$ regular random graphs with attractive couplings and 
strong local fields $\Theta=2$.
}
\end{figure}

To study the influence of coupling type, we have done additional experiments
for $(N = 50, d=6)$ random regular graphs with attractive couplings and
strong local fields ($\Theta = 2$). The results are shown in
\ref{fig:dreg_d6_N50_fe}.

By comparing Figures \ref{fig:dreg_d6_N50_sg} and \ref{fig:dreg_d6_N50_fe}, it
becomes clear that the influence of the coupling type is rather small.
Differences might be more pronounced in case of weak local fields (for which we
have not done additional experiments).

\subsubsection{Scaling with $N$}

\begin{figure}[t]
\centering
\includegraphics[scale=0.666666]{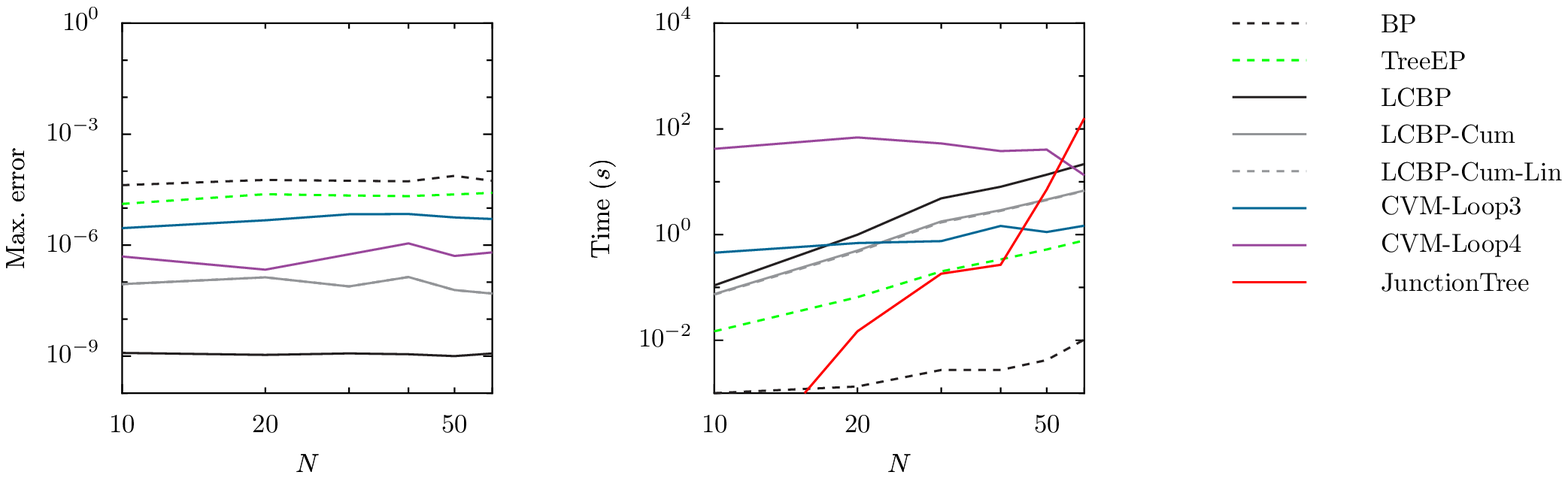}
\caption{\label{fig:dreg_d6_sg}
Error (left) and computation time (right) as a function of $N$ (the number of
variables), for random graphs with uniform degree $d=6$, mixed couplings,
$\beta = 0.1$ and $\Theta = 2$. Points are averages over 16 randomly generated
instances. Each method converged on all instances. 
}
\end{figure}

We have investigated how computation time scales with the number of variables
$N$,  for fixed $\beta = 0.1$, $\Theta = 2$ and $d = 6$ for mixed couplings. We
used a machine with more memory (16 GB) to be able to do exact inference
without swapping also for $N = 60$. The results can be found in Figure
\ref{fig:dreg_d6_sg}. For larger values of $N$, the computation time for exact
inference would increase exponentially with $N$.

The error of all methods is approximately constant. BP should scale
approximately linearly in $N$.  LCBP variants are expected to scale quadratic
in $N$ (since $d$ is fixed) which indeed appears to be the case. The
computation time of the exact JunctionTree method quickly increases due to
increasing treewidth; for $N = 60$ it is already ten times larger than the
computation time of the slowest approximate inference method. The computation
time of CVM-Loop3 and CVM-Loop4 seems to be approximately constant, probably
because the large number of overlaps of short loops for small values of $N$
causes difficulties.

We conclude that for large $N$, exact inference is infeasible, whereas LCBP
still yields very accurate results using moderate computation time. 

\subsection{Scaling with $d$}

\begin{figure}[t]
\centering
\includegraphics[scale=0.666666]{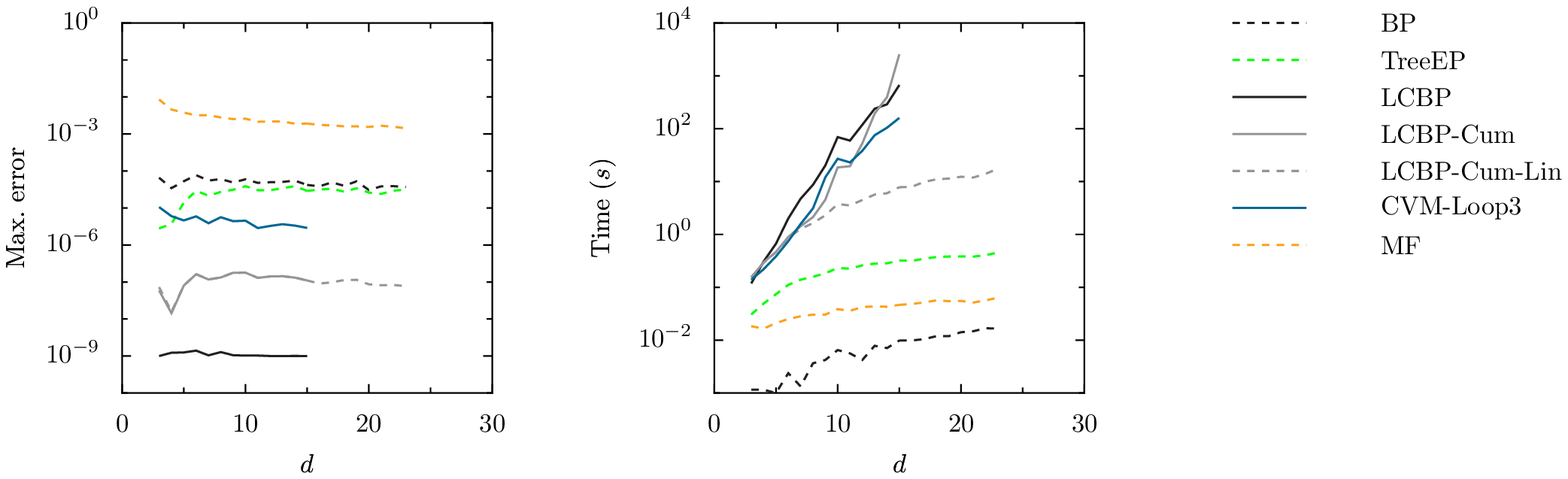}
\caption{\label{fig:dreg_N24_sg}
Error (left) and computation time (right) as a function of variable degree $d$ for 
regular random graphs of $N=24$ variables with mixed couplings for $\beta = 0.1$
and $\Theta = 2$. Points are averages over 16 randomly generated instances. Each
method converged on all instances.}
\end{figure}

It is also interesting to see how various methods scale with $d$, the variable
degree, which is directly related to the cavity size. We have done experiments
for random graphs of size $N=24$ with fixed $\beta = 0.1$ and $\Theta=2$ for
mixed couplings for different values of $d$ between 3 and 23. The results can
be found in Figure \ref{fig:dreg_N24_sg}. We aborted the calculations of the
slower methods (LCBP, LCBP-Cum, CVM-Loop3) at $d=15$.

Due to the particular dependence of the interaction strength on $d$, the
errors of most methods depend only slightly on $d$. TreeEP is an exception: for
larger $d$, the relative improvement of TreeEP over BP diminishes, and the
TreeEP error approaches the BP error.  CVM-Loop3 gives better quality, but
needs relatively much computation time and becomes very slow for large
$d$ due to the large increase in the number of loops of 3 variables.
LCBP is the most accurate method, but becomes very slow for large $d$.
LCBP-Cum is less accurate and becomes slower than LCBP for large $d$, because
of the additional overhead of the combinatorics needed to perform the update
equations. The accuracy of LCBP-Cum-Lin is indistinguishable from that of
LCBP-Cum, although it needs significantly less computation time. 

\subsection{Alternative methods to obtain initial approximate cavity distributions}

\begin{figure}[t]
\centering
\includegraphics[scale=0.666666]{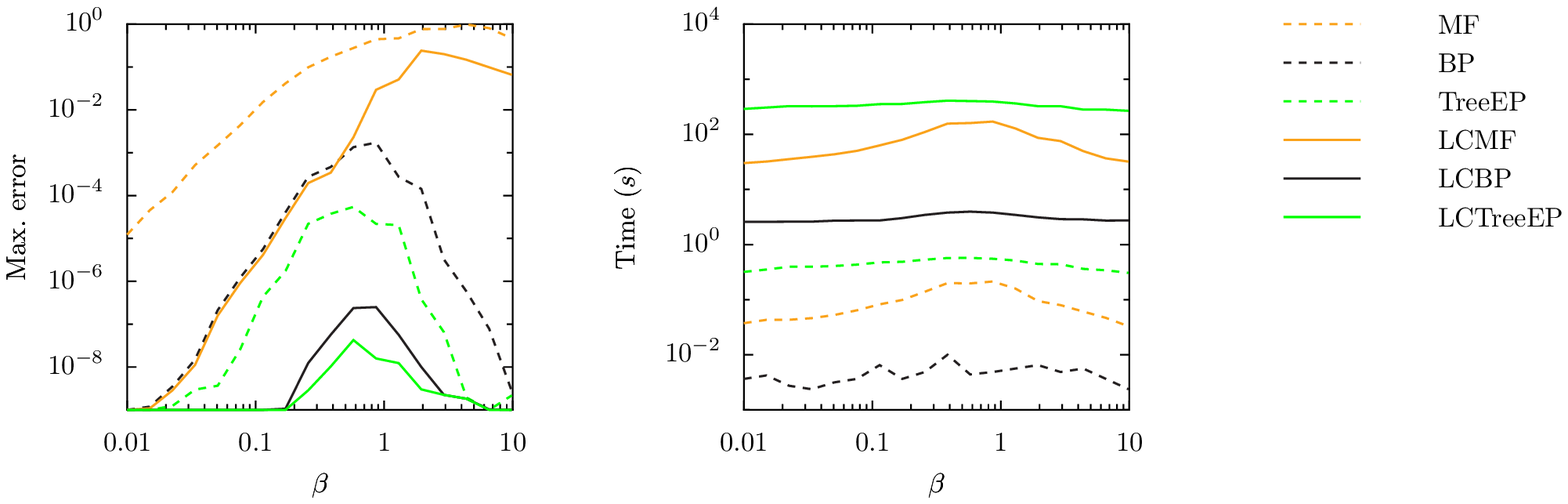}\\[1cm]
\begin{tabular}{ccc}
  \includegraphics[width=0.3\textwidth]{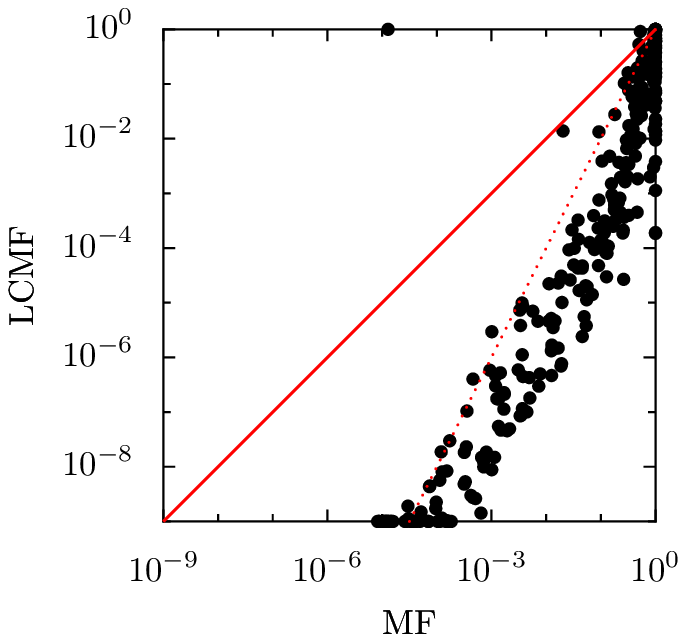}
& \includegraphics[width=0.3\textwidth]{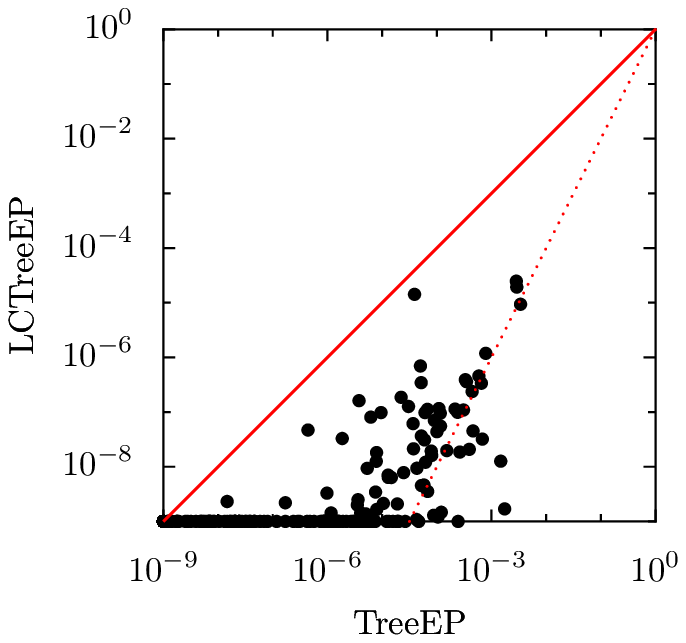}
& \includegraphics[width=0.3\textwidth]{dreg_d3_N100_sg_BP_LCBP.eps}
\end{tabular}
\caption{\label{fig:dreg_d3_N100_sg_scatters_alt}
Results for different methods of obtaining initial estimates of cavity
distributions, for $(N=100,d=3)$ regular random graphs 
with mixed couplings and strong local fields $\Theta=2$.}
\end{figure}

Until now we have used BP to estimate initial cavity approximations. We now
show that other approximate inference methods can be used as well and that a
similar relative improvement in accuracy is obtained. Figure
\ref{fig:dreg_d3_N100_sg_scatters_alt} shows the results of Algorithm
\ref{alg:LCBP} for cavity approximations initialized using the method described
in Section \ref{sec:initial_cavities_full} with MF and TreeEP instead of BP. For
reference, also the BP results are plotted. In all cases, the loop corrected
error is approximately the square of the error of the uncorrected approximate
inference method. Because BP is very fast yet relatively accurate, we focus on
LCBP in this article.

\subsection{Multi-variable factors}

We now go beyond pairwise interactions and study a class of random factor
graphs with binary variables and uniform factor degree $\nel{I} = k$ (for all
$I \in \facs$) with $k > 2$. The number of variables is $N$ and the number of
factors is $M$. The factor graphs are constructed by starting from an empty
graphical model $(\vars,\emptyset,\emptyset)$ and adding $M$ random factors,
where each factor is obtained in the following way: a subset $I = \{I_1, \dots,
I_k\} \subseteq \vars$ of $k$ different variables is drawn; a vector of $2^k$
independent random numbers $\{J_I(x_I)\}_{x_I \in \X{I}}$ is drawn from a
$\C{N}(0,\beta)$ distribution; the factor $\facx{I} := \exp J_I(x_i)$ is added
to the graphical model. We only use those constructed factor graphs that are 
connected.\footnote{The reason that we require the factor graph to
be connected is that not all our approximate inference method implementations
currently support connected factor graphs that consist of more than one
connected component.} The parameter $\beta$ again controls the interaction strength.

\begin{figure}[p]
\centering
\ifthenelse{\isundefined{\techrep}}{
	\begin{tabular}{cc}
	\includegraphics[scale=0.666666]{hoi_N50_M50_k3b.eps} &
	\includegraphics[scale=0.666666]{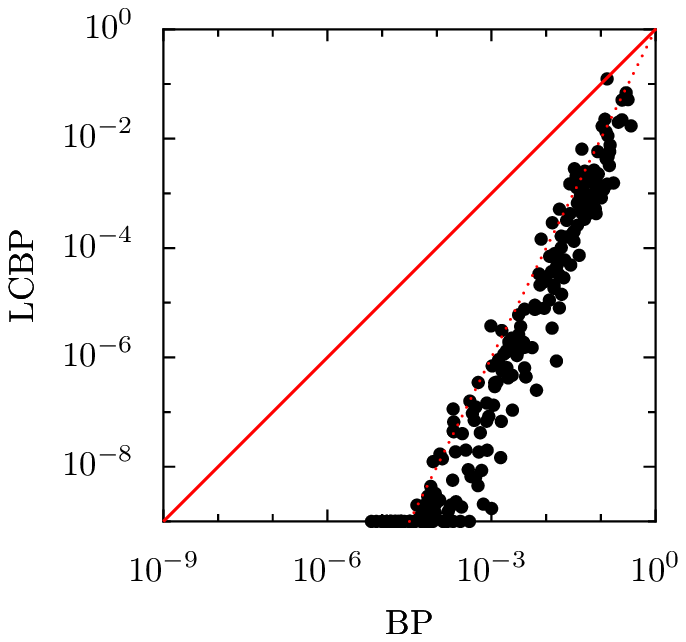}
	\end{tabular}
}{
	\includegraphics[scale=0.666666]{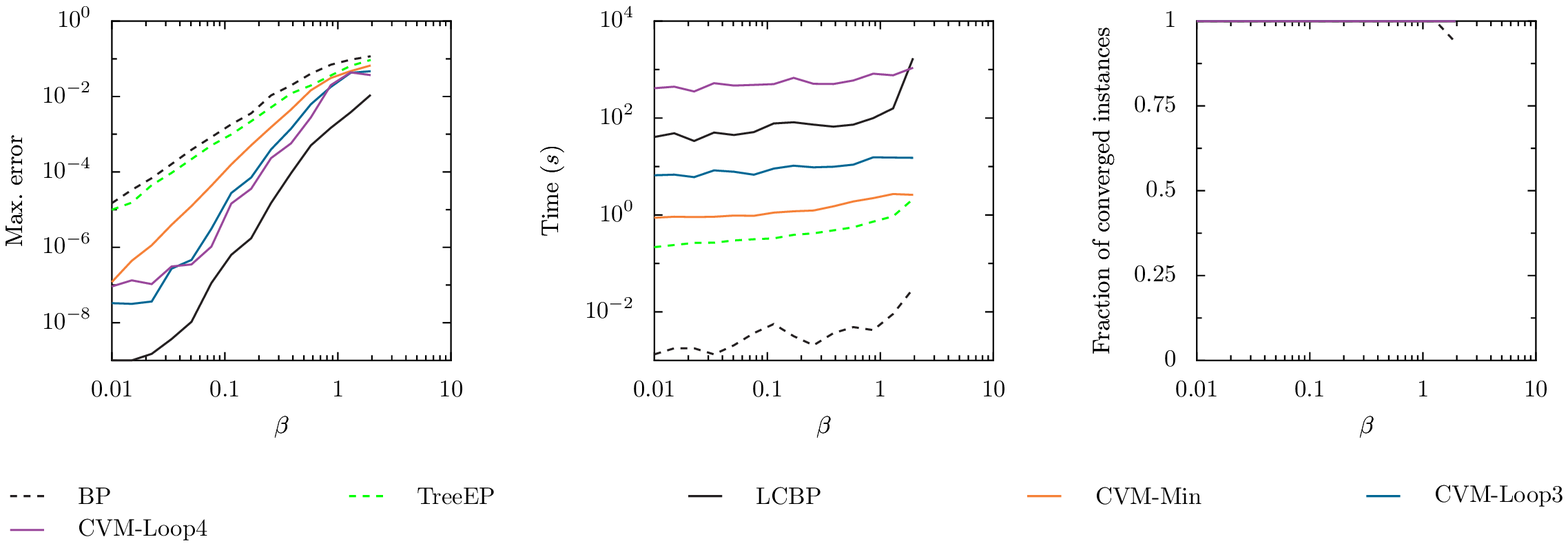}\\[1cm]
	\begin{tabular}{ccc}
	  \includegraphics[width=0.3\textwidth]{hoi_N50_M50_k3_BP_LCBP.eps}
	& \includegraphics[width=0.3\textwidth]{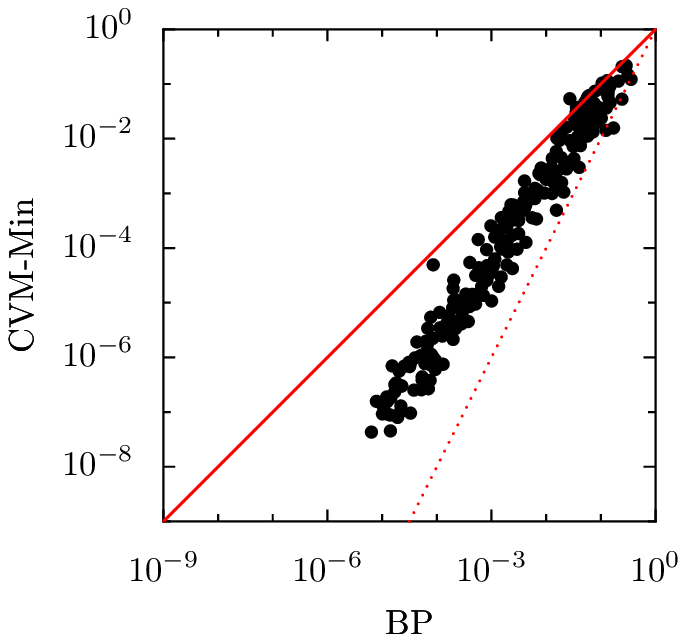}\\[0.5cm]
	  \includegraphics[width=0.3\textwidth]{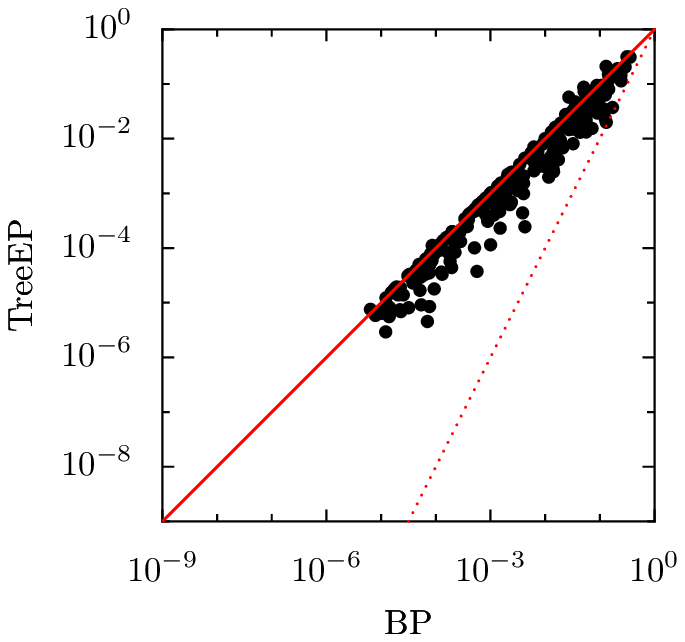}
	& \includegraphics[width=0.3\textwidth]{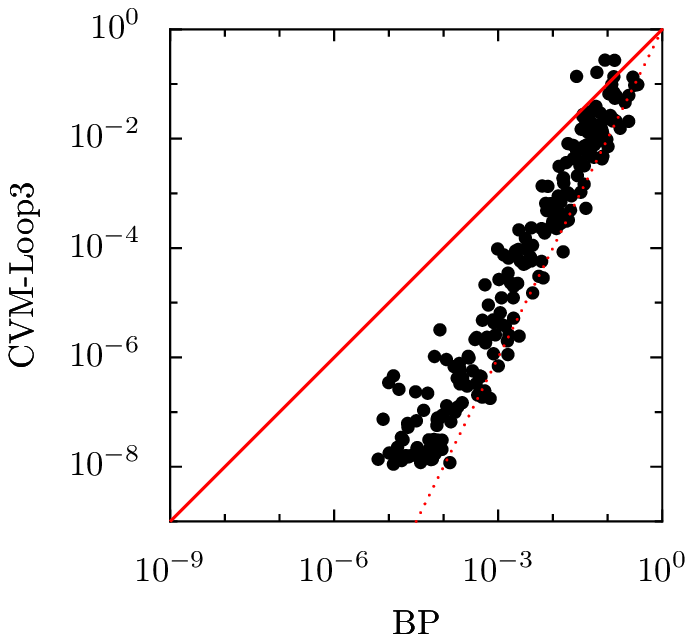}
	& \includegraphics[width=0.3\textwidth]{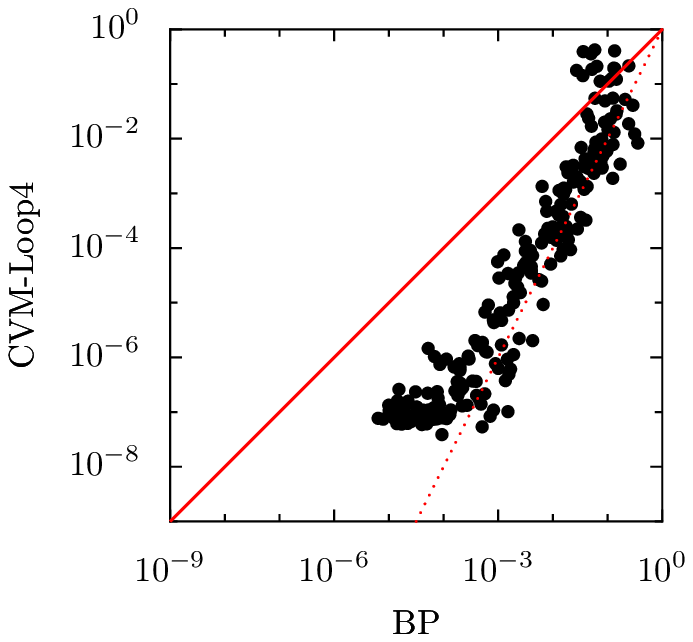}
	\end{tabular}
}
\caption{\label{fig:hoi_N50_M50_k3}
Results for $(N=50,M=50,k=3)$ random factor graphs.}
\end{figure}

We have done experiments for $(N=50,M=50,k=3)$ for various values of $\beta$
between 0.01 and 2. For each value of $\beta$, we have used 16 random
instances. For higher values of $\beta$, computation times increased and
convergence became problematic for some methods, which can probably be explained as the effects of a phase
transition. The results are shown in Figure \ref{fig:hoi_N50_M50_k3}.

Looking at the error and the computation time in Figure
\ref{fig:hoi_N50_M50_k3}, the following ranking can be made, where accuracy and
computation time both increase: BP, TreeEP, CVM-Min, CVM-Loop3, LCBP. CVM-Loop4
uses more computation time than LCBP but gives worse results. LCBP-Cum and
LCBP-Cum-Lin are not available due to the fact that the factors involve more
than two variables. The improvement of TreeEP over BP is rather small.

\subsection{ALARM network}

The ALARM
network\footnote{The ALARM network can be downloaded from \url{http://compbio.cs.huji.ac.il/Repository/Datasets/alarm/alarm.dsc}}
is a well-known Bayesian network consisting of 37 variables (some of which can
take on more than two possible values) and 37 factors (many of which involve
more than two variables). In addition to the usual approximate inference
methods, we have compared with GBP-Min, a GBP implementation of the minimal CVM
approximation that uses maximal factors as outer clusters.  The results are
reported in Table \ref{tab:ALARM}.

\begin{table}[b!]
\caption{\label{tab:ALARM} Results for the ALARM network}
\bigskip
\begin{center}
\begin{tabular}{l|rl}
Method                  & Time ($s$) & Error \\ 
\hline
BP                      &       0.00 &  $2.026\cdot 10^{-01}$ \\ 
TreeEP                  &       0.21 &  $3.931\cdot 10^{-02}$ \\ 
GBP-Min                 &       0.18 &  $2.031\cdot 10^{-01}$ \\ 
CVM-Min                 &       1.13 &  $2.031\cdot 10^{-01}$ \\ 
CVM-$\Delta$            &     280.67 &  $2.233\cdot 10^{-01}$ \\ 
CVM-Loop3               &       1.19 &  $4.547\cdot 10^{-02}$ \\ 
CVM-Loop4               &     154.97 &  $3.515\cdot 10^{-02}$ \\ 
CVM-Loop5               &    1802.83 &  $5.316\cdot 10^{-02}$ \\ 
CVM-Loop6               &   84912.70 & $5.752\cdot 10^{-02}$ \\ 
LCBP                    &      23.67 &  $3.412\cdot 10^{-05}$ \\ 
\end{tabular}
\end{center}
\end{table}

The accuracy of GBP-Min (and CVM-Min) is almost identical to that of BP for
this graphical model; GBP-Min converges without damping and is faster
than CVM-Min. TreeEP on the other hand significantly improves the BP result in
roughly the same time as GBP-Min needs. Simply enlarging the cluster size
(CVM-$\Delta$) slightly deteriorates the quality of the results and also causes
an enormous increase of computation time. The quality of the CVM-Loop results
is roughly comparable to that of TreeEP. Suprisingly, increasing the loop depth
beyond 4 deteriorates the quality of the results and results in an explosion of
computation time. We conclude that the CVM-Loop method is not a very good
approach to correcting loops in this case. 
LCBP uses considerable computation time,
but yields errors that are approximately $10^4$ times smaller than BP errors.
The cumulant based loop LCBP methods are not available, due to the presence of
factors involving more than two variables and variables that can take more than
two values.

\subsection{PROMEDAS networks}

In this subsection, we study the performance of LCBP on another ``real-world''
example, the PROMEDAS medical diagnostic network \citep{Wiegerinck+99}.  The diagnostic model in
PROMEDAS is based on a Bayesian network.  The global architecture of this
network is similar to QMR-DT \citep{Shwe+91}.  It consists of a diagnosis-layer
that is connected to a layer with findings\footnote{In addition, there is a
layer of variables, such as age and gender, that may affect the prior
probabilities of the diagnoses. Since these variables are always clamped for
each patient case, they merely change the prior disease probabilities and are
irrelevant for our current considerations.}.  Diagnoses (diseases) are modeled
as \emph{a priori} independent binary variables causing a set of symptoms
(findings), which constitute the bottom layer.  The PROMEDAS network currently
consists of approximately 2000 diagnoses and 1000 findings.

The interaction between diagnoses and findings is modeled with a noisy-OR
structure. The conditional probability of the finding given the parents is
modeled by $m+1$ numbers, $m$ of which represent the probabilities that the finding
is caused by one of the diseases and one that the finding is not caused by any
of the parents. 

The noisy-OR conditional probability tables with $m$ parents can be naively
stored in a table of size $2^m$. This is problematic for the PROMEDAS networks
since findings that are affected by more than 30 diseases are not uncommon in
the PROMEDAS network.  We use an efficient implementation of noisy-OR relations as
proposed by \cite{TakikawaDAmbrosio99} to reduce the size of these tables.
The trick is to introduce dummy variables $s$ and to make use of the property
\begin{eqnarray}
{\rm OR}(x|y_1,y_2,y_3) = \sum_{s}{\rm OR}(x|y_1,s){\rm OR}(s|y_2,y_3)
\end{eqnarray}
The factors on the right hand side involve at most 3 variables
instead of the initial 4 (left). Repeated application of this formula reduces
all factors to triple interactions or smaller.

When a patient case is presented to PROMEDAS, a subset of the findings will be
clamped and the rest will be unclamped. If our goal is to compute the marginal
probabilities of the diagnostic variables only, the unclamped findings and the
diagnoses that are not related to any of the clamped findings can be summed out
of the network as a preprocessing step.  The clamped findings cause an
effective interaction between their parents.  However, the noisy-OR structure
is such that when the finding is clamped to a negative value, the effective
interaction factorizes over its parents. Thus, findings can be clamped to
negative values without additional computation cost \citep{JaakkolaJordan99}.

The complexity of the problem now depends on the set of findings that is given
as input. The more findings are clamped to a positive value, the larger the
remaining network of disease variables and the more complex the inference task.
Especially in cases where findings share more than one common possible
diagnosis, and consequently loops occur, the model can become complex.  

We use the PROMEDAS model to generate virtual patient data by first clamping
one of the disease variables to be positive and then clamping each finding to
its positive value with probability equal to the conditional distribution of
the finding, given the positive disease. The union of all positive findings
thus obtained constitute one patient case. For each patient case, the
corresponding truncated graphical model is generated. The number of disease
nodes in this truncated graph is typically quite large.

\begin{figure}[p]
\centering
\begin{tabular}{ccc}
  \includegraphics[width=0.3\textwidth]{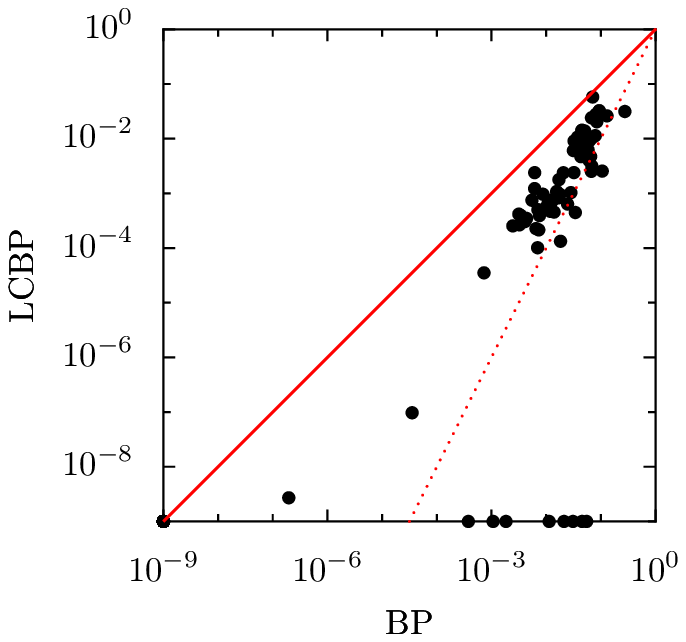}
& \includegraphics[width=0.3\textwidth]{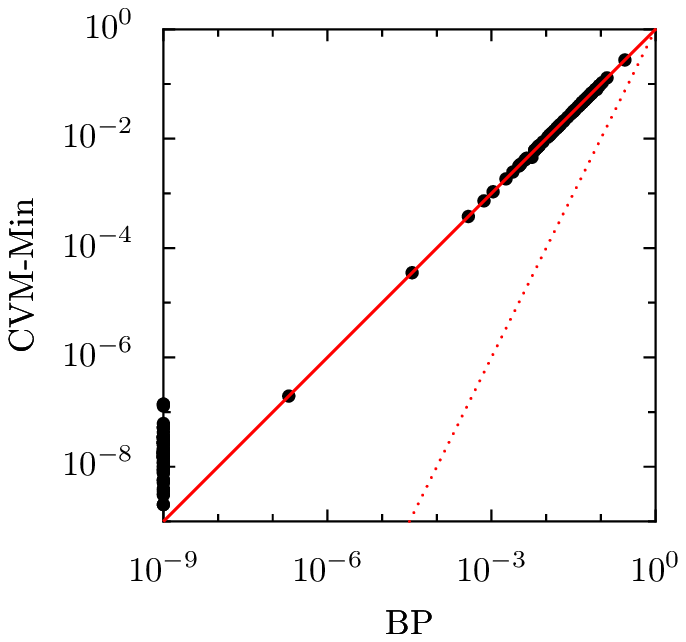}
& \includegraphics[width=0.3\textwidth]{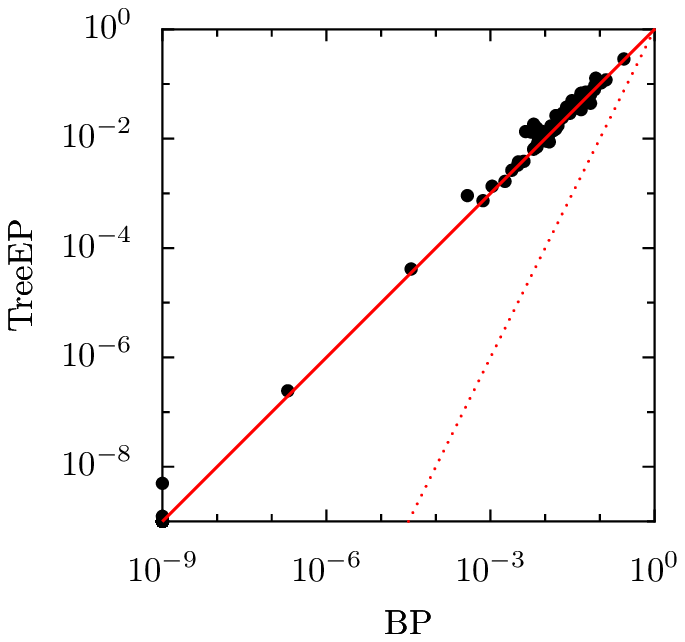}\\[0.5cm]
  \includegraphics[width=0.3\textwidth]{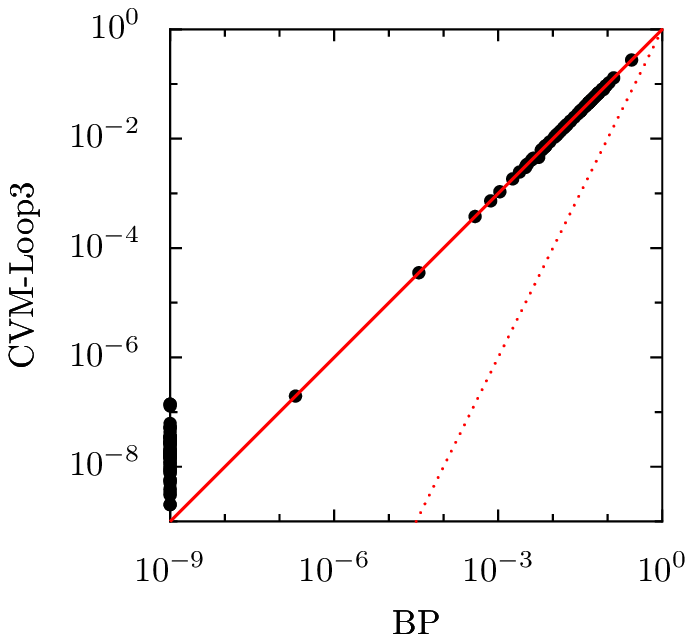}
& \includegraphics[width=0.3\textwidth]{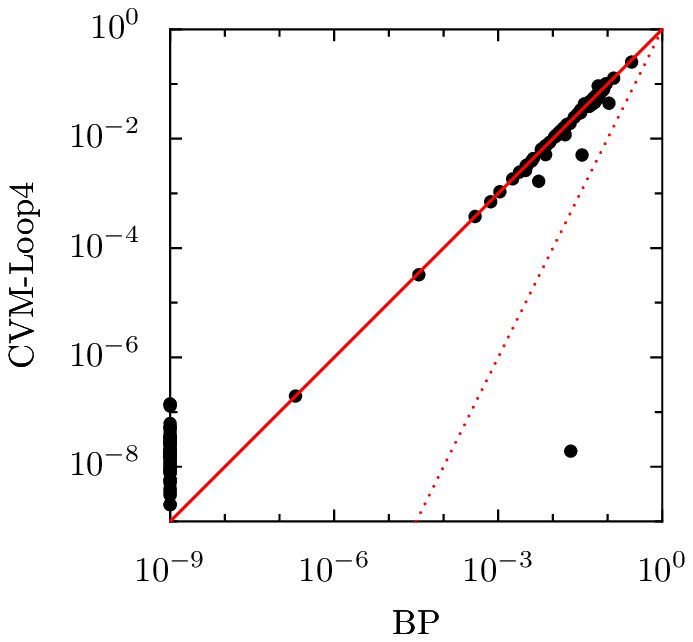}
& \includegraphics[width=0.3\textwidth]{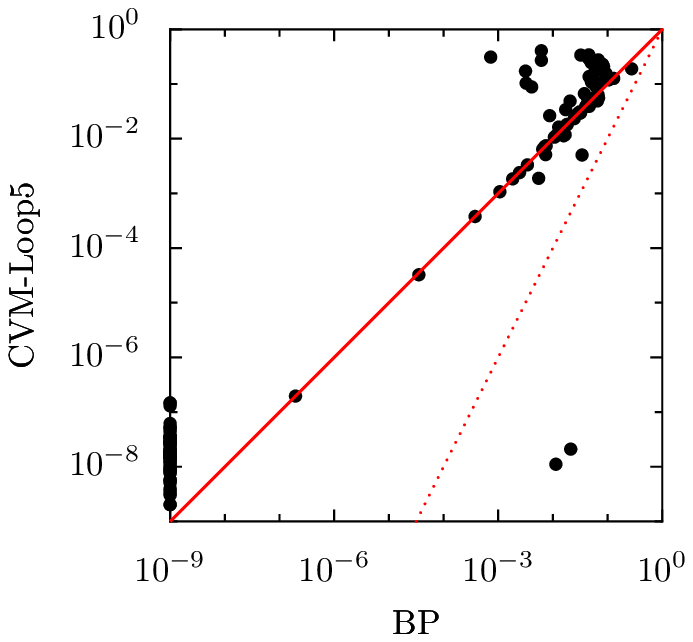}
\end{tabular}
\caption{\label{fig:promedas_scatters}
Scatter plots of errors for PROMEDAS instances.}
\end{figure}

\begin{figure}[h]
\centering
\includegraphics[width=0.3\textwidth]{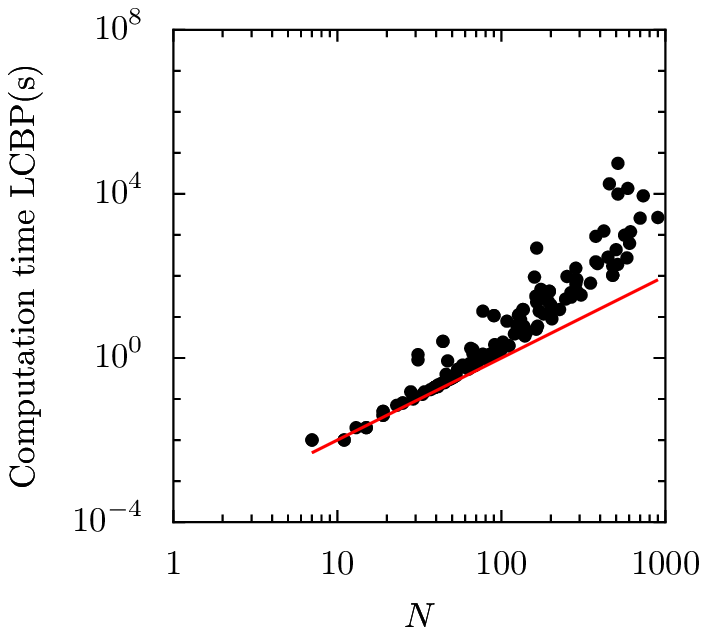}
\caption{\label{fig:promedas_time}
Computation time (in seconds) of LCBP for PROMEDAS instances vs.\ 
$N$, the number of variables in the preprocessed graphical model.
The solid line corresponds to $t \propto N^2$.}
\end{figure}

The results can be found in Figures \ref{fig:promedas_scatters} and
\ref{fig:promedas_time}. Surprisingly, neither TreeEP nor any of the CVM
methods gives substantial improvements over BP. TreeEP even gives worse 
results compared to BP. The CVM-Min and CVM-Loop3 results appear to be almost identical to
the BP results. CVM-Loop4 manages to improve over
BP in a few cases. Increased loop depth ($k=5,6$) results in worse quality in
many cases and also in an enormous increase in computation time.  

LCBP, on the other hand, is the only method that gives a significant
improvement over BP, in each case. Considering all patient cases, LCBP corrects
the BP error with more than one order of magnitude in half of the cases for
which BP was not already exact. The improvement obtained by LCBP has its price:
the computation time of LCBP is rather large compared to that of BP, as shown
in Figure \ref{fig:promedas_time}. The deviation from the quadratic scaling
$t \propto N^2$ is due to the fact that the size of the Markov blankets varies
over instances and instances with large $N$ often also have larger Markov blankets.
The cumulant based loop LCBP methods are not available, due to the presence of
factors involving more than two variables and variables that can take more than
two values.

\section{Discussion and conclusion}\label{sec:discussion}

We have proposed a method to improve the quality of an approximate inference
method (e.g.\ BP) by correcting for the influence of loops in the factor graph.
We found empirically that if one applies this Loop Correcting method, assuming
that no loops are present (by taking factorized initial approximate
cavity distributions), the method reduces to the minimal CVM approximation. We
have proved this for the case of factor graphs that do not have short loops of
exactly four nodes. If, on the other hand, the loop correction method is
applied in combination with BP estimates of the effective cavity interactions,
we have seen that the loop corrected error is approximately the square of the
uncorrected BP error.  Similar observations have been made for loop corrected
MF and TreeEP. For practical purposes, we suggest to apply loop corrections to
BP (``LCBP''), because the loop correction approach requires many runs of the
approximate inference method and BP is well suited for this job because of its
speed. We have compared the performance of LCBP with other approximate inference
methods that (partially) correct for the presence of loops. We have shown that
LCBP is the most accurate method and that it even works for relatively strong
interactions. 

On sparse factor graphs, TreeEP obtains improvements over BP by correcting for
loops that consist of part of the base tree and one additional interaction,
using little computation time. However, for denser graphs, the difference
between the quality of TreeEP and BP marginals diminishes. LCBP almost always
obtained more accurate results. However, LCBP also needs more computation time
than TreeEP.

The CVM-Loop approximation, which uses small loops as outer clusters, can also
provide accurate results if the number of short loops is not too large and the
number of intersections of clusters is limited. However, the computation time
becomes prohibitive in many cases. In order to obtain the same accuracy as
LCBP, the CVM-Loop approach usually needs significantly more computation time.
This behaviour is also seen on ``real world'' instances such as the ALARM
network and PROMEDAS test cases. There may exist other cluster choices that
give better results for the CVM approximation, but no general method for
obtaining ``good'' cluster choices seems to be known (see also
\citep{WellingMinkaTeh05} for a discussion of what constitutes a ``good''
CVM cluster choice).

We have also compared the performance of LCBP with the original implementation proposed
by \cite{MontanariRizzo05}.  This implementation works with cumulants instead
of interactions and we believe that this is the reason that it has more
difficulties in the regime of strong interactions. Although the differences
were rather small in some cases, LCBP obtained better results than LCBP-Cum
using approximately similar amounts of computation time. The linearized version
LCBP-Cum-Lin, which is applicable to factor graphs with large Markov blankets, 
performed surprisingly well, often obtaining similar accuracy as LCBP-Cum. 
For random graphs with high degree $d$ (i.e.\ large Markov blankets), it
turned out to be the most accurate of the
applicable approximate inference methods.  It is rather fortunate that the
negative effect of the linearization error on the accuracy of the result
becomes smaller as the degree increases, since it is precisely for high degree
where one needs the linearization because of performance issues. 

In the experiments reported here, the standard JunctionTree method was almost
always faster than LCBP. The reason is that we have intentionally selected
experiments for which exact inference is still feasible, in order to be able to
compare the quality of various approximate inference methods. However, as
implied by Figure \ref{fig:dreg_d6_sg}, there is no reason to
expect that LCBP will suddenly give inaccurate results when exact inference
is no longer feasible. Thus we suggest that
LCBP may be used to obtain accurate marginal estimates in cases where exact
inference is impossible because of high treewidth. As illustrated in Figure
\ref{fig:dreg_d6_sg}, the computation time of LCBP scales very
different from that of the JunctionTree method: whereas the latter is
exponential in treewidth, LCBP is exponential in the size of the Markov
blankets.

The fact that computation time of LCBP (in its current form) scales
exponentially with the size of the Markov blankets can be a severe limitation
in practice. Many real world Bayesian networks have large Markov blankets,
prohibiting application of LCBP. The linear cumulant based implementation
LCBP-Cum-Lin proposed by \cite{MontanariRizzo05} does not suffer from this
problem, as it is quadratic in the size of the Markov blankets.  Unfortunately,
this particular implementation can in its current form only be applied to
graphical models that consist of binary variables and factors that involve at
most two variables (which excludes any interesting Bayesian network, for
example). Furthermore, problems may arise if some factors contain zeroes. For
general application of loop correcting methods, it will be of paramount
importance to derive an implementation that combines the generality of LCBP
with the speed of LCBP-Cum-Lin. At this point, it is not obvious whether it
would be better to use cumulants or interactions as the parameterization of the
cavity distribution. This topic will be left for future research. The work
presented here provides some intuition that may be helpful for constructing a
general and fast loop correcting method that is applicable to arbitrary factor
graphs that can have large Markov blankets.


Another important direction for future research would be to find an extension
of the loop correcting framework that also gives a loop corrected approximation
of the normalization constant $Z$ in \eref{eq:probability_distribution}.
Additionally, (and possibly related to that), it would be desirable to find an
approximate ``free energy'', a function of the beliefs, whose stationary points
coincide with the fixed points of the Algorithm \ref{alg:LCBP}. This can be
done for many approximate inference methods (MF, BP, CVM, EP) so it is natural
to expect that the Loop Correction algorithm can also be seen as a minimization
procedure of a certain approximate free energy.  Despite some efforts, we have
not yet been able to find such a free energy. 

Recently, other loop correcting approaches (to the Bethe approximation) have
been proposed in the statistical physics community
\citep{ParisiSlanina05,ChertkovChernyak06b}. In particular,
\cite{ChertkovChernyak06b} have derived a series expansion of the \emph{exact}
normalizing constant $Z$ in terms of the BP solution. The first term of the series is
precisely the Bethe free energy evaluated at the BP fixed point. The number of terms
in the series is finite, but can be very large, even larger than the number of
total states of the graphical model. Each term is associated with a
``generalized loop'', which is a subgraph of the factor graph for which each
node has at least connectivity two. By truncating the series, it is possible to
obtain approximate solutions that improve on BP by taking into account a subset
of all generalized loops \citep{GomezMooijKappen06}. Summarizing, this approach
to loop corrections takes a subset of loops into account in an exact way,
whereas the loop correcting approach presented in this article takes all loops
into account in an approximate way. More experiments should be done to compare
both approaches.

Concluding, we have proposed a method to correct approximate inference methods
for the influence of loops in the factor graph. We have shown that it can
obtain very accurate results, also on real world graphical models,
outperforming existing approximate inference methods in terms of quality,
robustness or applicability. We have shown that it can be applied to problems
for which exact inference is infeasible. The rather large computation time
required is an issue which deserves further consideration; it may be possible
to use additional approximations on top of the loop correcting framework that
trade quality for computation time.


\acks{The research reported here is part of the Interactive Collaborative Information
Systems (ICIS) project, supported by the Dutch Ministry of Economic Affairs,
grant BSIK03024. We thank Bastian Wemmenhove for stimulating discussions and
for providing the PROMEDAS test cases.}


\newpage


\appendix

\section*{Appendix: Original approach proposed by \cite{MontanariRizzo05}}

For completeness, we describe the implementation based on cumulants as 
originally proposed by \cite{MontanariRizzo05}. The approach can be applied
in recursive fashion. Here we will only discuss the first recursion level.

Consider a graphical model which has only binary ($\pm1$-valued) variables
and factors that involve at most two variables. The corresponding
probability distribution can be parameterized in terms of the local fields
$\{\theta_i\}_{i\in\vars}$ and the couplings $\{J_{ij} = J_{ji}\}_{i\in\vars,j\in\del{i}}$:
  \begin{equation*}
  P(x) = \frac{1}{Z} \exp\left( \sum_{i\in\vars} \theta_i x_i + \frac{1}{2} \sum_{i\in\vars}\sum_{j\in\del{i}} J_{ij} x_i x_j \right).
  \end{equation*}

Let $i \in \vars$ and consider the corresponding cavity network of $i$. For
$\C{A} \subseteq \del{i}$, the cavity \emph{moment} $\M{i}{\C{A}}$ is defined
as the following expectation value under the cavity distribution:
  \begin{equation*}
  \M{i}{\C{A}} := \sum_{\xd{i}} \Zmx{i} \prod_{j\in\C{A}} x_j,
  \end{equation*}
where we will not explicitly distinguish between approximate and exact
quantities, following the physicists' tradition.\footnote{In \citep{MontanariRizzo05}, the notation 
$\tilde C^{(i)}_{\C{A}}$ is used for the cavity moment $\M{i}{\C{A}}$.}
The cavity \emph{cumulants} (also called ``connected correlations'') 
$\Cu{i}{\C{A}}$ are related to the moments in the following way:
  \begin{equation*}
  \M{i}{\C{A}} = \sum_{\C{B} \in \mathrm{Part}(\C{A})} \prod_{\C{E} \in \C{B}} \Cu{i}{\C{E}}
  \end{equation*}
where $\mathrm{Part}(\C{A})$ is the set of partitions of $\C{A}$.

We introduce some notation: we define for $\C{A} \subseteq \del{i}$:
  \begin{equation*}
  t_{i\C{A}} := \prod_{k \in \C{A}} \tanh J_{ik}.
  \end{equation*}
Further, for a set $X$, we denote the even subsets of $X$ as
$\C{P}_+(X) := \{Y \subseteq X: \nel{Y} \text{ is even}\}$ and the odd subsets
of $X$ as $\C{P}_-(X) := \{Y \subseteq X : \nel{Y} \text{ is odd}\}$.

Using standard algebraic manipulations, one can show that for $j \in \del{i}$,
the expectation value of $x_j$ in the absence of the 
interaction $\fac{ij} = \exp(J_{ij}x_ix_j)$ can be expressed in terms of cavity moments of $i$
as follows:
  \begin{equation}\label{eq:def_B_i_j}
  \frac{\displaystyle \sAeij t_{i\A} \M{i}{\A \cup j} + \tanh\theta_i \sAoij t_{i\A} \M{i}{\A \cup j}}{\displaystyle \sAeij t_{i\A} \M{i}{\A} + \tanh\theta_i \sAoij t_{i\A} \M{i}{\A}}.
  \end{equation}
On the other hand, the same expectation value can also be expressed in terms of cavity moments of $j$ as follows:
  \begin{equation}\label{eq:def_K_j_i}
  \frac{\tanh \theta_j \displaystyle \sBeji t_{j\B} \M{j}{\B} + \sBoji t_{j\B} \M{j}{\B}}{\displaystyle \sBeji t_{j\B} \M{j}{\B} + \tanh \theta_j \sBoji t_{j\B} \M{j}{\B}}.
  \end{equation}
The consistency equations are now given by equating \eref{eq:def_B_i_j} to \eref{eq:def_K_j_i}
for all $i \in \vars$, $j \in \del{i}$.

The expectation value of $x_i$ (in the presence of all interactions) can be similarly expressed in terms of cavity moments of $i$:
  \begin{equation}\label{eq:def_M_i}
  M_i := \sum_{x_i=\pm1}P(x_i)x_i = \frac{\tanh \theta_i \displaystyle \sAei t_{i\A} \M{i}{\A} + \sAoi t_{i\A} \M{i}{\A}}{\displaystyle \sAei t_{i\A} \M{i}{\A} + \tanh \theta_i \sAoi t_{i\A} \M{i}{\A}}.
  \end{equation}

\subsection*{Neglecting higher order cumulants}

Montanari and Rizzo proceed by neglecting cavity cumulants
$\Cu{i}{\C{A}}$ with $\nel{\C{A}} > 2$.  Denote by $\mathrm{Part}_2(\C{A})$
the set of all partitions of $\C{A}$ into subsets which have cardinality 2 at
most. Thus, neglecting higher order cavity cumulants amounts to the following
approximation:
  \begin{equation}\label{eq:def_Ma}
  \M{i}{\C{A}} \approx \sum_{\C{B} \in \mathrm{Part}_2(\C{A})} \prod_{\C{E} \in \C{B}} \Cu{i}{\C{E}}.
  \end{equation}
By some algebraic manupulations, one can express the consistency equations
\eref{eq:def_B_i_j}${} = {}$\eref{eq:def_K_j_i} in this approximation as follows:
  \begin{multline}\label{eq:MR_full}
  \Ma{i}{j} = \frac{\tanh \theta_j \displaystyle \sBeji t_{j\B} \Ma{j}{\B} + \sBoji t_{j\B} \Ma{j}{\B}}{\displaystyle \sBeji t_{j\B} \Ma{j}{\B} + \tanh \theta_j \sBoji t_{j\B} \Ma{j}{\B}} \\
{} - \sum_{k\in \dele{i}{j}} t_{ik} \Cu{i}{jk} \frac{\displaystyle \tanh\theta_i \sAeijk t_{i\A} \Ma{i}{\A} + \sAoijk t_{i\A} \Ma{i}{\A}}{\displaystyle \sAeij t_{i\A} \Ma{i}{\A} + \tanh\theta_i \sAoij t_{i\A} \Ma{i}{\A}}
  \end{multline}
One can use \eref{eq:def_Ma} to write \eref{eq:MR_full} in terms of the singleton
cumulants $\{\Ma{i}{j}\}_{i\in\vars,j\in\del{i}}$ and the pair cumulants 
$\{\Cu{i}{jk}\}_{i\in\vars,j\in\del{i},k\in\dele{i}{j}}$. Given (estimates of) the
pair cumulants, the consistency equations \eref{eq:MR_full} are thus fixed point
equations in the singleton cumulants.

The procedure is now:
\begin{itemize}
\item Estimate the pair cumulants $\{\Cu{i}{jk}\}_{i\in\vars,j\in\del{i},k\in\dele{i}{j}}$ using BP in combination with linear response
(called ``response propagation'' in \cite{MontanariRizzo05}).
\item Calculate the fixed point $\{\Ma{i}{j}\}_{i\in\vars,j\in\del{i}}$ of \eref{eq:MR_full} using the estimated the pair cumulants.
\item Use \eref{eq:def_M_i} in combination with \eref{eq:def_Ma} to calculate the final 
expectation values $\{M_j\}_{j\in\vars}$ using the estimated pair cumulants and the fixed 
point of \eref{eq:MR_full}.
\end{itemize}

\subsection*{Linearized version}

The update equations can be linearized by expanding up to first order in the
pair cumulants $\Cu{i}{jk}$. This yields
the following linearized consistency equation \citep{MontanariRizzo05}:
  \begin{equation}\label{eq:MR_linear}
  M^{(i)}_j = \T{j}{i} - \sum_{l\in\dele{i}{j}} \Omega^{(i)}_{j,l} t_{il} C^{(i)}_{jl} + \sum_{\{l_1,l_2\}:l_1,l_2 \in \dele{j}{i}} \Ga{j}{i,l_1 l_2} t_{j l_1} t_{j l_2} C^{(j)}_{l_1 l_2}
  \end{equation}
where
  \begin{align*}
  \T{i}{\C{A}} & := \tanh \left( \theta_i + \sum_{k \in \dele{i}{\C{A}}} \arctanh(t_{ik} M^{(i)}_k) \right) \\
  \Om{i}{jl} & := \frac{\T{i}{jl}}{1 + t_{il} M^{\setm i}_l \T{i}{jl}} \\
  \Ga{j}{i,l_1 l_2} & := \frac{\T{j}{i l_1 l_2} - \T{j}{i}}{1 + t_{jl_1} t_{jl_2} M^{(j)}_{l_1} M^{(j)}_{l_2} + t_{j l_1} M^{(j)}_{l_1} \T{j}{il_1 l_2} + t_{jl_2} M^{(j)}_{l_2} \T{j}{i l_1 l_2}}.
  \end{align*}
The final magnetizations \eref{eq:def_M_i} are, up to first order in the pair cumulants:
  \begin{equation*}
  M_j = \T{j}{} + \sum_{\{l_1,l_2\}:l_1,l_2 \in \del{j}^2} \Ga{j}{l_1 l_2} t_{j l_1} t_{j l_2} C^{(j)}_{l_1 l_2} + \C{O}(C^2)
  \end{equation*}
where
  \begin{equation*}
  \Ga{j}{l_1 l_2} := \frac{\T{j}{l_1 l_2} - \T{j}{}}{1 + t_{jl_1} t_{jl_2} M^{\setm j}_{l_1} M^{(j)}_{l_2} + t_{j l_1} M^{(j)}_{l_1} \T{j}{l_1 l_2} + t_{jl_2} M^{(j)}_{l_2} \T{j}{l_1 l_2}}.
  \end{equation*}

\vskip 0.2in
\bibliography{bp}

\begin{thebibliography}{24}
\providecommand{\natexlab}[1]{#1}
\providecommand{\url}[1]{\texttt{#1}}
\expandafter\ifx\csname urlstyle\endcsname\relax
  \providecommand{\doi}[1]{doi: #1}\else
  \providecommand{\doi}{doi: \begingroup \urlstyle{rm}\Url}\fi

\bibitem[Bethe(1935)]{Bethe35}
H.~Bethe.
\newblock Statistical theory of superlattices.
\newblock \emph{Proc. R. Soc. A}, 150:\penalty0 552--575, 1935.

\bibitem[Chertkov and Chernyak(2006)]{ChertkovChernyak06b}
Michael Chertkov and Vladimir~Y Chernyak.
\newblock Loop series for discrete statistical models on graphs.
\newblock \emph{Journal of Statistical Mechanics: Theory and Experiment},
  2006\penalty0 (06):\penalty0 P06009, 2006.
\newblock URL \url{http://stacks.iop.org/1742-5468/2006/P06009}.

\bibitem[Elidan et~al.(2006)Elidan, McGraw, and Koller]{ElidanMcGrawKoller06}
G.~Elidan, I.~McGraw, and D.~Koller.
\newblock Residual belief propagation: {I}nformed scheduling for asynchronous
  message passing.
\newblock In \emph{Proceedings of the Twenty-second Conference on Uncertainty
  in AI (UAI)}, Boston, Massachussetts, July 2006.

\bibitem[G{\'o}mez et~al.(2006)G{\'o}mez, Mooij, and
  Kappen]{GomezMooijKappen06}
V.~G{\'o}mez, J.~Mooij, and H.~Kappen.
\newblock \emph{In preparation}, 2006.

\bibitem[Heskes et~al.(2003)Heskes, Albers, and Kappen]{HeskesAlbersKappen03}
Tom Heskes, C.A. Albers, and Hilbert~J. Kappen.
\newblock Approximate inference and constrained optimization.
\newblock In \emph{Proc. of the 19th Annual Conf. on Uncertainty in Artificial
  Intelligence (UAI-03)}, pages 313--320, San Francisco, CA, 2003. Morgan
  Kaufmann Publishers.

\bibitem[Jaakkola and Jordan(1999)]{JaakkolaJordan99}
Tommi Jaakkola and Michael~I. Jordan.
\newblock Variational probabilistic inference and the {QMR}-{DT} network.
\newblock \emph{Journal of Artificial Intelligence Research}, 10:\penalty0
  291--322, 1999.
\newblock URL \url{http://www.jair.org/papers/paper583.html}.

\bibitem[Kikuchi(1951)]{Kikuchi51}
R.~Kikuchi.
\newblock A theory of cooperative phenomena.
\newblock \emph{Phys. Rev.}, 81:\penalty0 988--1003, 1951.

\bibitem[Kschischang et~al.(2001)Kschischang, Frey, and
  Loeliger]{KschischangFreyLoeliger01}
Frank~R. Kschischang, Brendan~J. Frey, and Hans-Andrea Loeliger.
\newblock Factor graphs and the {Sum-Product Algorithm}.
\newblock \emph{IEEE Trans. Inform. Theory}, 47\penalty0 (2):\penalty0
  498--519, February 2001.

\bibitem[M{\'e}zard et~al.(1987)M{\'e}zard, Parisi, and
  Virasoro]{MezardParisiVirasoro87}
M.~M{\'e}zard, G.~Parisi, and M.~A. Virasoro.
\newblock \emph{Spin glass theory and beyond}.
\newblock World Scientific, Singapore, 1987.

\bibitem[Minka(2001)]{Minka01}
Thomas Minka.
\newblock {Expectation Propagation} for approximate {Bayesian} inference.
\newblock In \emph{Proc. of the 17th Annual Conf. on Uncertainty in Artificial
  Intelligence (UAI-01)}, pages 362--369, San Francisco, CA, 2001. Morgan
  Kaufmann Publishers.

\bibitem[Minka and Qi(2004)]{MinkaQi04}
Thomas Minka and Yuan Qi.
\newblock Tree-structured approximations by {Expectation Propagation}.
\newblock In Sebastian Thrun, Lawrence Saul, and Bernhard {Sch\"{o}lkopf},
  editors, \emph{Advances in Neural Information Processing Systems 16},
  Cambridge, MA, 2004. MIT Press.

\bibitem[Montanari and Rizzo(2005)]{MontanariRizzo05}
Andrea Montanari and Tommaso Rizzo.
\newblock How to compute loop corrections to the {Bethe} approximation.
\newblock \emph{Journal of Statistical Mechanics: Theory and Experiment},
  2005\penalty0 (10):\penalty0 P10011, 2005.
\newblock URL \url{http://stacks.iop.org/1742-5468/2005/P10011}.

\bibitem[Mooij and Kappen(2005)]{MooijKappen05c}
J~M Mooij and H~J Kappen.
\newblock On the properties of the bethe approximation and loopy belief
  propagation on binary networks.
\newblock \emph{Journal of Statistical Mechanics: Theory and Experiment},
  2005\penalty0 (11):\penalty0 P11012, 2005.
\newblock URL \url{http://stacks.iop.org/1742-5468/2005/P11012}.

\bibitem[Parisi(1988)]{Parisi88}
G.~Parisi.
\newblock \emph{Statistical Field Theory}.
\newblock Addison-Wesley, Redwood City, Ca, 1988.

\bibitem[Parisi and Slanina(2005)]{ParisiSlanina05}
Giorgio Parisi and Frantisek Slanina.
\newblock Loop expansion around the {Bethe-Peierls} approximation for lattice
  models.
\newblock \emph{arXiv.org preprint}, cond-mat/0512529, 2005.

\bibitem[Pearl(1988)]{Pearl88}
J.~Pearl.
\newblock \emph{Probabilistic Reasoning in Intelligent systems: Networks of
  Plausible Inference}.
\newblock Morgan Kaufmann, San Francisco, CA, 1988.

\bibitem[Pelizzola(2005)]{Pelizzola05}
A.~Pelizzola.
\newblock Cluster variation method in statistical physics and probabilistic
  graphical models.
\newblock \emph{J. Phys. A: Math. Gen.}, 38:\penalty0 R309--R339, August 2005.

\bibitem[Shwe et~al.(1991)Shwe, Middleton, Heckerman, Henrion, Horvitz,
  Lehmann, and Cooper]{Shwe+91}
M.~A. Shwe, B.~Middleton, D.~E. Heckerman, M.~Henrion, E.~J. Horvitz, H.~P.
  Lehmann, and G.~F. Cooper.
\newblock Probabilistic diagnosis using a reformulation of the
  {INTERNIST-1/QMR} knowledge base. {I. The} probabilistic model and inference
  algorithms.
\newblock \emph{Methods of information in Medicine}, 30\penalty0 (4):\penalty0
  241--255, October 1991.

\bibitem[Takikawa and D'Ambrosio(1999)]{TakikawaDAmbrosio99}
Masami Takikawa and Bruce D'Ambrosio.
\newblock Multiplicative factorization of noisy-max.
\newblock In \emph{Proceedings of the 15th Annual Conference on Uncertainty in
  Artificial Intelligence (UAI-99)}, pages 622--63, San Francisco, CA, 1999.
  Morgan Kaufmann.

\bibitem[Welling and Teh(2004)]{WellingTeh04}
Max Welling and Yee~Whye Teh.
\newblock Linear response for approximate inference.
\newblock In Sebastian Thrun, Lawrence Saul, and Bernhard {Sch\"{o}lkopf},
  editors, \emph{Advances in Neural Information Processing Systems 16},
  Cambridge, MA, 2004. MIT Press.

\bibitem[Welling et~al.(2005)Welling, Minka, and Teh]{WellingMinkaTeh05}
Max Welling, Thomas Minka, and Yee~Whye Teh.
\newblock {Structured Region Graphs}: Morphing {EP} into {GBP}.
\newblock In \emph{Proceedings of the 21th Annual Conference on Uncertainty in
  Artificial Intelligence (UAI-05)}, page 609, Arlington, Virginia, 2005. AUAI
  Press.

\bibitem[Wiegerinck et~al.(1999)Wiegerinck, Kappen, ter Braak, ter Burg,
  Nijman, O, and Neijt]{Wiegerinck+99}
W.~Wiegerinck, H.~J. Kappen, E.~W. M.~T. ter Braak, W.~J. P.~P. ter Burg, M.~J.
  Nijman, Y.~L. O, and J.~P. Neijt.
\newblock Approximate inference for medical diagnosis.
\newblock \emph{Pattern Recognition Letters}, 20:\penalty0 1231--1239, 1999.

\bibitem[Yedidia et~al.(2005)Yedidia, Freeman, and
  Weiss]{YedidiaFreemanWeiss05}
J.~S. Yedidia, W.~T. Freeman, and Y.~Weiss.
\newblock Constructing free-energy approximations and {Generalized} {Belief}
  {Propagation} algorithms.
\newblock \emph{IEEE Transactions on Information Theory}, 51\penalty0
  (7):\penalty0 2282--2312, July 2005.

\bibitem[Yuille(2002)]{Yuille02}
A.~L. Yuille.
\newblock {CCCP} algorithms to minimize the {Bethe} and {Kikuchi} free
  energies: Convergent alternatives to belief propagation.
\newblock \emph{Neural Computation}, 14\penalty0 (7):\penalty0 1691--1722,
  2002.

\end{thebibliography}

\clearpage

\end{document}